%% file: example_paper.tex
\documentclass{article}

\usepackage{microtype}
\usepackage{graphicx}
\usepackage{subcaption}
\usepackage{booktabs} 

\usepackage[pagebackref,breaklinks,colorlinks,citecolor=cvprblue,hypertexnames=false]{hyperref}

\usepackage[dvipsnames,table]{xcolor}
\definecolor{lightblue}{RGB}{220, 230, 241}
\definecolor{lightpink}{RGB}{248, 228, 235}
\definecolor{lightyellow}{RGB}{255, 249, 227}
\usepackage{tikz} 
\usepackage{amsmath}

\usepackage[accepted]{icml2025}

\usepackage{amsmath}
\usepackage{amssymb}
\usepackage{mathtools}
\usepackage{amsthm}

\usepackage[dvipsnames,table]{xcolor}
\definecolor{lightblue}{RGB}{220, 230, 241}
\definecolor{lightpink}{RGB}{248, 228, 235}
\definecolor{lightyellow}{RGB}{255, 249, 227}
\usepackage{tikz} 
\usepackage{pdfpages}
\usepackage{placeins}
\usepackage{minitoc}
\usepackage{fancyhdr}  
\usepackage{float}
\usepackage{array}
\newcolumntype{P}[1]{>{\centering\arraybackslash}m{#1}}
\usepackage{newunicodechar}

\usepackage{multirow}

\usepackage[capitalize,noabbrev]{cleveref}

\theoremstyle{plain}

\theoremstyle{definition}

\theoremstyle{remark}

\usepackage[textsize=tiny]{todonotes}

\icmltitlerunning{Bridging vision language model (VLM) evaluation gaps with a framework for scalable and cost-effective benchmark generation}

\begin{document}

\twocolumn[
\icmltitle{Bridging vision language model (VLM) evaluation gaps with a framework for scalable and cost-effective benchmark generation}



\icmlsetsymbol{tech}{+}
\icmlsetsymbol{dagger}{\dag}

\begin{icmlauthorlist}
\icmlauthor{Tim Rädsch}{dkfz,hidkfz,heidelberguni,tech,dagger}
\icmlauthor{Leon Mayer}{dkfz,tech}
\icmlauthor{Simon Pavicic}{dkfz}
\icmlauthor{A. Emre Kavur}{dkfz,hidkfz}
\icmlauthor{Marcel Knopp}{dkfz,heidelberguni}
\icmlauthor{Barış Öztürk}{dkfz}
\icmlauthor{Klaus Maier-Hein}{dkfz,hidkfz,heidelberguni}
\icmlauthor{Paul F. Jaeger}{dkfz,hidkfz}
\icmlauthor{Fabian Isensee}{dkfz,hidkfz}
\icmlauthor{Annika Reinke}{dkfz,hidkfz,dagger}
\icmlauthor{Lena Maier-Hein}{dkfz,hidkfz,heidelberguni,dagger} 
\vspace{0.5em} \\
\small $\textsuperscript{1}$German Cancer Research Center (DKFZ) Heidelberg, Germany, $\textsuperscript{2}$Helmholtz Imaging, DKFZ, Heidelberg, Germany, $\textsuperscript{3}$Heidelberg University, Heidelberg, Germany, + core contributor, \textdagger{} project lead
\end{icmlauthorlist}

\icmlaffiliation{dkfz}{German Cancer Research Center (DKFZ) Heidelberg, Germany}
\icmlaffiliation{hidkfz}{Helmholtz Imaging, DKFZ, Heidelberg, Germany}
\icmlaffiliation{heidelberguni}{Heidelberg University, Heidelberg, Germany}

\icmlcorrespondingauthor{{tim.raedsch; l.maier-hein}}{@dkfz-heidelberg.de}

\icmlkeywords{Machine Learning, ICML}

\vskip 0.3in
]




\begin{figure*}
  \centering
  \includegraphics[width=\textwidth]{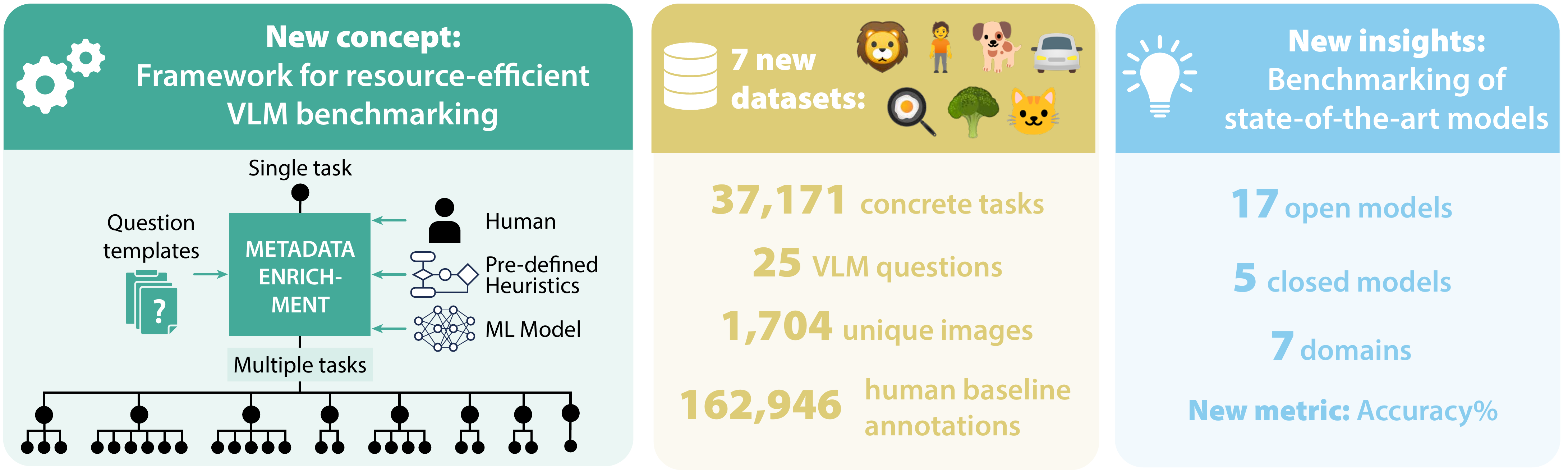}
\caption{\textbf{Summary of contributions}. (1) New concept: We propose a new framework for the resource-efficient creation of domain-specific VLM benchmarks. It is based on the concept of task augmentation designed for creating multiple tasks from a single existing task using metadata annotations from multiple sources (humans, pre-defined heuristics, models). (2) 7 new datasets: We apply our framework to generate seven domain-specific VLM benchmarks with highly reliable reference data. As a unique feature compared to existing benchmarks, we quantify the ambiguity of each question for each image by acquiring human answers from a total of six raters. (3) New insights: We apply our framework to a total of 22 open and frontier closed models to demonstrate the benefit of task augmentation and to shed light on current VLM capabilities.}
  \label{fig:figure1}
\end{figure*}

\input{sec/0_abstract}    
\input{sec/1_intro}
\input{sec/2_related_work}
\input{sec/3_methods}

\input{sec/4_experiments_and_results}

\input{sec/5_discussion}

\section*{Impact Statement}
This paper advances Machine Learning by enabling researchers to benchmark with their own data on a minimal budget. All human annotations were sourced from a reputable company following ethical guidelines.

\bibliography{example_paper}
\bibliographystyle{icml2025}

\newpage

\appendix
\onecolumn

\input{sec/X_suppl}


\end{document}

%% file: sec/0_abstract.tex
\begin{abstract}
Reliable evaluation of AI models is critical for scientific progress and practical application. While existing VLM benchmarks provide general insights into model capabilities, their heterogeneous designs and limited focus on a few imaging domains pose significant challenges for both cross-domain performance comparison and targeted domain-specific evaluation. To address this, we propose three key contributions: (1) a framework for the resource-efficient creation of domain-specific VLM benchmarks enabled by task augmentation for creating multiple diverse tasks from a single existing task, (2) the release of new VLM benchmarks for seven domains, created according to the same homogeneous protocol and including 162,946 thoroughly human-validated answers, and (3) an extensive benchmarking of 22 state-of-the-art VLMs on a total of 37,171 tasks, revealing performance variances across domains and tasks, thereby supporting the need for tailored VLM benchmarks. Adoption of our methodology will pave the way for the resource-efficient domain-specific selection of models and guide future research efforts toward addressing core open questions.
\end{abstract}

%% file: sec/1_intro.tex
\addtocontents{toc}{\protect\setcounter{tocdepth}{-1}}  

\section{Introduction}
\label{sec:intro}
The reliable and objective performance assessment, i.e., validation of AI models is crucial for both the measurement of scientific progress and translation into practice. Benchmarking for traditional narrow, task-specific AI already comes with numerous challenges~\cite{myllyaho2021systematic}, but validation has proven to be even more complex and error-prone in the emerging field of generalist multimodal foundation models~\cite{schaeffer2024areemergent}. In the context of Vision-Language Models (VLMs), one issue that has received limited attention is the heterogeneous and often non-targeted nature of model validation~\cite{tong2024cambrian1fullyopenvisioncentric, tong2024eyes}. Widely used VLM benchmarks span diverse domains and encompass a variety of tasks, providing a broad view of model capabilities across different contexts~\cite{fu2024blink,liu2024mmbench,ying2024mmtbench,altahan2024unibench,yue2023mmmu}. 

We identify three key trends that highlight the critical need for personalized benchmarking approaches:

\textbf{Domain-specific benchmark demand:} Numerous datasets and benchmarks are continually being released in the general computer vision field. According to our analyses,  $\sim$400 out of the 2,700 CVPR 2024 publications propose a new or modified dataset as detailed in~\cref{appendix_cvpr_paper_analysis}. These benchmarks cover a wide range of domains, from autonomous driving to wildlife monitoring, underscoring the need for \textit{domain-specific} benchmarks.

\textbf{Popular arena platforms do not scale from an individual user’s perspective:}
Arena-style platforms such as Chatbot Arena\footnote{lmarena.ai/?leaderboard; see the Arena(Vision) tab} or WildVision Arena\footnote{huggingface.co/spaces/WildVision/vision-arena} allow users to submit single tasks and rate the outputs of different (anonymized) models. The aggregated user ratings, in turn, can be used for the objective and comparative assessment of models. While this allows for personalized and domain-relevant evaluation, large-scale assessment from a single user perspective would be cumbersome due to the required annotation effort.

\textbf{Homogeneous evaluation:} Most existing VLM benchmarks \cite{fu2024blink, yue2023mmmu, wang2024journeybench, zhang2024mmereal}, generally evaluate models using a single question per image. While this can suffice when large datasets are available—allowing for a broad range of tasks—domain experts with smaller, curated datasets face a more significant limitation. From a resource standpoint, image acquisition may also be expensive, and few tasks emerge if there is only one question per image. Furthermore, such an approach provides little insight into whether a VLM truly comprehends broad aspects of an image’s semantic content.

Taking these three trends together we conclude that there is a lack of guidance on how to set up a framework that enables personalized, domain-specific benchmarking in a resource-efficient manner.
Such a framework must address the scarcity of labeled data, leverage task diversity by systematically generating multiple questions per image, and maintain resource efficiency to ensure accessibility for researchers working in specialized fields, such as wildlife monitoring, or autonomous driving. 

In this work, we propose a resource-efficient framework for creating domain-specific VLM benchmarks via task augmentation. Our approach transforms a single type of annotation—instance segmentation—into a diverse set of tasks that test a broad range of perception abilities, such as object counting, occlusion detection, brightness comparison, and more.
Specifically, we focus on 2D natural images that either (1) already include instance segmentations or (2) can be annotated using recent advances in semi-automatic labeling tools (e.g., SAM~\cite{ravi2024sam}). This approach allows even domains with limited labeled data to efficiently generate custom evaluation tasks. Our main contribution, summarized in~\cref{fig:figure1}, is a \textbf{resource-efficient framework} for creating domain-specific VLM benchmarks via task augmentation, transforming a single type of annotation (instance segmentation) into a diverse set of tasks. We \textbf{apply} this framework \textbf{to create seven new domain-specific VLM benchmarks} and comprehensively \textbf{evaluate 22 open and closed VLMs on over 37,000 tasks }(for the full model list see \cref{appendix_model_overview}). To establish strong reference points for model evaluation, we collected an additional 162,946 human baseline answers corresponding to 37,171 questions across 1,704 images.

\begin{figure*}[ht]
    \centering
    \includegraphics[width=1\linewidth]{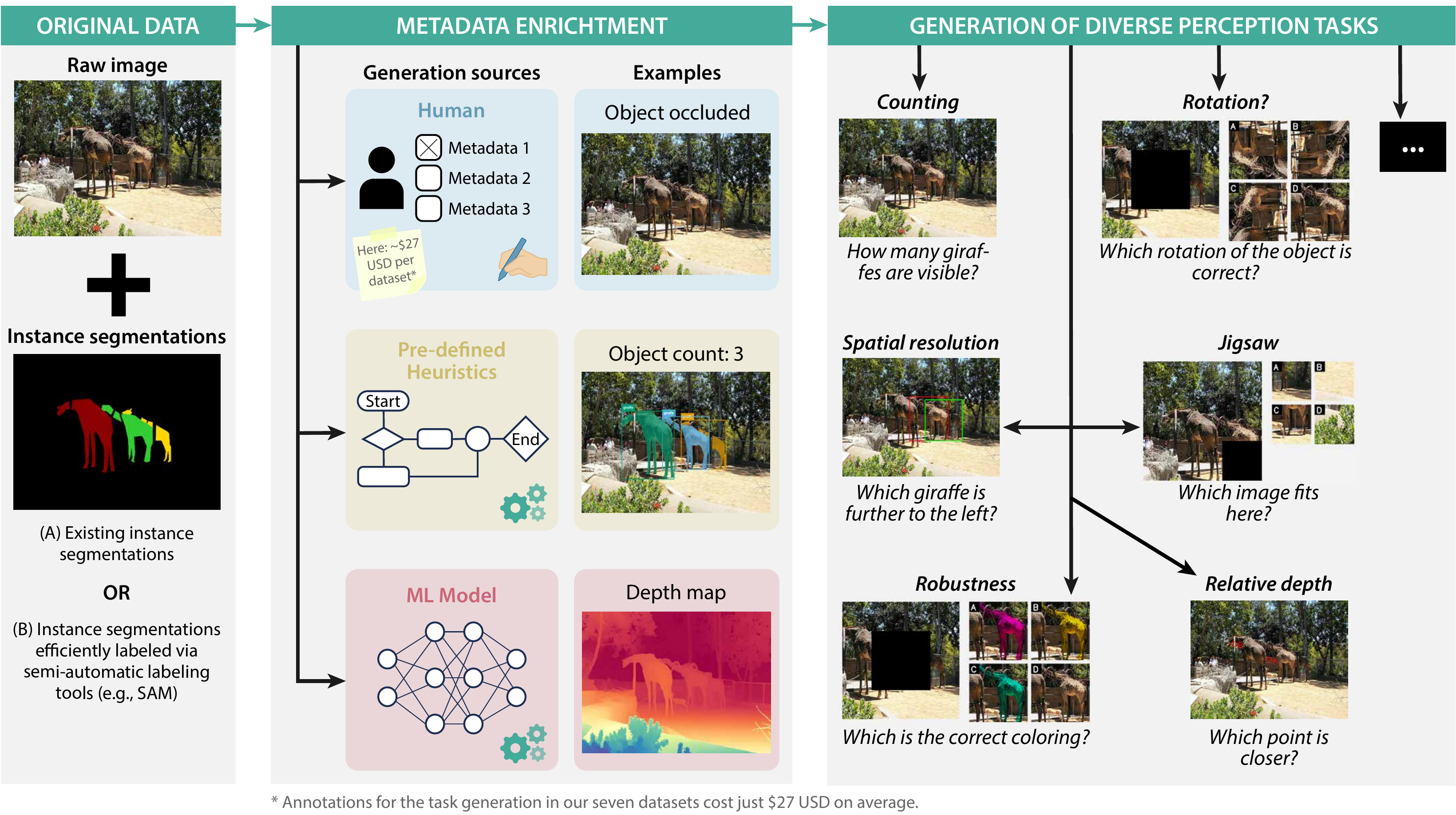}
    \caption{\textbf{Framework for resource-efficient in-domain benchmarking.} Starting from a single task with fine-grained annotations (here: instance segmentations), metadata for each image is obtained from both automatic sources (heuristics and models) and a small number of manual sources (human annotations). This process transforms the initial task into a collection of tasks, enabling resource-efficient and easy to use in-domain benchmarking of general VLM capabilities while maintaining cross-domain comparability.}
    \label{fig:metadataaug}
\end{figure*}

%% file: sec/2_related_work.tex
\section{Related Work}
\label{sec:related_work}

\subsection{Vision-Language Benchmarks}
Recent studies propose a range of evaluation benchmarks for VLMs, varying in size, number, and type of VL capabilities. Examples include Blink \cite{fu2024blink} and MMBench~\cite{liu2024mmbench} ($>$3,000 multiple-choice questions each), and MME~\cite{fu2024mme} (Yes/No questions on perception and cognition). The largest benchmarks include MMT-Bench~\cite{ying2024mmtbench} ($>$31,000 questions), MME-RealWorld~\cite{zhang2024mmereal} ($>$29,000 image-question pairs), and MMMU \cite{yue2023mmmu} ($>$11,500 questions). While these benchmarks cover multiple VL capabilities and domains, they require extensive labeling efforts. For example, MME-RealWorld involved 25 annotators and seven VLM experts, MMMU relied on 50 college students, while MMT-Bench lacks details on annotator numbers. Other benchmarks focus on much smaller question sets~\cite{chen2024we, yu2024mm}, integrating multiple existing benchmarks~\cite{jiang2024eff,altahan2024unibench}, or collecting individual human preferences~\cite{lu2024wildvision,xu2023lvlmehub}. \citet{tong2024cambrian1fullyopenvisioncentric} present a critical examination of multimodal LLM benchmarks.

Despite the variety of datasets and tasks, a resource-efficient and generalizable approach that enables extensive evaluation of VLMs across multiple (domain-specific) tasks is still lacking. Our framework addresses this gap by empowering users to create domain-specific VLM perception benchmarks from just a few images.

\subsection{Task augmentation and metadata}
Task augmentation refers to generating multiple diverse tasks from a single existing task~\cite{muennighoff2023octopack}. While task augmentation has been addressed from various directions \cite{johnson2017clevr,zhang2024task, Zamir_2018_CVPR,wang2023instruct4v, wang2024journeybench,kuznetsova2020v4,krishna2017genome}, an easy to use framework for evaluating VLMs by domain users on their own images is still missing. The closest works to ours are \citet{zhang2024task} and \citet{zhang2024provision}, which programmatically generate benchmarks using a library of visual assets and task templates. A comprehensive comparison to other task augmentations works and their applicability is provided in~\cref{appendix_task_augmentation_comparison}.

\subsection{Resource-efficient VLM benchmarking}
Most existing benchmarks often focus on performance metrics without considering the human and computational resources required to generate a benchmark (see, e.g.,~\cite{fu2024blink,liu2024mmbench}). The work that has been done on efficient benchmarking has been focused in the realm of unimodal language models~\cite{polo2024tinybenchmarks, perlitz2023efficient}. An exception has been~\citet{ging2024open}, who investigated the automatic creation of VLM benchmarks from classification datasets. Nevertheless, the increasing prominence of VLMs in research and industry ~\cite{li2024multimodal, yang2023dawn} is not yet reflected in efforts to increase efficiency during benchmark creation.

\begin{figure*}[t]
    \centering
\includegraphics[width=\linewidth]{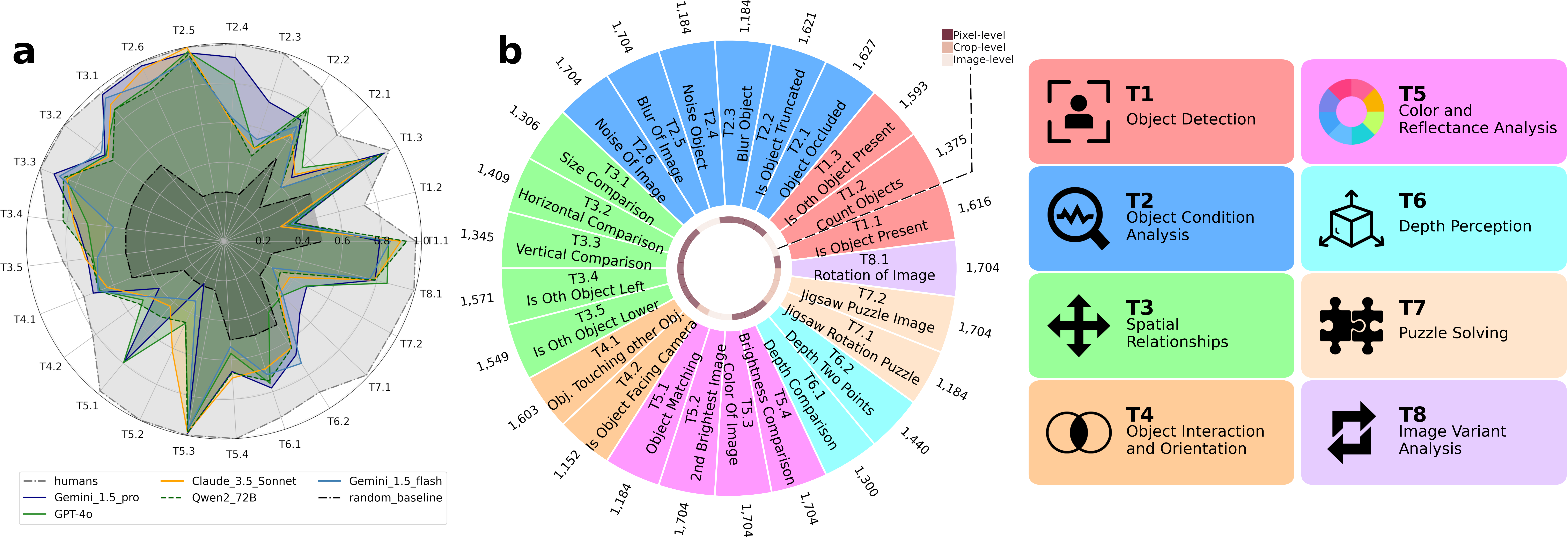}
\caption{\textbf{Our framework yields a diverse set of tasks.} (a) The spider diagram illustrates high Accuracy variability across tasks for the VLMs. We present the results of all the best ranked models while a comprehensive performance summary for all 22 tested models can be found in~\Cref{appendix_spider_figures}. (b) Based on a single image with instance segmentations, our framework enables the  generation of 25 tasks from eight different vision-language categories, ranging from pixel-level to image-level perception.}
    \label{fig:fig3}
\end{figure*}

%% file: sec/3_methods.tex
\section{Methods}

\subsection{Framework for resource-efficient in-domain benchmarking
}
The framework for resource-efficient in-domain benchmarking is depicted in~\cref{fig:metadataaug}. Starting with domain images that include instance segmentations (existing or created with semi-automatic labeling tools, such as SAM~\cite{ravi2024sam}), metadata for each image is acquired from multiple sources (humans, pre-defined heuristics, and models) to transform the single task into a collection of perception tasks.

For our seven new datasets, we use existing instance segmentation as the core perceptual task to generate the diverse set of VLM benchmark tasks depicted in \Cref{appendix_overviewvlmtasks} (examples in ~\Cref{fig:example_cow} and more detailed in ~\Cref{sec_sub:app_person}).

The metadata enrichment is derived from three sources: 

1) \underline{Human annotators} were used to generate information that cannot be extracted from the existing annotations or using established models. To this end, we outsourced annotations to a professional annotation company (Quality Match GmbH in Heidelberg). Specifically, human raters were tasked with determining the presence of occlusion and truncation in the images. Furthermore, they were asked to assess the direction in which the objects were facing. These annotations cost $\sim$27 USD on average with a total turnover time of two days. 

2) \underline{Pre-defined heuristics and rules} were employed to transform existing information into metadata. For example, instance segmentations were utilized to quantify the number of objects within a specific class or to determine whether specific instance segmentation masks were touching each other. 

3) \underline{An existing depth foundation model}, Depth Anything v2~\cite{yang2024depth}, was used to generate depth maps for each image.

\subsection{Seven new datasets from diverse domains}
We applied our proposed framework to images from seven different domains. Overall, the input images and instance segmentations for our framework were extracted from KITTI ~\cite{geiger2012kitti}, COCO~\cite{lin2014microsoft}, and COCONut~\cite{deng2024coconut}. In summary, we added ~300,000 metadata annotations to a total of 1,704 images across seven domains. This includes 15 annotations per object (e.g. occlusion, relative\_size, segmask\_touches\_segmask, or average\_depth). For truncation, occlusion, and direction, we obtained up to five annotations per object from human annotators (UI example is displayed in ~\Cref{appendix_example_quality_match}). Early stopping was applied when four annotators reached a consensus. The complete list is provided in Appendix \Cref{appendix_three_sources_of_metadata}.

The metadata were then used to define a set of 25 different VLM tasks (see~\cref{fig:fig3}), including six tasks concerning the entire image, 13 related to individual objects, and six focused on object pairs.

\textbf{Setup for automatic task processing after metadata extraction:}
To create a concrete list of vision-language tasks for each image we employed a systematic process. We began by prioritizing images in the datasets that featured a higher number of classes and objects to maximize task diversity and complexity. Next, specific criteria for each task were evaluated to ensure appropriate task generation for each image. For instance, in tasks requiring the comparison of two objects, it was essential that both objects were present in the image and belonged to the relevant classes. Furthermore, we established minimum thresholds for various measures, such as requiring a substantial depth difference between objects, to ensure the correct answers for the task could be reliably determined. Overall, our objective was to generate as many of the 25 different tasks as possible for each image. No LLMs or VLMs were used for task generation, as these methods are prone to injecting hallucinations~\cite{wang2023instruct4v,wang2024journeybench}. We prioritized quality and reliability instead.

\textbf{Human ambiguity baseline}: To rate the difficulty and ambiguity for each of the 37,171 tasks, we further acquired annotations from six human raters per image. We implemented early stopping if four raters reached agreement on a task. Overall, this resulted in 162,946 human reference annotations. An overview of the resulting datasets is provided in~\cref{tab:dataset_statistics}and exemplary images for all generated datasets are included in~\cref{domain_sets_appendix}.

\begin{table}[t]
\centering
\setlength{\tabcolsep}{4pt}  
\footnotesize
\begin{tabular}{|c|c|c|c|c|p{1.3cm}|}
\hline
\textbf{Domain} & \textbf{Icon} & \textbf{\#Images} & \textbf{\#Objects} & \textbf{\#Tasks} & \textbf{\#Human Annot.} \\
\hline
\textbf{Wildlife} & \includegraphics[width=0.4cm]{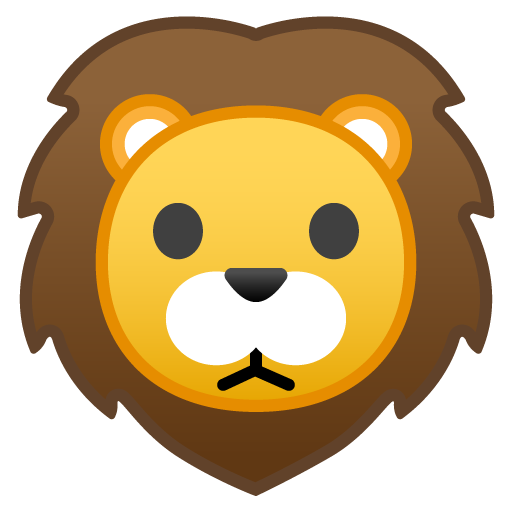} & 268 & 853 & 5,528 & 24,024 \\
\textbf{Persons} & \includegraphics[width=0.4cm]{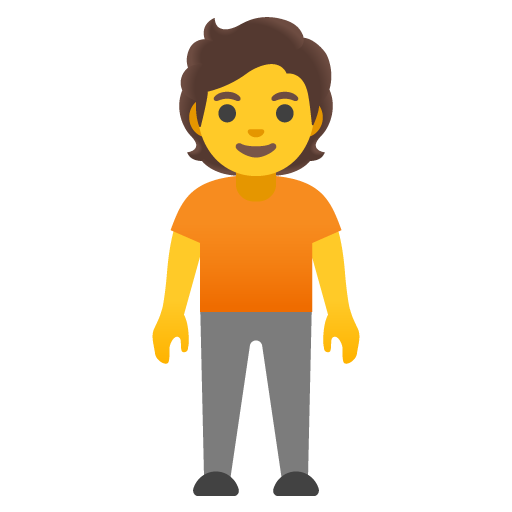} & 250 & 7,812 & 6,122 & 26,548 \\
\textbf{Vehicles} & \includegraphics[width=0.4cm]{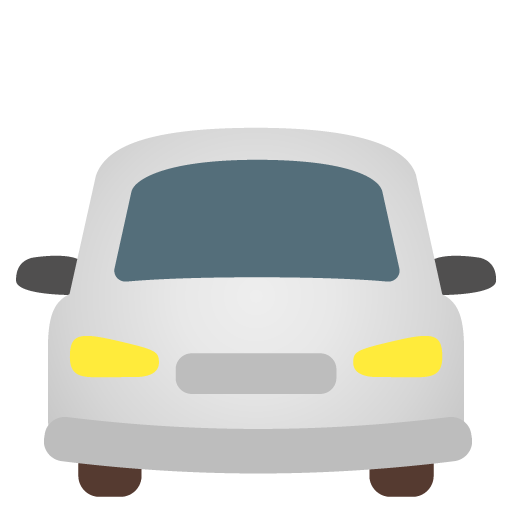} & 235 & 2,199 & 5,219 & 22,976 \\
\textbf{Animals} & \includegraphics[width=0.4cm]{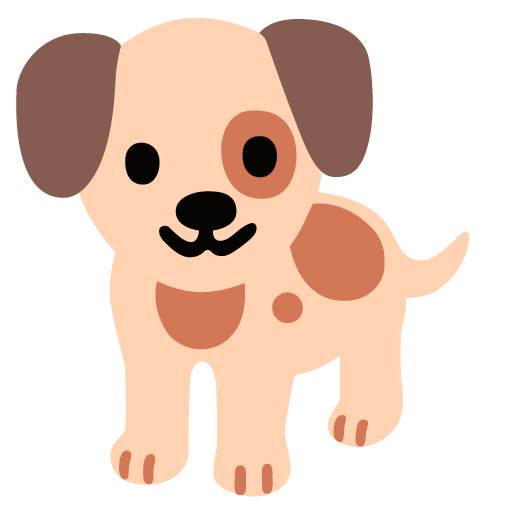} & 273 & 1,162 & 5,724 & 24,907 \\
\textbf{Kitchen} & \includegraphics[width=0.4cm]{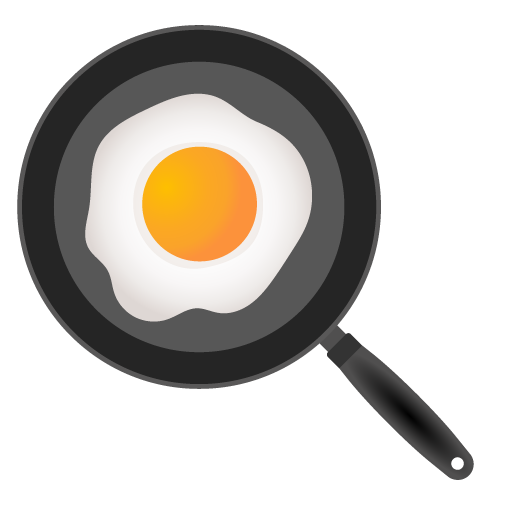} & 272 & 2,143 & 5,332 & 23,793 \\
\textbf{Food} & \includegraphics[width=0.4cm]{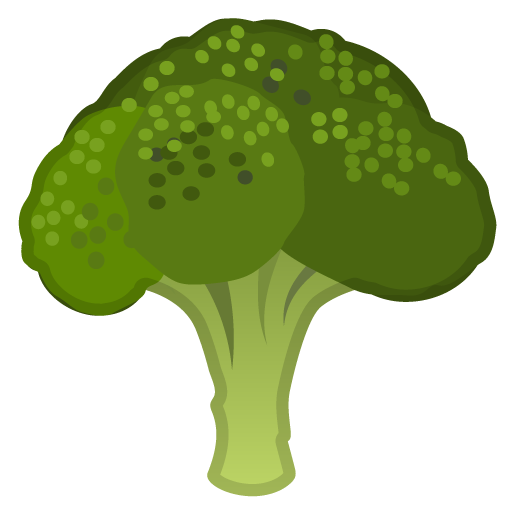} & 236 & 5,673 & 5,249 & 23,221 \\
\textbf{Kitti} & \includegraphics[width=0.4cm]{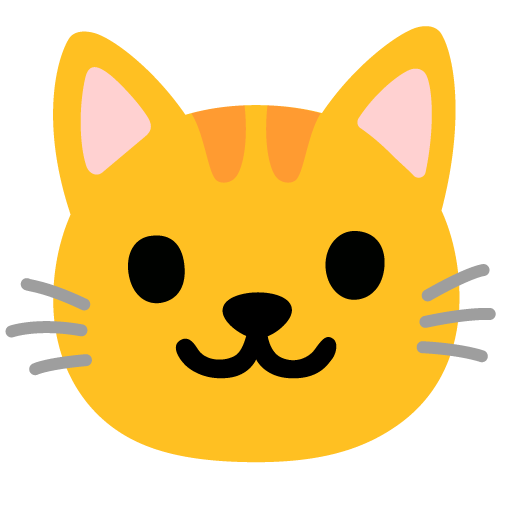} & 170 & 1,458 & 3,997 & 17,477 \\
\hline
\textbf{Total} & & \textbf{1,704} & \textbf{21,300} & \textbf{37,171} & \textbf{162,946} \\
\hline
\end{tabular}
\caption{\textbf{Dataset statistics across different domains.} The table presents the total number of images, objects, tasks, and human annotations across all domains.}
\label{tab:dataset_statistics}
\end{table}

\subsection{Benchmarking strategy}
VLM benchmarking results can vary substantially with various factors, such as the images used, the domain, and the applied prompts. This often renders comparison of results across papers infeasible. For example, Accuracy is a prevalence-dependent metric, meaning that results should not be compared across datasets. To address this bottleneck, we fully homogenized our benchmarking framework using the proposed framework.

\textbf{Model selection:}
We selected 22 frontier and open VLMs of various sizes and from various providers and sources, as illustrated in \Cref{appendix_model_overview}. The oldest model was released in January 2024, while the most recent one included was released at the end of September 2024.

 \textbf{Benchmarking workflow:} 
 To ensure fair and consistent evaluation of all selected VLMs, we developed a standardized benchmarking workflow applied uniformly across all models. We assessed them in a zero-shot setting without any additional fine-tuning or domain-specific training. We strictly followed the configurations and setups recommended by each model's authors, using the exact settings provided in their official repositories (e.g., on Hugging Face) to ensure that each model was evaluated under conditions intended by its creators. Each model was provided with a carefully crafted text prompt alongside the corresponding image. 
To eliminate potential ambiguities in the questions, we conducted iterative testing of these prompts among human evaluators in our department. Through four rounds of refinement, we adjusted the prompts until all four human evaluators consistently agreed on their interpretation. Furthermore, we evaluated the sensitivity of the VLMs to variations in image markers, as many questions involved marked objects. Altering the box colors used to highlight objects—from green and red to other colors—resulted in slight performance fluctuations in both directions across different VLMs. To maintain consistency, we used the commonly recognized colors red and green, assigning them to objects at random.

\textbf{VLM tasks:} We evaluated the models on a comprehensive set of 25 tasks derived from our task augmentation framework (overview in~\cref{fig:fig3}, full list in~\cref{appendix_overviewvlmtasks} and examples per dataset in~\Cref{domain_sets_appendix}). Each task was associated with specific evaluation criteria and standardized prompts. For instance, when dealing with multiple-choice questions or tasks involving object selection, we established clear guidelines on how options were presented and how objects were chosen within images. This attention to detail ensured that the evaluation was both rigorous and reproducible.

\begin{table*}[ht]
\centering
\small
\begin{tabular}{lcccccccc}

\toprule
 & Overall & Wildlife & Animals & Kitti & Person & Vehicles & Food & Kitchen \\
& & \includegraphics[width=0.03\textwidth]{figures/icons_iclr/wildlife.png} & \includegraphics[width=0.03\textwidth]{figures/icons_iclr/animals.png} & \includegraphics[width=0.03\textwidth]{figures/icons_iclr/kitti.png} & \includegraphics[width=0.03\textwidth]{figures/icons_iclr/person.png} & \includegraphics[width=0.03\textwidth]{figures/icons_iclr/vehicles.png} & \includegraphics[width=0.03\textwidth]{figures/icons_iclr/food.png} & \includegraphics[width=0.03\textwidth]{figures/icons_iclr/kitchen.png} \\\midrule
Human & 93.7 & 93.4 & 93.9 & 94.6 & 95.2 & 92.4 & 93.6 & 92.6 \\
\toprule
Gemini\_1.5\_pro & \cellcolor[rgb]{1,0.843,0}72.4 & \cellcolor[rgb]{1,0.843,0}74.6 & \cellcolor[rgb]{1,0.843,0}75.4 & \cellcolor[rgb]{1,0.843,0}78.0 & \cellcolor[rgb]{1,0.843,0}70.8 & \cellcolor[rgb]{1,0.843,0}71.7 & \cellcolor[rgb]{1,0.843,0}70.0 & \cellcolor[rgb]{1,0.843,0}66.6 \\
GPT-4o & \cellcolor[rgb]{0.753,0.753,0.753}69.8 & \cellcolor[rgb]{0.565,0.933,0.565}71.4 & \cellcolor[rgb]{0.565,0.933,0.565}71.4 & \cellcolor[rgb]{0.753,0.753,0.753}76.2 & \cellcolor[rgb]{0.753,0.753,0.753}69.0 & \cellcolor[rgb]{0.753,0.753,0.753}69.3 & \cellcolor[rgb]{0.753,0.753,0.753}67.0 & \cellcolor[rgb]{0.804,0.498,0.196}64.4 \\
Claude\_3.5\_Sonnet & \cellcolor[rgb]{0.804,0.498,0.196}69.0 & \cellcolor[rgb]{0.753,0.753,0.753}73.7 & \cellcolor[rgb]{0.804,0.498,0.196}74.3 & \cellcolor[rgb]{0.565,0.933,0.565}72.4 & \cellcolor[rgb]{0.804,0.498,0.196}65.3 & \cellcolor[rgb]{0.565,0.933,0.565}67.8 & \cellcolor[rgb]{0.565,0.933,0.565}65.7 & \cellcolor[rgb]{0.565,0.933,0.565}63.8 \\
Qwen2\_72B & \cellcolor[rgb]{0.565,0.933,0.565}68.8 & \cellcolor[rgb]{0.867,0.627,0.867}70.8 & \cellcolor[rgb]{0.753,0.753,0.753}75.0 & \cellcolor[rgb]{0.804,0.498,0.196}74.6 & \cellcolor[rgb]{0.678,0.847,0.902}61.7 & \cellcolor[rgb]{0.804,0.498,0.196}68.2 & \cellcolor[rgb]{0.804,0.498,0.196}66.2 & \cellcolor[rgb]{0.753,0.753,0.753}64.7 \\
Llama\_3.2\_90B & \cellcolor[rgb]{0.678,0.847,0.902}65.9 & \cellcolor[rgb]{0.678,0.847,0.902}71.3 & \cellcolor[rgb]{0.678,0.847,0.902}70.2 & \cellcolor[rgb]{0.867,0.627,0.867}68.6 & \cellcolor[rgb]{0.565,0.933,0.565}64.6 & \cellcolor[rgb]{0.867,0.627,0.867}63.1 & \cellcolor[rgb]{0.678,0.847,0.902}62.9 & \cellcolor[rgb]{0.867,0.627,0.867}60.8 \\
Gemini\_1.5\_flash & \cellcolor[rgb]{0.867,0.627,0.867}65.7 & \cellcolor[rgb]{0.804,0.498,0.196}71.5 & \cellcolor[rgb]{0.867,0.627,0.867}70.2 & \cellcolor[rgb]{0.678,0.847,0.902}70.8 & \cellcolor[rgb]{0.867,0.627,0.867}60.2 & \cellcolor[rgb]{0.678,0.847,0.902}65.2 & \cellcolor[rgb]{0.867,0.627,0.867}61.1 & \cellcolor[rgb]{0.678,0.847,0.902}61.0 \\
\bottomrule
\end{tabular}
\caption{\textbf{The rankings of models differ strongly across the tested domains.} Model Accuracies across different generated datasets. The 'Overall' column represents the mean accuracy across all datasets. \textcolor[rgb]{1,0.843,0}{\rule{0.5cm}{0.5cm}} 1st place (Gold) \quad \textcolor[rgb]{0.753,0.753,0.753}{\rule{0.5cm}{0.5cm}} 2nd place (Silver) \quad \textcolor[rgb]{0.804,0.498,0.196}{\rule{0.5cm}{0.5cm}} 3rd place (Bronze) \quad \textcolor[rgb]{0.565,0.933,0.565}{\rule{0.5cm}{0.5cm}} 4th place \quad \textcolor[rgb]{0.678,0.847,0.902}{\rule{0.5cm}{0.5cm}} 5th place \quad \textcolor[rgb]{0.867,0.627,0.867}{\rule{0.5cm}{0.5cm}} 6th place. Only the top six models are shown. The 'Overall' column represents the mean accuracy across all datasets. Due to space constraints, results for additional models are provided in Appendix \Cref{tab:model_accuracies}. Note that Accuracy does not account for shared images between questions; this issue is addressed in~\Cref{fig:acc_per_global_fig}.}
\label{tab:model_accuracies_top7}
\end{table*}

\textbf{Metrics and rankings:} Choosing an adequate strategy for performance assessment is far from trivial and a research topic of its own~\cite{maier2024metrics,reinke2024understanding}. In this work, we were specifically interested in relative performance differences rather than in the specific ability of VLMs to serve a specific task. To obtain aggregated performance values across images, we define the \textbf{Accuracy\%($t$)} metric with a threshold \(t \in [0, 1]\).  
For each image \(i\) in a dataset \(D\), let \(Q_i\) denote the set of questions associated with that image.  
Let \(C_{i,q,m} \in \{0, 1\}\) indicate whether model \(m\) correctly answered question \(q\) for image \(i\) (1 for correct, 0 otherwise). The model \(m\) is considered to meet the threshold \(t\) on image \(i\) if the fraction of questions \(q\) in \(Q_i\) answered correctly by the model is at least \(t\). Formally, we define:
\begin{gather*}
\text{Accuracy\%}_m(t) = \\
\frac{1}{\lvert D\rvert}\,\sum_{i \in D}\,
    I\!\Bigl(\bigl(\tfrac{1}{\lvert Q_i\rvert}\,\sum_{q \in Q_i}C_{i,q,m}\bigr) \ge t\Bigr)
    \times 100
\end{gather*}
Here, \(I(\cdot)\) is an indicator function defined as:
\[
I(x \ge t)
 \;=\;
 \begin{cases}
  1, & \text{if } x \ge t,\\
  0, & \text{otherwise}.
 \end{cases}
\]
\paragraph{Explanation:}
\begin{itemize}
  \item \(\sum_{q \in Q_i} C_{i,q,m}\): Total number of correctly answered questions for image \(i\).
  \item \(\frac{1}{\lvert Q_i\rvert}\sum_{q \in Q_i} C_{i,q,m}\): Fraction of questions answered correctly for image \(i\).
  \item \(t \in [0,1]\): Desired minimum accuracy level assessed for each \(Q_i\).
\end{itemize}

%% file: sec/4_experiments_and_results.tex
\section{Experiments and Results}
The primary purpose of our experiments was to showcase the benefit of our task augmentation approach (sec. \ref{sec:4.1}). To assess the value of each task for VLM benchmarking, we related it to average model performance, resources needed to create the task, and corresponding human ambiguity (sec. \ref{sec:4.2}). Finally, we leveraged our concept and data to explore the capabilities of the most recent open and closed VLMs (sec. \ref{sec:4.3}.).

\subsection{Benefit of the proposed framework
\label{sec:4.1}
}
\Cref{fig:metadataaug} shows aggregated performance values for all models, separated by imaging domain. As the tasks and prompts were homogenized, the results clearly indicate that performance varies substantially across domains, supporting the hypothesis that in-domain validation is crucial for real-world translation. Note that this holds true despite the fact that we purposely chose domains that are relatively common (presumably captured in the model training) and closely related to one another.

Furthermore, as shown in~\cref{fig:fig3}a, the performance of models varies substantially across VLM tasks, suggesting that the tasks generated by our framework are diverse. The hardest tasks on average across domains are (1) T7.2 “Jigsaw Puzzle Completion”, (2), T1.2 “Object Counting”, (3), T7.1 “Rotated Jigsaw Puzzle Completion”, (4), T2.1 “Object Occlusion Detection”, and (5) T5.2 “Second Brightest Image Selection”. The easiest task on average was T1.3 “Additional Object Presence Detection” (see~\cref{fig:fig6_instance}).

\begin{figure*}[ht]
    \centering
    
    \begin{subfigure}{0.95\linewidth}
        \centering
        \includegraphics[width=0.95\linewidth]{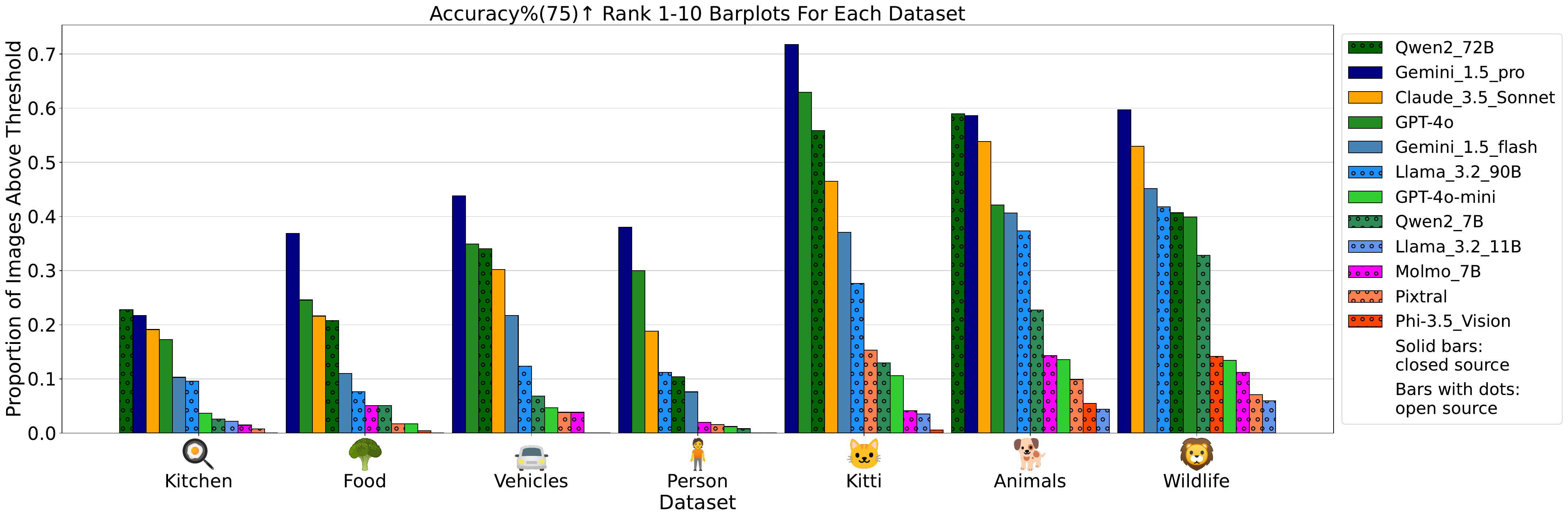}
        \caption{\textbf{The need for specific in-domain evaluation is demonstrated by the high performance variability across imaging domains.} 
        The performance of the overall best model Gemini 1.5 pro varies between domains from 22\% (Kitchen dataset) to 72\% (Kitti dataset). 
        For the displayed Accuracy\%(75), humans achieve an almost perfect score of 1 for all datasets (see Appendix). The top 10 models per dataset shown. 
        We display the full plots for all thresholds and models in \Cref{appendix_various_fixed_percentage_lines}.}
        \label{fig:acc_sub1}
    \end{subfigure}
    
    
    \begin{subfigure}{0.95\linewidth}
        \centering
        \includegraphics[width=0.95\linewidth]{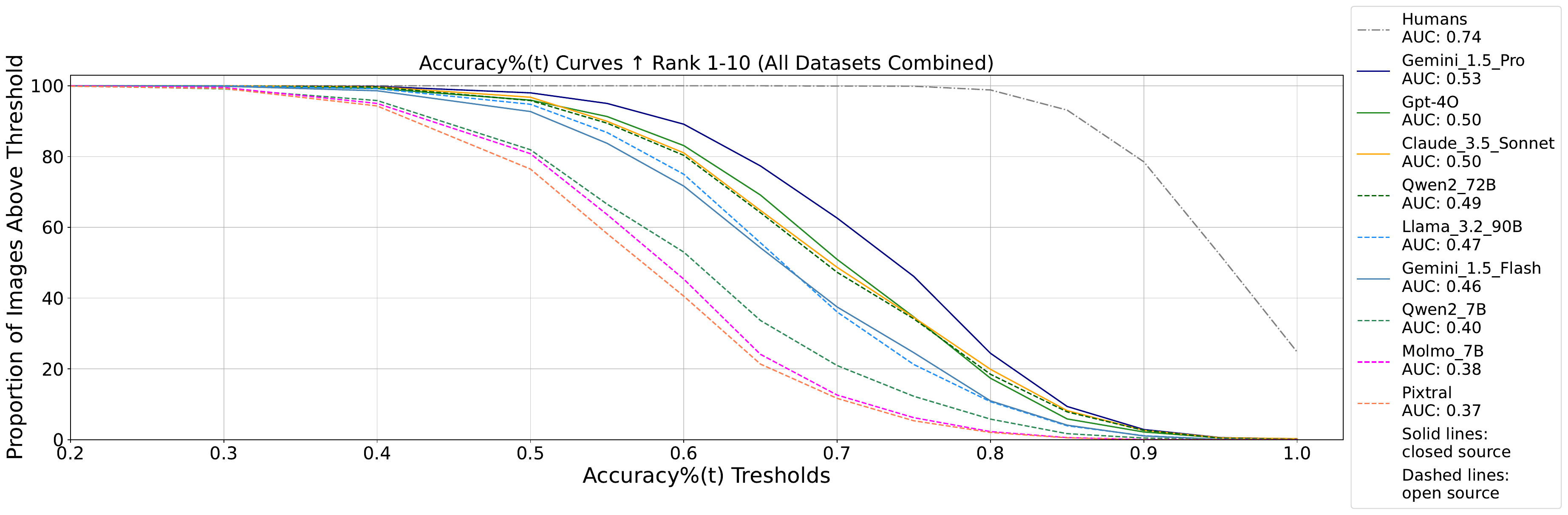}
        \caption{The Area under the Accuracy\%($t$) Curve serves as a \textbf{metric for comprehensive image understanding}. 
        With a maximum possible value of 1, higher values indicate better performance. 
        Notably, the current state-of-the-art model, Gemini\_1.5\_pro, achieves only 0.53, highlighting significant room for improvement. 
        Only the top 10 models are shown. 
        Separate curves for all 22 tested models and datasets are displayed in~\cref{appendix_full_auc_curves}.}
        \label{fig:full_ACC_perc_curves}
    \end{subfigure}
    
    \caption{\textbf{Performance varies across domains, highlighting the need for specialized in-domain evaluation; even the best models still lag behind human performance}. 
    The Accuracy\%($t$) metric represents the percentage of images for which at least a specified proportion of questions are correctly answered. 
    It can (a) be computed for specific thresholds or (b) be aggregated over multiple thresholds to remove dependence on a specific $t$. 
    The Area under the Accuracy\%($t$) Curve captures model performance in a single value, ranging from 0.37 to 0.53 for the top 10 models tested.}
    \label{fig:acc_per_global_fig}
\end{figure*}
\subsection{Human Ambiguity}
\label{sec:4.2}
As demonstrated in Appendix \ref{appendix_task_ranking_models_humans}, there is a high discrepancy in task rankings between humans and models. While the "Jigsaw Puzzle Completion” tasks ranked amongst the most challenging for the models, humans found "Object Occlusion Detection" and “Object Touching Detection” to be the most difficult.
 
From a resource perspective, tasks should be (1) hard to solve for models and (2) require as little human annotation as possible. This potential trade-off is captured in Appendix \autoref{appendix_fig6_instance}. It can be seen that many hard tasks, including the top four, can already be extracted from instance segmentations alone.

\subsection{Insights on current models}
\label{sec:4.3}
\Cref{fig:acc_per_global_fig} summarizes the performance of a model selection and reference baselines. Further detailed analysis, including all tested models, examples, and errors for each generated dataset are provided in the Appendix. The following insights can be extracted:

\textbf{Confirming common findings from the community:}
Our analysis reinforces several established patterns in the field. Closed models continue to demonstrate superior performance across tasks and domains, although open models have significantly narrowed this performance gap. In particular, Qwen2 72B stands out as the strongest performer among open models. The superiority of human evaluation remains evident, with human raters achieving near-perfect performance on most tasks, though they notably struggle with specific challenges such as counting, occlusion, and direction-related tasks—counting being particularly problematic. Regarding model scaling, larger variants typically show better performance, with some notable exceptions such as Molmo 7B outperforming Pixtral 12B.

\textbf{Interesting new findings:}
The need for specific in-domain evaluation is highlighted by the high performance variability across imaging domains for the same perception tasks, see \Cref{tab:model_accuracies_top7} and~\Cref{fig:metadataaug}. The overall best model, Gemini 1.5 Pro, varies between domains from 22\% (Kitchen dataset) to 72\% (Kitti dataset). Qwen2 72B slightly surpasses Gemini 1.5 Pro on the kitchen and animals datasets but ranks only fifth on the person dataset. Additional insights emerge from model comparisons, with Qwen2 7B consistently outperforming Molmo 7B across most datasets, and Gemini Flash 1.5 showing superior Point Depth Comparison capabilities over Gemini Pro. These results indicate that our newly introduced metric, Accuracy\%($t$), can effectively capture model performance in a single value.

%% file: sec/5_discussion.tex
\section{Discussion}

This paper contributes to the advancement of VLM benchmarking in three ways: 

\textbf{1) Framework for resource-efficient and domain-specific benchmarking:} We showed that task augmentation, using instance segmentation as the root task, enables the generation of a diverse set of VLM tasks and could thus evolve as a core method for resource-efficient domain-specific VLM benchmarking. The insights gained on the varying difficulty of presented VLM tasks will further guide the design of future benchmarks. The framework can be easily applied to other domains, even with a small number of images. The computational and monetary costs for each generated dataset are minimal and displayed in~\cref{appendix_comp_and_mon_cost}. 

\textbf{2) Seven new openly available datasets:} Our seven new datasets will help assess generalist capabilities of future VLMs. Furthermore, we release the six human annotations per task (totaling 162,946 annotations) to assist researchers working on human annotations. 

\textbf{3) New insights:} The insights on current capabilities of closed and open VLMs highlight the narrowing gap between closed and open models. Most importantly, we showcased the need for domain-specific validation.

Core strengths of our contribution include the broad applicability of our concept, the open dataset and benchmark contribution, and the wide range of state-of-the-art closed and open models investigated here.

As an implicit contribution, we introduced the new metric Accuracy\%($t$), which offers several key strengths. First, it captures model performance in a single very intuitive value. The metric is extendable with additional tasks, allowing for gradually increasing difficulty, and can be adapted to evaluate domain-specific tasks effectively. It is worth mentioning, however, that the specific properties of the metric require further analyses~\cite{reinke2024understanding}. For example, some questions require specific image conditions, such as the presence of multiple objects for comparison. This can result in a varying number of questions per image, which, in turn, has an influence on the metric. Furthermore, tasks are treated equally without any weighting, which may overlook differences in task difficulty or importance. Users can, however, easily modify the weighting scheme to better reflect their specific evaluation priorities.

A limitation of our work is model family dependence, as many models come from closely related families, which may hinder statistical analysis. For closed-source models, specific information about training and data is often unavailable, creating transparency issues. We provide further statistical analysis, such as ranking variability in~\cref{app_sec_stats}. Model performance showed small variations with prompt phrasing, which we mitigated through iterative testing for consistency.  Additionally, our human annotations were performed by professional annotators, which may introduce ambiguity since annotators aim to complete tasks quickly.

Future work should focus on expanding the number of tasks generated, further enhancing the diversity and comprehensiveness of VLM benchmarks. Additionally, our method can be adapted to different domains with domain-specific questions or scaled up to support continuous extension, providing a versatile approach for evaluating models across diverse applications.

\subsubsection*{Code / Datasets / Human Annotations}
Code, datasets, and annotations will be made available.

\begin{figure}
  \centering
  \includegraphics[width=0.95\linewidth]{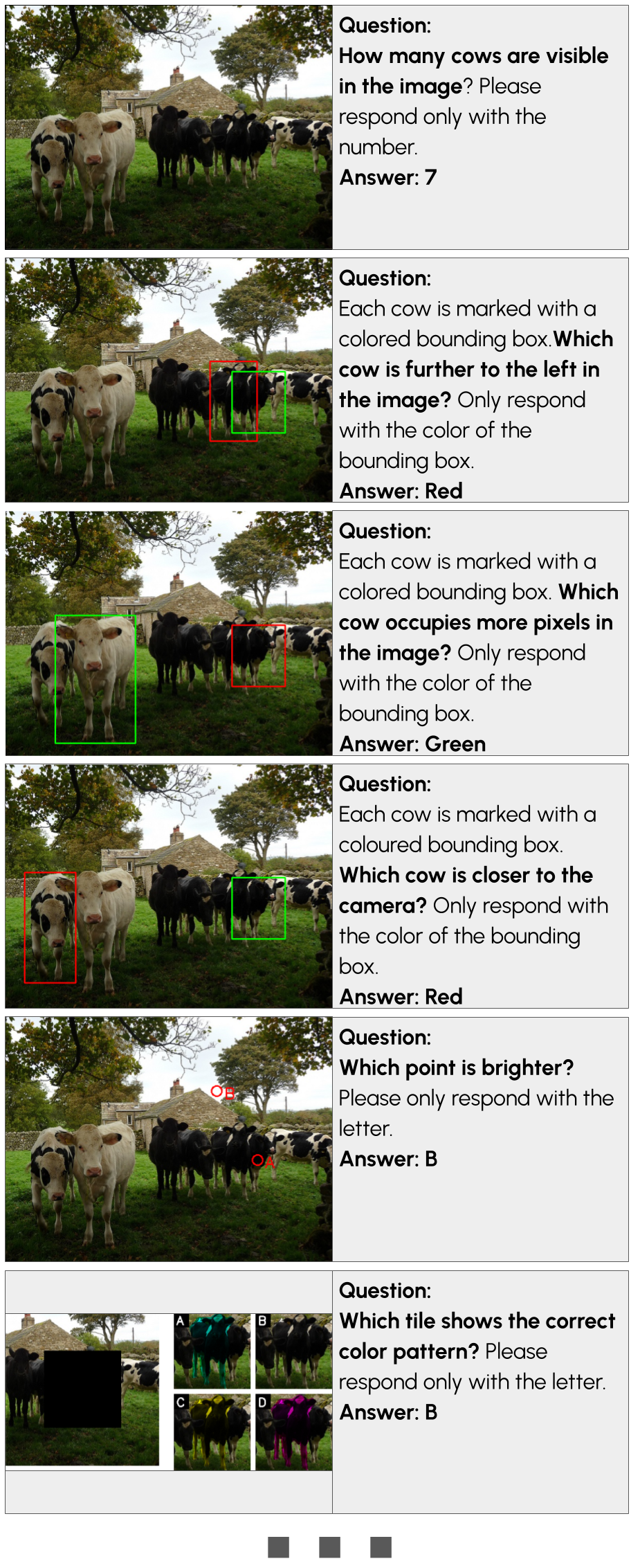}
   \caption{\textbf{Our framework yields a diverse set of tasks.} Exemplary tasks that were generated with the framework for a given image. A broad range of examples and errors for each generated dataset is provided in~\cref{domain_sets_appendix}.}
   \label{fig:example_cow}
\end{figure}

%% file: sec/X_suppl.tex
\addtocontents{toc}{\protect\setcounter{tocdepth}{3}}  

\setcounter{page}{1}
\onecolumn  

\renewcommand{\numberline}[1]{#1\hspace{1em}}  

\hypertarget{toc}{}

We provide further detail on i) the ressource-efficient framework, ii) the seven generated domain-specific datasets, and iii) the benchmarking insights and model evaluations.
\let\clearpage\relax
\tableofcontents  


\pagestyle{fancy}  
\fancyhf{}  
\fancyfoot[C]{\textcolor{red}{\hyperlink{toc}{Back to Appendix Table of Contents}} \textbar\ Page \thepage}

\renewcommand{\headrulewidth}{0pt}  
\renewcommand{\footrulewidth}{0pt}  

\newpage

\section{RESOURCE-EFFICIENT VLM BENCHMARKING FRAMEWORK}
\subsection{CVPR 2024 Paper Analysis}
\label{appendix_cvpr_paper_analysis}

\begin{table}[h]
\small
\centering
\begin{tabular}{|l|l|l|p{3cm}|} \hline 
\textbf{CVPR 2024 }&   & Manually verified
papers&Agreement LLMs and Human verification\\ \hline 
Total number of papers & 2,708  & --&--\\ \hline 
With new or modified dataset: & 397  & 40 (10\%)&1\\ \hline 
Without new or modified dataset: & 2,311  & 50 (2\%)&1\\ \hline
\end{tabular}
\caption{\textbf{A notable portion of CVPR 2024 papers contribute new or modified datasets, highlighting a rising trend in dataset-focused research}. CVPR 2024 paper analysis summary.}
\end{table}
We analyzed all papers from CVPR 2024 using three different large language models (LLMs). If the majority of models indicated that a paper introduced a new or modified dataset, we tagged it accordingly. This process identified 397 publications proposing a new or modified dataset. To validate the accuracy of the tagging, we randomly selected 10\% of these flagged papers for a human review. All human-verified publications were confirmed to propose a new dataset.

\newpage

\subsection{Example Of Human Generated Metadata}
\label{appendix_example_quality_match}
\begin{figure}[h]
    \centering
    \includegraphics[width=0.9\linewidth]{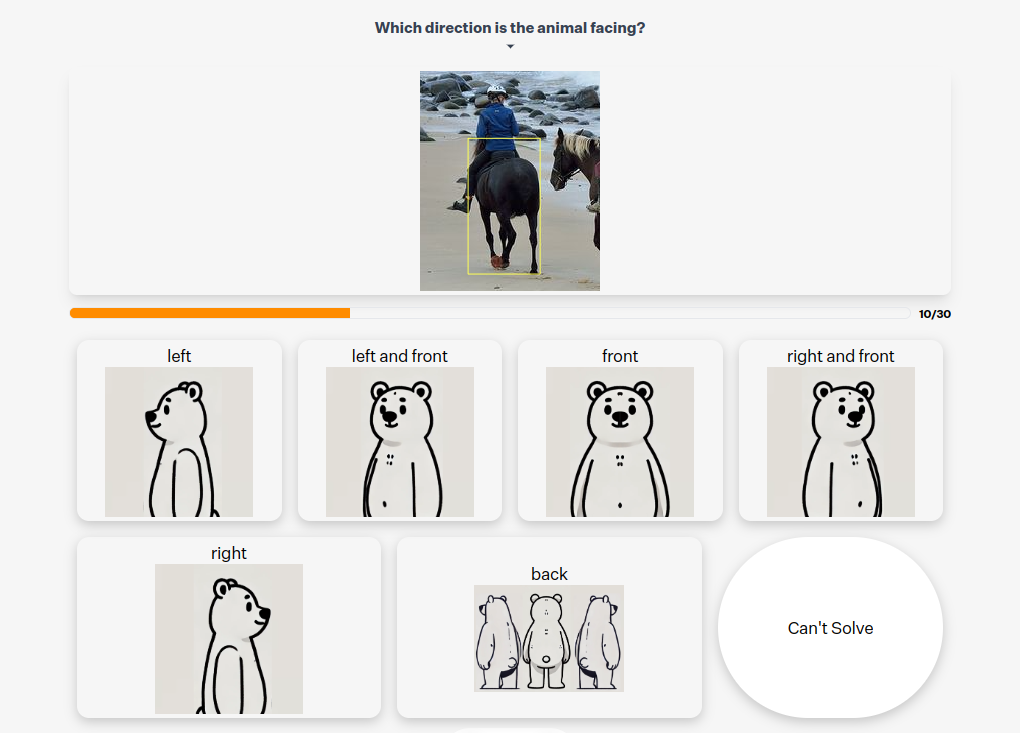}
    \caption{\textbf{Example of human-generated metadata enriching object annotations in initialization tasks.} These annotations demonstrate the process of enriching objects with human-provided metadata during the initial setup phase.}
    \label{fig:quality_match_example}
\end{figure}

\newpage

\subsection{Metadata Sources}
\begin{table}[h]
\centering

\begin{tabular}{|p{5.0cm} |l|} \hline 
\multicolumn{2}{|l|}{\textbf{Human Raters}} \\ \hline
\textbf{Attribute} & \textbf{Description} \\ \hline 
Occluded & Object occluded or fully visible (other object in front) \\ \hline 
Truncated & Object truncated or fully visible (edge of image) \\ \hline 
Direction & Direction the object is facing \\ \hline 
\multicolumn{2}{|l|}{\textbf{Existing Annotations}} \\ \hline
\textbf{Attribute} & \textbf{Description} \\ \hline 
relative\_size & Relative size compared to image size \\ \hline 
bbox\_touches\_bbox & Bounding box touching another bounding box \\ \hline 
segmask\_touches\_segmask & Segmentation mask touching another segmentation mask \\ \hline 
segmask\_touches\_segmask\_with & Specific segmentation masks touching each other \\ \hline 
segmentation\_area & Area covered by segmentation \\ \hline 
brightness\_score & Brightness score \\ \hline 
michelson\_contrast\_score & Michelson contrast score \\ \hline 
bbox\_x\_min, bbox\_y\_min,\newline  bbox\_x\_max, bbox\_y\_max & Bounding box coordinates \\ \hline 
class\_name & Class name of the object \\ \hline 
\multicolumn{2}{|l|}{\textbf{Model Generated}} \\ \hline
\textbf{Attribute} & \textbf{Description} \\ \hline 
average\_depth & Average depth of the object \\ \hline 
top\_95\_depth & Depth of the top 95\% portion of the object \\ \hline 
bottom\_5\_depth & Depth of the bottom 5\% portion of the object \\ \hline

\end{tabular}
\caption{\textbf{Overview of metadata sources used for enriching instance segmentation datasets.} Metadata was created from existing annotations, specialized models, or manually annotated by human raters.}
\label{appendix_three_sources_of_metadata}
\end{table}

\newpage

\subsection{VLM Tasks Overview}
\label{appendix_overviewvlmtasks}
\begin{figure}[h]
    \centering
\includegraphics[width=\linewidth]{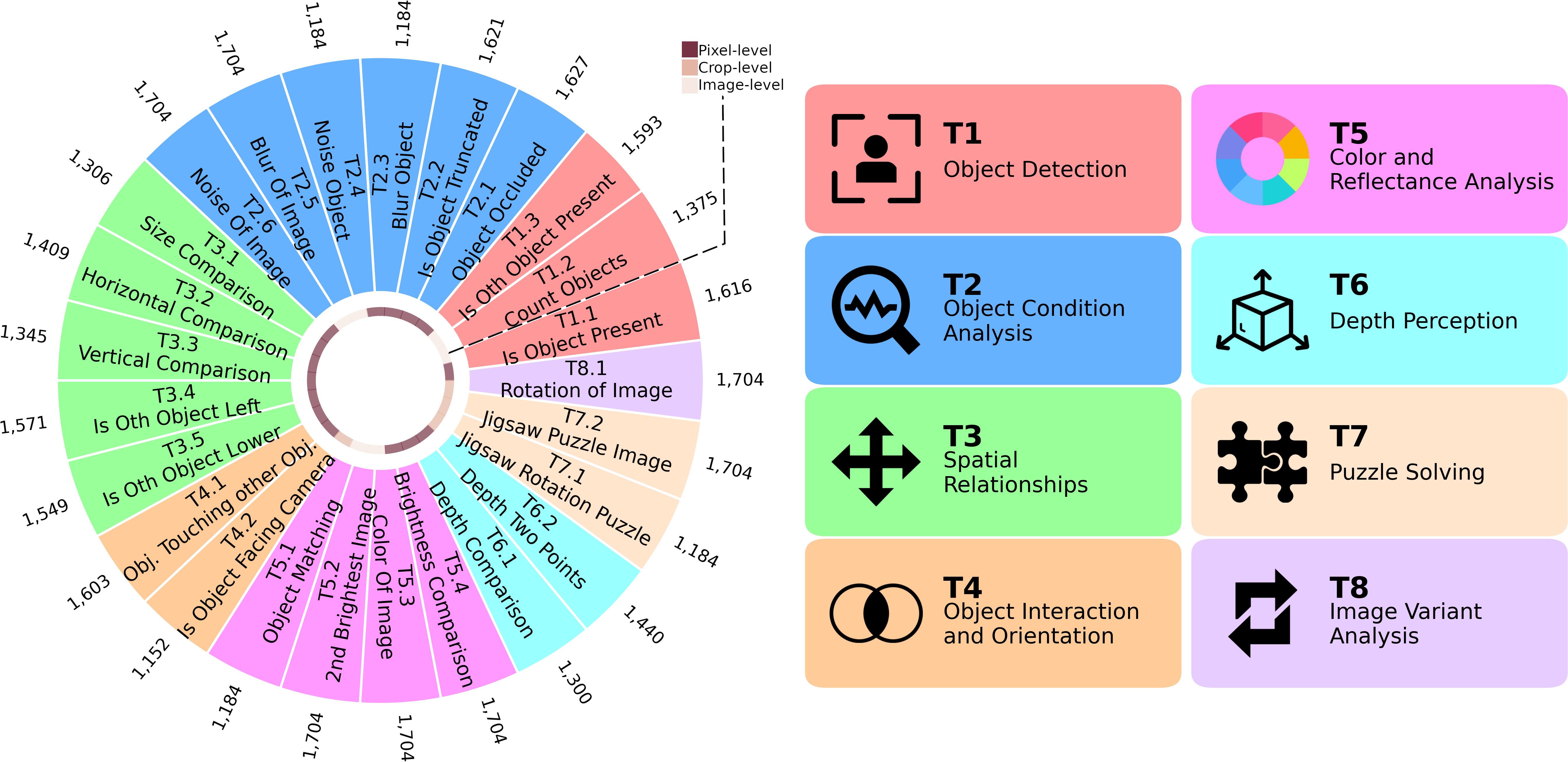}
\caption{\textbf{Our framework yields a diverse set of tasks.} Based on a single image with instance segmentations, our framework enables the  generation of 25 tasks from eight different vision-language categories, ranging from pixel-level to image-level perception.}
    \label{fig:fig3_app}
\end{figure}

Here we present the VLM tasks overview and its corresponding meta categories in~\autoref{fig:fig3_app}. Further information on each task is provided in~\autoref{tab:vlm_benchmark} on the next page.

\begin{table}[t]
\centering
\small

\begin{tabular}{|p{0.05\textwidth}|p{0.15\textwidth}|p{0.5\textwidth}|p{0.15\textwidth}|}
\hline
\textbf{ID} & \textbf{Task Name} & \textbf{Task Description} & \textbf{Answer Type} \\
\hline
T1.1 & Is Object Present & Determines whether a specified object is present in the image. & Binary \\
\hline
T1.2 & Count Objects & Determines the number of objects in the image & Count \\
\hline
T1.3 & Is Oth Object Present & Determines whether or not there is more than one object in the image & Binary \\
\hline
T2.1 & Is Object Occluded & Determines if the specified object is partially or fully occluded. & Quiz (A/B/C/D) \\
\hline
T2.2 & Is Object Truncated & Determines if the specified object is truncated in the image frame. & Binary \\
\hline
T2.3 & Blur Object & Determines whether an object is blurred & Quiz (A/B/C/D) \\
\hline
T2.4 & Noise Object & Determines whether an object contains noise & Quiz (A/B/C/D) \\
\hline
T2.5 & Blur Of Image & Determines which image variant is least blurred & Quiz (A/B/C/D) \\
\hline
T2.6 & Noise Of Image & Determines which image variant is not corrupted & Quiz (A/B/C/D) \\
\hline
T3.1 & Size Comparison & Determines which of two objects is larger & Color \\
\hline
T3.2 & Horizontal Comparison & Determines which object is further to the left of the image & Color \\
\hline
T3.3 & Vertical Comparison & Determines which object is further to the bottom of the image & Color \\
\hline
T3.4 & Is Oth Object Left & Determines whether there is another image further to the left of an object & Binary \\
\hline
T3.5 & Is Oth Object Lower & Determines whether there is another image further to the bottom of an object & Binary \\
\hline
T4.1 & Is Object Touching other Object & Determines if two objects are touching each other & Binary \\
\hline
T4.2 & Is Object Facing Camera & Determines if the object is facing the camera & Quiz (A/B/C/D) \\
\hline
T5.1 & Color Object Matching & Determines which of four tiles show the correct color for the given image & Quiz (A/B/C/D) \\
\hline
T5.2 & 2nd Brightest Image & Determines which of the images is the 2nd brightest image & Quiz (A/B/C/D) \\
\hline
T5.3 & Color Of Image & Determines which image variant is not corrupted & Quiz (A/B/C/D) \\
\hline
T5.4 & Brightness Comparison of Two Points & Determines which of two points is brighter & Binary \\
\hline
T6.1 & Depth Comparison & Determines which of two objects is closer to the camera & Color \\
\hline
T6.2 & Depth Two Points Image & Determines which point is closer & Binary \\
\hline
T7.1 & Jigsaw rotation Puzzle & Determines which of four rotated tiles fits best into a cut out area of the image & Quiz (A/B/C/D) \\
\hline
T7.2 & Jigsaw Puzzle Image & Determines which of four tiles fits best into a cut out area of the image & Quiz (A/B/C/D) \\
\hline
T8.1 & Rotation Of Image & Determines which image variant is not rotated & Quiz (A/B/C/D) \\
\hline
\end{tabular}
\caption{\textbf{Overview of VLM Benchmark Tasks generated with the framework.} We provide a small task description and answer type for each generated task. Examples across datasets are displayed in~\autoref{sec:app_examples_for_each_dataset}. }
\label{tab:vlm_benchmark}
\end{table}

\newpage

\newpage

\subsection{Task Augmentation Methods Comparison}
\label{appendix_task_augmentation_comparison}

\begin{table}[H]
\centering
\footnotesize
\renewcommand{\arraystretch}{1.2} 
\setlength{\tabcolsep}{4pt} 
\begin{tabular}{|P{2cm}|P{1.2cm}|P{2.2cm}|P{1.8cm}|P{2cm}|P{2.2cm}|P{2.2cm}|P{1.3cm}|}
\hline
\textbf{Metric} & \textbf{CLEVR} & \textbf{Task Me Anything} & \textbf{Taskonomy} & \textbf{Wang 2023} & \textbf{JourneyBench} & \textbf{ProVision} & \textbf{Ours} \\ \hline
\textbf & \cite{johnson2017clevr} & \cite{zhang2024task} & \cite{zamir2018taskonomy} & \cite{wang2023instruct4v} & \cite{wang2024journeybench} & \cite{zhang2024provision} & \textbf{Ours} \\ \hline
Real images/objects & \includegraphics[width=0.02\textwidth]{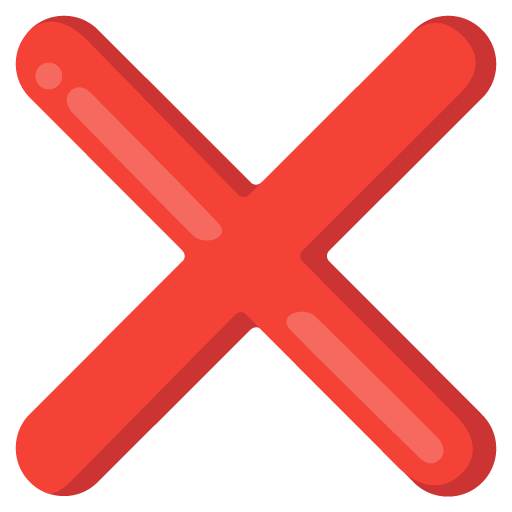} & \includegraphics[width=0.02\textwidth]{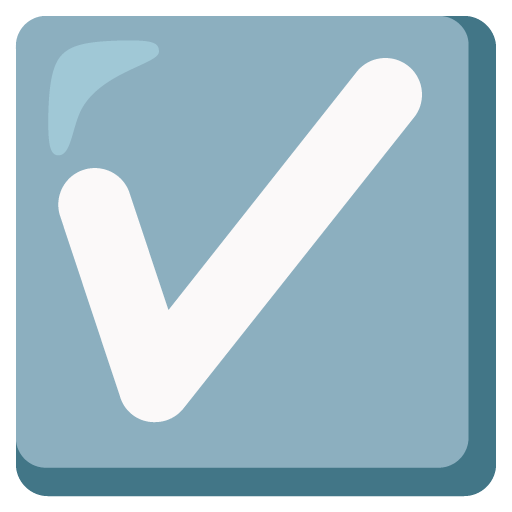} (partly, needs scene graph) & \includegraphics[width=0.02\textwidth]{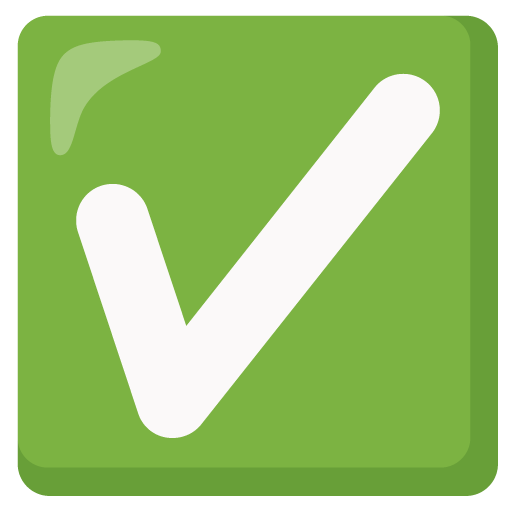} & \includegraphics[width=0.02\textwidth]{figures/icons_iclr/icon_hook_green.png} & \includegraphics[width=0.02\textwidth]{figures/icons_iclr/icon_cross_red.png} & \includegraphics[width=0.02\textwidth]{figures/icons_iclr/icon_hook_green.png} & \includegraphics[width=0.02\textwidth]{figures/icons_iclr/icon_hook_green.png} \\ \hline
Diversity core perception tasks & \includegraphics[width=0.02\textwidth]{figures/icons_iclr/icon_hook_green.png} (subjective) & \includegraphics[width=0.02\textwidth]{figures/icons_iclr/icon_hook_green.png} & \includegraphics[width=0.02\textwidth]{figures/icons_iclr/icon_hook_green.png} & \includegraphics[width=0.02\textwidth]{figures/icons_iclr/icon_hook_green.png} & \includegraphics[width=0.02\textwidth]{figures/icons_iclr/icon_hook_green.png} & \includegraphics[width=0.02\textwidth]{figures/icons_iclr/icon_hook_green.png} & \includegraphics[width=0.02\textwidth]{figures/icons_iclr/icon_hook_green.png} \\ \hline
Focus on resource efficiency & Synthetic data & Strong synthetic data focus and flexible & Not relevant, no new tasks/data can be added & \includegraphics[width=0.02\textwidth]{figures/icons_iclr/icon_cross_red.png} & \includegraphics[width=0.02\textwidth]{figures/icons_iclr/icon_cross_red.png} (2,200 hours of human annotation) & \includegraphics[width=0.02\textwidth]{figures/icons_iclr/icon_hook_green.png} & \includegraphics[width=0.02\textwidth]{figures/icons_iclr/icon_hook_green.png} \\ \hline
Enables others to use their own data and benchmark / Easily extendable & \includegraphics[width=0.02\textwidth]{figures/icons_iclr/icon_hook_green.png} & \includegraphics[width=0.02\textwidth]{figures/icons_iclr/icon_hook_green.png} & \includegraphics[width=0.02\textwidth]{figures/icons_iclr/icon_cross_red.png} & \includegraphics[width=0.02\textwidth]{figures/icons_iclr/icon_hook_green.png} (in theory possible, but no code) & \includegraphics[width=0.02\textwidth]{figures/icons_iclr/icon_hook_green.png} & \includegraphics[width=0.02\textwidth]{figures/icons_iclr/icon_hook_green.png} & \includegraphics[width=0.02\textwidth]{figures/icons_iclr/icon_hook_green.png} \\ \hline
Not reliant on generative models & \includegraphics[width=0.02\textwidth]{figures/icons_iclr/icon_hook_green.png} & \includegraphics[width=0.02\textwidth]{figures/icons_iclr/icon_hook_green.png} & \includegraphics[width=0.02\textwidth]{figures/icons_iclr/icon_hook_green.png} & \includegraphics[width=0.02\textwidth]{figures/icons_iclr/icon_cross_red.png} & \includegraphics[width=0.02\textwidth]{figures/icons_iclr/icon_cross_red.png} & \includegraphics[width=0.02\textwidth]{figures/icons_iclr/icon_cross_red.png} & \includegraphics[width=0.02\textwidth]{figures/icons_iclr/icon_hook_green.png} \\ \hline
Object-centric & \includegraphics[width=0.02\textwidth]{figures/icons_iclr/icon_hook_green.png} & \includegraphics[width=0.02\textwidth]{figures/icons_iclr/icon_hook_green.png} & \includegraphics[width=0.02\textwidth]{figures/icons_iclr/icon_cross_red.png} & \includegraphics[width=0.02\textwidth]{figures/icons_iclr/icon_hook_green.png} & \includegraphics[width=0.02\textwidth]{figures/icons_iclr/icon_cross_red.png} (hard to say) & \includegraphics[width=0.02\textwidth]{figures/icons_iclr/icon_hook_green.png} & \includegraphics[width=0.02\textwidth]{figures/icons_iclr/icon_hook_green.png} \\ \hline
Validated across multiple visual content domains & \includegraphics[width=0.02\textwidth]{figures/icons_iclr/icon_cross_red.png} & \includegraphics[width=0.02\textwidth]{figures/icons_iclr/icon_cross_red.png} & \includegraphics[width=0.02\textwidth]{figures/icons_iclr/icon_cross_red.png} & \includegraphics[width=0.02\textwidth]{figures/icons_iclr/icon_hook_green.png} & \includegraphics[width=0.02\textwidth]{figures/icons_iclr/icon_hook_green.png} & \includegraphics[width=0.02\textwidth]{figures/icons_iclr/icon_hook_green.png} & \includegraphics[width=0.02\textwidth]{figures/icons_iclr/icon_hook_green.png} \\ \hline
Human Ambiguity Scores & \includegraphics[width=0.02\textwidth]{figures/icons_iclr/icon_cross_red.png} & \includegraphics[width=0.02\textwidth]{figures/icons_iclr/icon_cross_red.png} & \includegraphics[width=0.02\textwidth]{figures/icons_iclr/icon_cross_red.png} & \includegraphics[width=0.02\textwidth]{figures/icons_iclr/icon_cross_red.png} & \includegraphics[width=0.02\textwidth]{figures/icons_iclr/icon_cross_red.png} & \includegraphics[width=0.02\textwidth]{figures/icons_iclr/icon_cross_red.png} & \includegraphics[width=0.02\textwidth]{figures/icons_iclr/icon_hook_green.png} \\ \hline
Easily scalable & \includegraphics[width=0.02\textwidth]{figures/icons_iclr/icon_hook_green.png} & \includegraphics[width=0.02\textwidth]{figures/icons_iclr/icon_hook_green.png} & \includegraphics[width=0.02\textwidth]{figures/icons_iclr/icon_cross_red.png} & \includegraphics[width=0.02\textwidth]{figures/icons_iclr/icon_hook_green.png} & \includegraphics[width=0.02\textwidth]{figures/icons_iclr/icon_hook_green.png} & \includegraphics[width=0.02\textwidth]{figures/icons_iclr/icon_hook_green.png} & \includegraphics[width=0.02\textwidth]{figures/icons_iclr/icon_hook_green.png}\\ \hline
Task creation code public & \includegraphics[width=0.02\textwidth]{figures/icons_iclr/icon_hook_green.png} & \includegraphics[width=0.02\textwidth]{figures/icons_iclr/icon_hook_green.png} & \includegraphics[width=0.02\textwidth]{figures/icons_iclr/icon_cross_red.png} & \includegraphics[width=0.02\textwidth]{figures/icons_iclr/icon_cross_red.png} & \includegraphics[width=0.02\textwidth]{figures/icons_iclr/icon_hook_green.png}  & \includegraphics[width=0.02\textwidth]{figures/icons_iclr/icon_hook_green.png} & \includegraphics[width=0.02\textwidth]{figures/icons_iclr/icon_hook_green.png} \\ \hline
Evaluated on SOTA VLMs & \includegraphics[width=0.02\textwidth]{figures/icons_iclr/icon_cross_red.png} & \includegraphics[width=0.02\textwidth]{figures/icons_iclr/icon_hook_grey.png}(limited \# of proprietary models) & \includegraphics[width=0.02\textwidth]{figures/icons_iclr/icon_cross_red.png} & \includegraphics[width=0.02\textwidth]{figures/icons_iclr/icon_hook_grey.png} (but SoTA* 2023) & \includegraphics[width=0.02\textwidth]{figures/icons_iclr/icon_hook_grey.png}(limited \# of proprietary models) & \includegraphics[width=0.02\textwidth]{figures/icons_iclr/icon_cross_red.png} &\includegraphics[width=0.02\textwidth]{figures/icons_iclr/icon_hook_green.png} \\ \hline
\end{tabular}
\caption{\textbf{Our framework uniquely combines comprehensive evaluation capabilities with methodological advantages.} Systematic comparison of task augmentation approaches across key metrics, highlighting distinct features in VLM evaluation, programmatic task generation, and efficiency measures relative to existing frameworks.}
\label{tab:benchmark_context_to_related_word}
\end{table}

Our work is positioned within the broader context of research on large VLMs, programmatic task generation and task augmentation. A comparison to other relevant work~\cite{johnson2017clevr, zhang2024task,zamir2018taskonomy,wang2023instruct4v, wang2024journeybench} is provided in~\autoref{tab:benchmark_context_to_related_word}. 

We deliberately excluded large datasets primarily designed for task generation, such as Visual Genome~\cite{krishna2017genome} and the Open Images Dataset~\cite{kuznetsova2020v4}, as these datasets are not suitable for directly using their own images for evaluation. Instead, our focus is on empowering domain-specific users to efficiently test a wide range of capabilities with a small number of real-world images containing relevant objects, while operating under limited resources. For instance, the human annotations required to generate the tasks in our framework can be achieved at a minimal cost of approximately \$27 USD (average across all seven datasets). Our approach is complementary to existing research, offering a lightweight and accessible alternative for specific use cases. Thus enabling researchers and practitioners to iterate more quickly and effectively in the right direction.

\newpage

\subsection{Computational and Monetary Cost}
\label{appendix_comp_and_mon_cost}

\begin{table}[H]
\centering
\footnotesize
\renewcommand{\arraystretch}{1.2}
\setlength{\tabcolsep}{4pt}
\begin{tabular}{|l|c|c|c|c|c|}
\hline
\multirow{2}{*}{Domain} & \multirow{2}{*}{Icon} & \multicolumn{3}{c|}{Metadata Enrichment} & \multirow{2}{*}{\begin{tabular}[c]{@{}c@{}}Perception\\Task\\Generation\\(in seconds)\end{tabular}} \\
\cline{3-5}
& & \begin{tabular}[c]{@{}c@{}}Pre-defined\\Heuristics\\(in seconds)\end{tabular} & \begin{tabular}[c]{@{}c@{}}ML Model\\(in seconds)\end{tabular} & \begin{tabular}[c]{@{}c@{}}Human Annotations\\for task generation\end{tabular} & \\
\hline
Wildlife & \includegraphics[width=0.4cm]{figures/icons_iclr/wildlife.png} & 15 & 50 & \multirow{7}{*}{\begin{tabular}[c]{@{}c@{}}$\sim$ 27\$ USD\\on average\\per dataset\\(2-day average\\turnover)\end{tabular}} & 124 \\
Persons & \includegraphics[width=0.4cm]{figures/icons_iclr/person.png} & 354 & 496 & & 160 \\
Vehicles & \includegraphics[width=0.4cm]{figures/icons_iclr/vehicles.png} & 73 & 211 & & 129 \\
Animals & \includegraphics[width=0.4cm]{figures/icons_iclr/animals.png} & 41 & 138 & & 129 \\
Kitchen & \includegraphics[width=0.4cm]{figures/icons_iclr/kitchen.png} & 48 & 156 & & 161 \\
Food & \includegraphics[width=0.4cm]{figures/icons_iclr/food.png} & 295 & 117 & & 144 \\
Kitti & \includegraphics[width=0.4cm]{figures/icons_iclr/kitti.png} & 23 & 108 & & 146 \\
\hline
\end{tabular}
\caption{\textbf{Our framework is resource-efficient in both computation and cost.} Displayed are the timings (in seconds) for metadata enrichment and question generation across domains, demonstrating the scalability and efficiency of our approach. Annotation costs averaged \$27 per dataset, calculated based on the number of objects and manually added metadata points. The turnover time for task annotations was two days. All computations were performed on an RTX 3090 GPU and Ryzen 9 5900X CPU.}
\label{table:metadata_metrics}
\end{table}

\newpage

\section{DOMAIN-SPECIFIC DATASETS}
\label{domain_sets_appendix}
\includepdf[pages=1-2,pagecommand=\subsection{Animals Dataset Examples}
\label{sec:app_examples_for_each_dataset},nup = 2x1, offset=0 0, frame=false, scale=0.75]{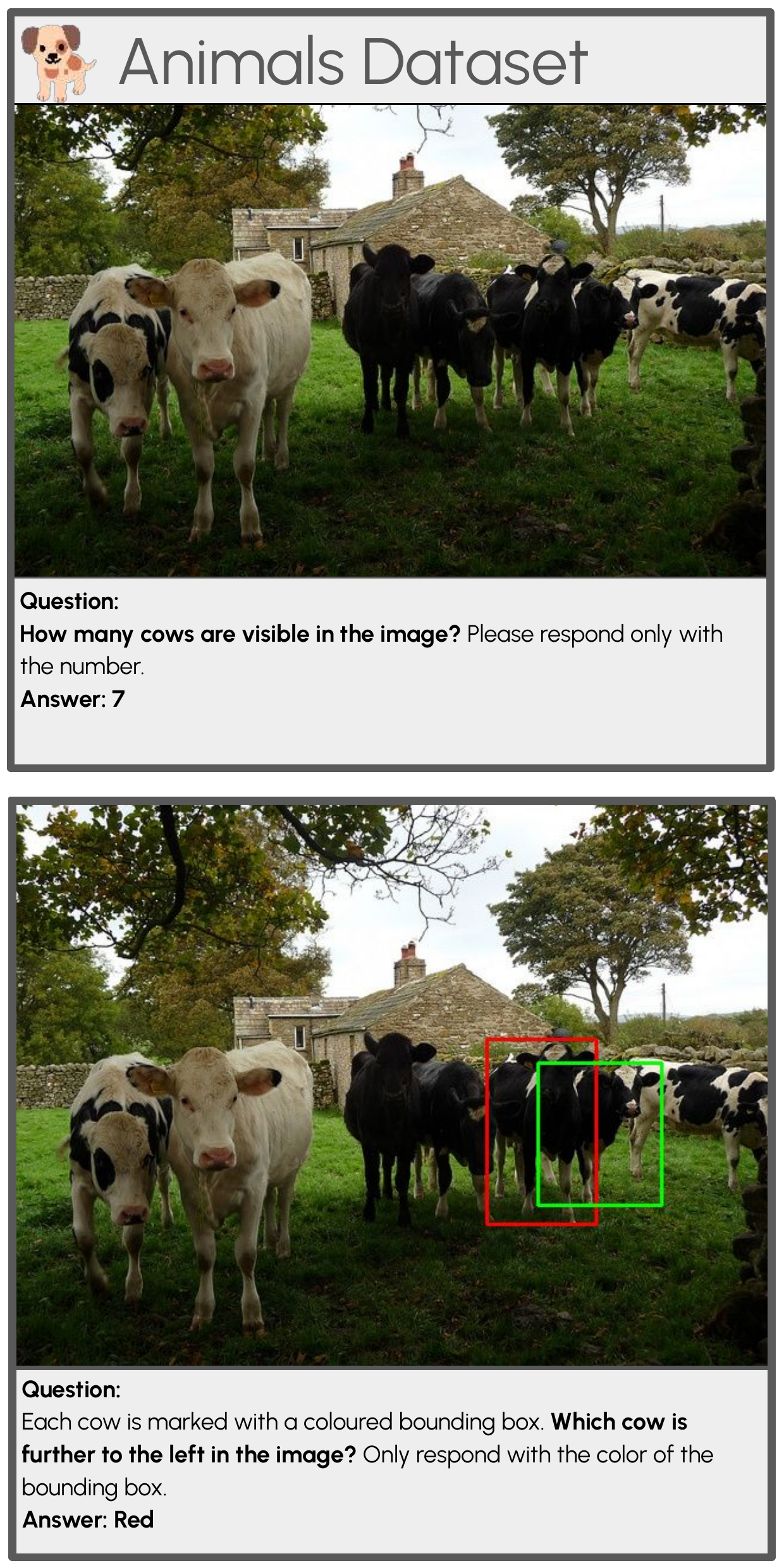}
\includepdf[pages=3-4,pagecommand= \thispagestyle{fancy},nup = 2x1, offset=0 0, frame=false, scale=0.75]{figures/appendix/cvpr_appendix_examples_7_datasets.pdf}

\includepdf[pages=5-6,pagecommand=\subsection{Person Dataset Examples\label{sec_sub:app_person}},nup = 2x1, offset=0 0, frame=false, scale=0.75]{figures/appendix/cvpr_appendix_examples_7_datasets.pdf}
\includepdf[pages=7-8,pagecommand=\thispagestyle{fancy}, nup = 2x1, offset=0 0, frame=false, scale=0.75]{figures/appendix/cvpr_appendix_examples_7_datasets.pdf}

\includepdf[pages=9-10,pagecommand=\subsection{Food Dataset Examples\label{sec_sub:app_food}},nup = 2x1, offset=0 0, frame=false, scale=0.75]{figures/appendix/cvpr_appendix_examples_7_datasets.pdf}
\includepdf[pages=11-12,pagecommand=\thispagestyle{fancy}, nup = 2x1, offset=0 0, frame=false, scale=0.75]{figures/appendix/cvpr_appendix_examples_7_datasets.pdf}

\includepdf[pages=13-14,pagecommand=\subsection{Vehicles Dataset Examples\label{sec_sub:app_vehicles}},nup = 2x1, offset=0 0, frame=false, scale=0.75]{figures/appendix/cvpr_appendix_examples_7_datasets.pdf}
\includepdf[pages=15-16,pagecommand=\thispagestyle{fancy}, nup = 2x1, offset=0 0, frame=false, scale=0.75]{figures/appendix/cvpr_appendix_examples_7_datasets.pdf}

\includepdf[pages=17-18,pagecommand=\subsection{Wildlife Dataset Examples\label{sec_sub:app_wildlife}},nup = 2x1, offset=0 0, frame=false, scale=0.75]{figures/appendix/cvpr_appendix_examples_7_datasets.pdf}
\includepdf[pages=19-20,pagecommand=\thispagestyle{fancy},nup = 2x1, offset=0 0, frame=false, scale=0.75]{figures/appendix/cvpr_appendix_examples_7_datasets.pdf}

\includepdf[pages=21-22,pagecommand=\subsection{Kitchen Dataset Examples\label{sec_sub:app_kitchen}},nup = 2x1, offset=0 0, frame=false, scale=0.75]{figures/appendix/cvpr_appendix_examples_7_datasets.pdf}
\includepdf[pages=23-24,pagecommand=\thispagestyle{fancy}, nup = 2x1, offset=0 0, frame=false, scale=0.75]{figures/appendix/cvpr_appendix_examples_7_datasets.pdf}

\includepdf[pages=25-26,pagecommand=\subsection{Kitti Dataset Examples\label{sec_sub:app_kitti}}
,nup = 2x1, offset=0 0, frame=false, scale=0.75]{figures/appendix/cvpr_appendix_examples_7_datasets.pdf}
\includepdf[pages=27-28,pagecommand=\thispagestyle{fancy}, nup = 2x1, offset=0 0, frame=false, scale=0.75]{figures/appendix/cvpr_appendix_examples_7_datasets.pdf}

\subsection{API Errors and Safety Settings of the Gemini API}
\begin{figure}[H]
    \centering
    \includegraphics[width=0.8\linewidth]{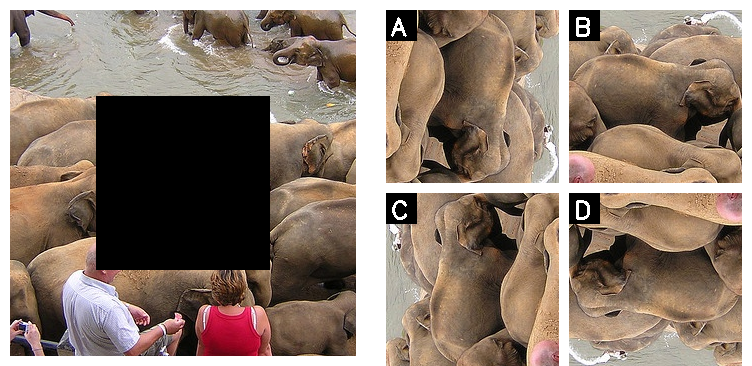}
    \caption{\textbf{Safety systems in vision-language models can be triggered by benign inputs.} Example showing an image that activated Google Gemini's content filtering mechanisms despite containing no harmful content.}
    \label{app:fig_gemini_errors1}
\end{figure}

\begin{figure}[H]
    \centering
    \includegraphics[width=0.5\linewidth]{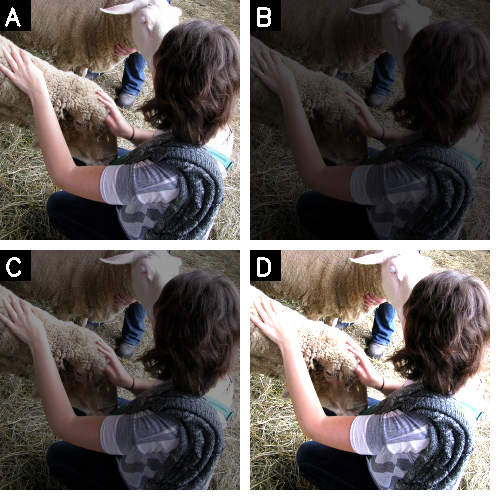}
    \caption{\textbf{Safety systems in vision-language models can be triggered by benign inputs.} Example showing an image that activated Google Gemini's content filtering mechanisms despite containing no harmful content.}    
    \label{app:fig_gemini_errors2}
\end{figure}

When conducting experiments with the Gemini API, we had to modify the default safety settings to accommodate our use case, which was already surprising. While the text safety settings could be adjusted, the image safety settings were locked and required access through a higher-tier customer account. This limitation was particularly notable given that our experiments exclusively involved standard computer vision images. Consequently, this restriction resulted in ~2\% of our tasks remaining unanswered. In~\autoref{app:fig_gemini_errors1} and ~\autoref{app:fig_gemini_errors2} we display some tasks that triggered the safety settings.

\newpage



\section{BENCHMARKING INSIGHTS AND MODEL EVALUATIONS}
\label{app_sec_stats}
\subsection{VLM Model Overview}
\label{appendix_model_overview}
\begin{table}[h]
\centering

\small
\scalebox{1}{
\begin{tabular}{|l|c|l|l|p{2cm}|l|}
\hline
\textbf{Access} & \textbf{Size} & \textbf{Name} & \textbf{Version} & \textbf{Organization} & \textbf{Release Date} \\
\hline
Closed & - & GPT-4o & gpt-4o-2024-08-06 & OpenAI & 2024-08-08 \\
Closed & - & GPT-4o-mini & gpt-4o-mini-2024-07-18 & OpenAI & 2024-07-18 \\
\hline
Closed & - & Gemini 1.5 Pro & gemini-1.5-pro-001 & Google & 2024-05-24 \\
Closed & - & Gemini 1.5 Flash & gemini-1.5-flash-001 & Google & 2024-05-24 \\
\hline
Closed & - & Claude 3.5 Sonnet & claude-3-5-sonnet-20240620 & Anthropic & 2024-06-20 \\
\hline
Open & 1B & InternVL2-1B & InternVL2-1B & OpenGVLab & 2024-07-04 \\
Open & 8B & InternVL2-8B & InternVL2-8B & OpenGVLab & 2024-07-04 \\
Open & 40B & InternVL2-40B & InternVL2-40B & OpenGVLab & 2024-07-04 \\
\hline
Open & 7B & Qwen2 7B & Qwen2-VL-7B-Instruct & Alibaba & 2024-08-30 \\
Open & 72B & Qwen2 72B & Qwen2-VL-72B-Instruct & Alibaba & 2024-08-30 \\
\hline
Open & 7B & LLaVA-NeXT 7B & llava-v1.6-mistral-7b-hf & U. of Wisconsin–Madison & 2024-01-30 \\
Open & 34B & LLaVA-NeXt 34B & lava-v1.6-34b-hf & U. of Wisconsin–Madison & 2024-01-30 \\
\hline
Open & 7B & Chameleon 7B & chameleon-7b & Meta & 2024-05-16 \\
\hline
Open & 4.2B & Phi-3 Vision & Phi-3-vision-128k-instruct & Microsoft & 2024-04-23 \\
Open & 4.2B & Phi-3.5 Vision & Phi-3.5-vision-instruct & Microsoft & 2024-08-20 \\
\hline
Open & 770M & Florence-2 & Florence-2-large-ft & Microsoft & 2024-06-15 \\
\hline
Open & 3B & PaliGemma 3B 224x224 & paligemma-3b-mix-224 & Google & 2024-05-14 \\
Open & 3B & PaliGemma 3B 448x448 & paligemma-3b-mix-448 & Google & 2024-05-14 \\
\hline
Open & 12B & Pixtral & Pixtral-12B-2409 & Mistral & 2024-09-17 \\
\hline
Open & 11B & Llama 3.2 11B & llama-3-2-11b-vision-instruct & Meta & 2024-09-25 \\
Open & 90B & Llama 3.2 90B & llama-3-2-90b-vision-instruct & Meta & 2024-09-25 \\
\hline
Open & 7B & Molmo 7B & Molmo-7B-D & Allen Institute for AI & 2024-09-24 \\
\hline
\end{tabular}
}
\caption{\textbf{Our 22 evaluated SOTA vision-language models span both open and closed-source architectures.} The benchmark includes 22 VLMs with precisely specified versions, representing current capabilities across both proprietary and publicly available models.}
\label{tab:vlm-benchmark}
\end{table}

\newpage

\subsection{Accuracy\%(t) Thresholds for All Datasets}
\label{appendix_various_fixed_percentage_lines}

\begin{figure}[h]
    \centering
    \includegraphics[width=0.87\linewidth]{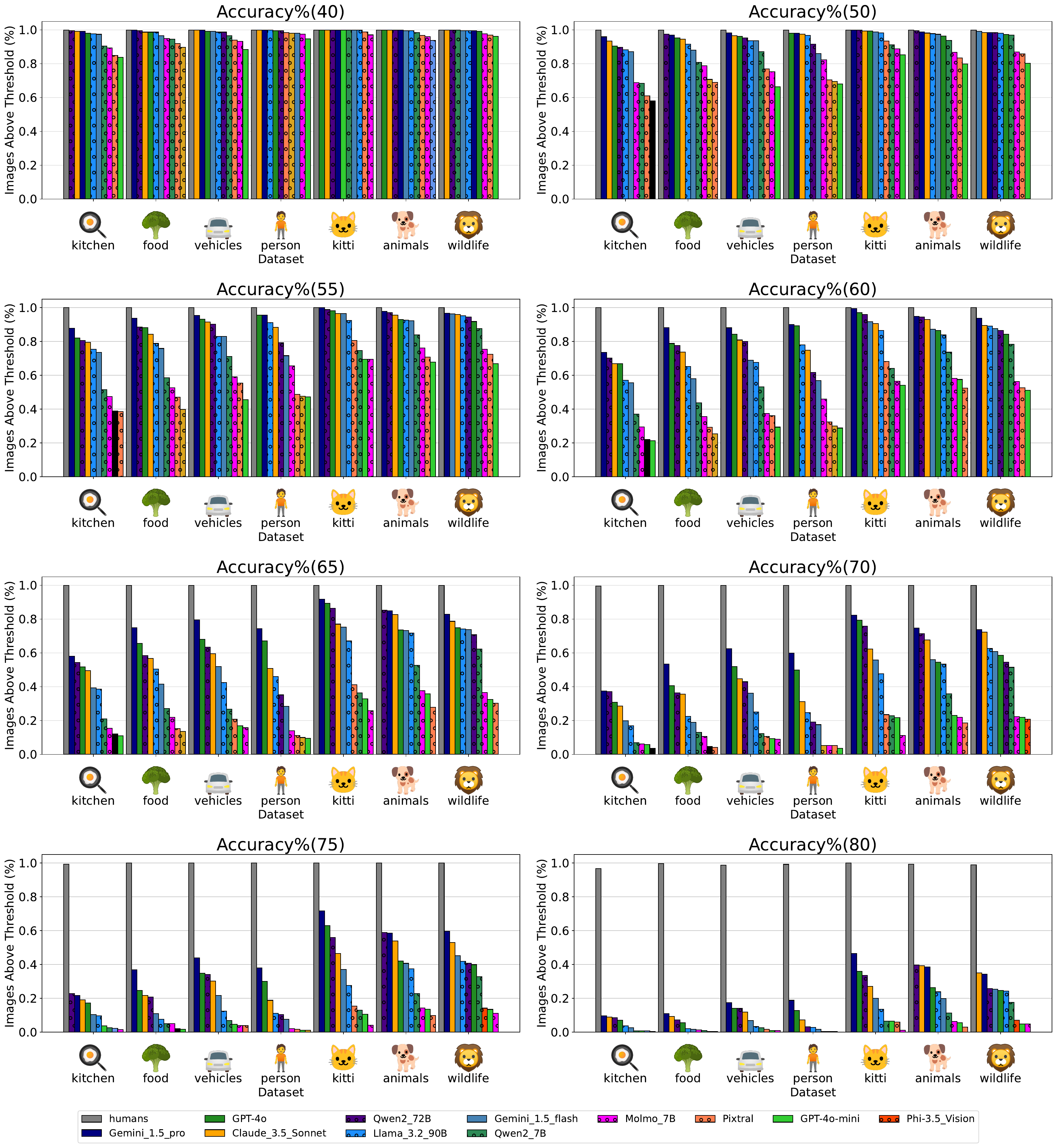}
\caption{\textbf{Accuracy\% (t) \textuparrow thresholds for the top 10 models across all datasets}. Humans (solid grey bar) consistently achieve near-perfect scores of 1 across all displayed thresholds and datasets. Performance varies significantly across imaging domains, with dotted lines representing open-source models and solid lines indicating closed-source models.}
    \label{fig:app_auc_perc_top_10}
\end{figure}

\begin{figure}[h]
    \centering
    \includegraphics[width=0.85\linewidth]{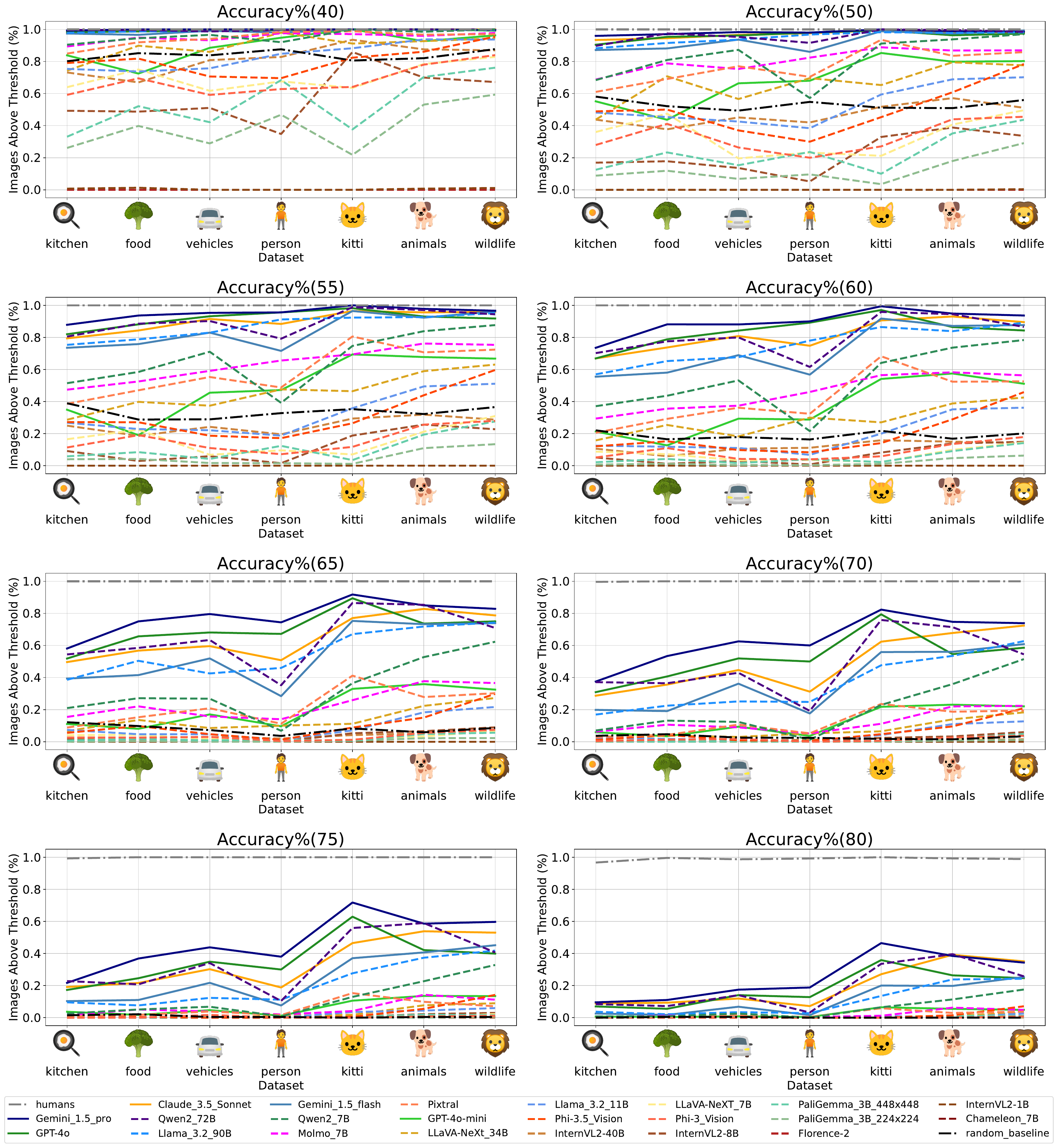}
    \caption{\textbf{Accuracy\%(t) \textuparrow thresholds for all models for each dataset}. Displayed as line plots, given the large number of models. Humans (dashed-dotted grey line) consistently achieve near-perfect scores of 1 across all displayed thresholds and datasets. Performance varies significantly across imaging domains, with dotted lines representing open-source models and solid lines indicating closed-source models.}
    \label{fig:app_auc_perc_all_models}
\end{figure}

Here we present the Accuracy\%(t) metric at thresholds 0.4, 0.5, 0.55, 0.6, 0.65, 0.70, 0.75, and 0.80. The top 10 models for each dataset are displayed as bar plots in \autoref{fig:app_auc_perc_top_10}. To increase interpretability of single models across datasets, \autoref{fig:app_auc_perc_all_models} displays all 22 models as line plots.

\newpage

\FloatBarrier  
\newpage

\newpage

\subsection{Task Ranking Comparison Between Models And Humans}
\label{appendix_task_ranking_models_humans}
\begin{figure}[h]
    \centering
    \includegraphics[width=1\linewidth]{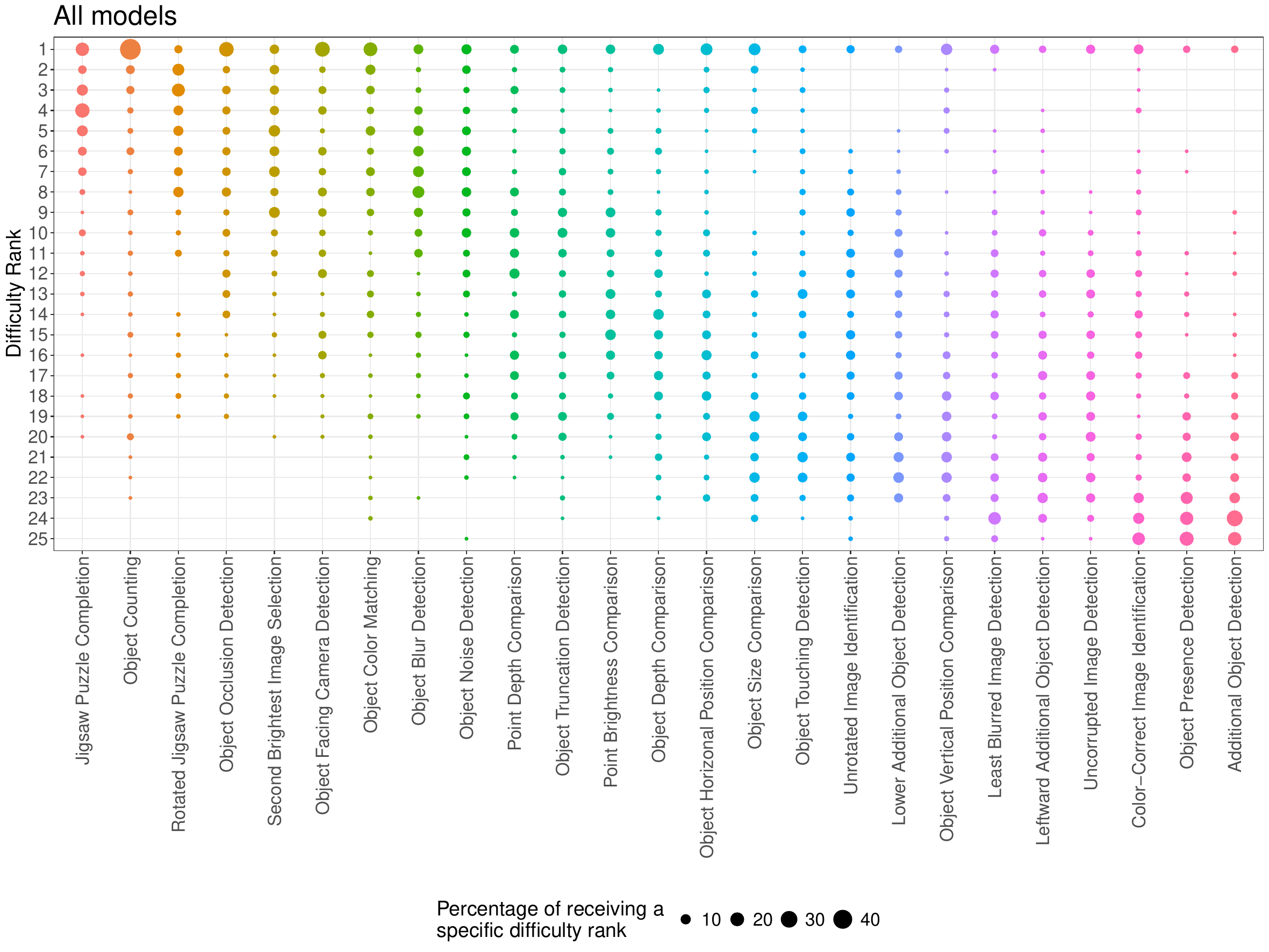}
    \caption{\textbf{All tested vision-language models, both open and closed-source, exhibit consistent patterns in task difficulty.} Aggregated ranking of tasks from easiest to most challenging, revealing systematic strengths and limitations shared across the complete set of evaluated models regardless of their accessibility.}
    \label{fig:fig5_taskdifficulty_all}
\end{figure}

\begin{figure}[H]
    \centering
    \includegraphics[width=1\linewidth]{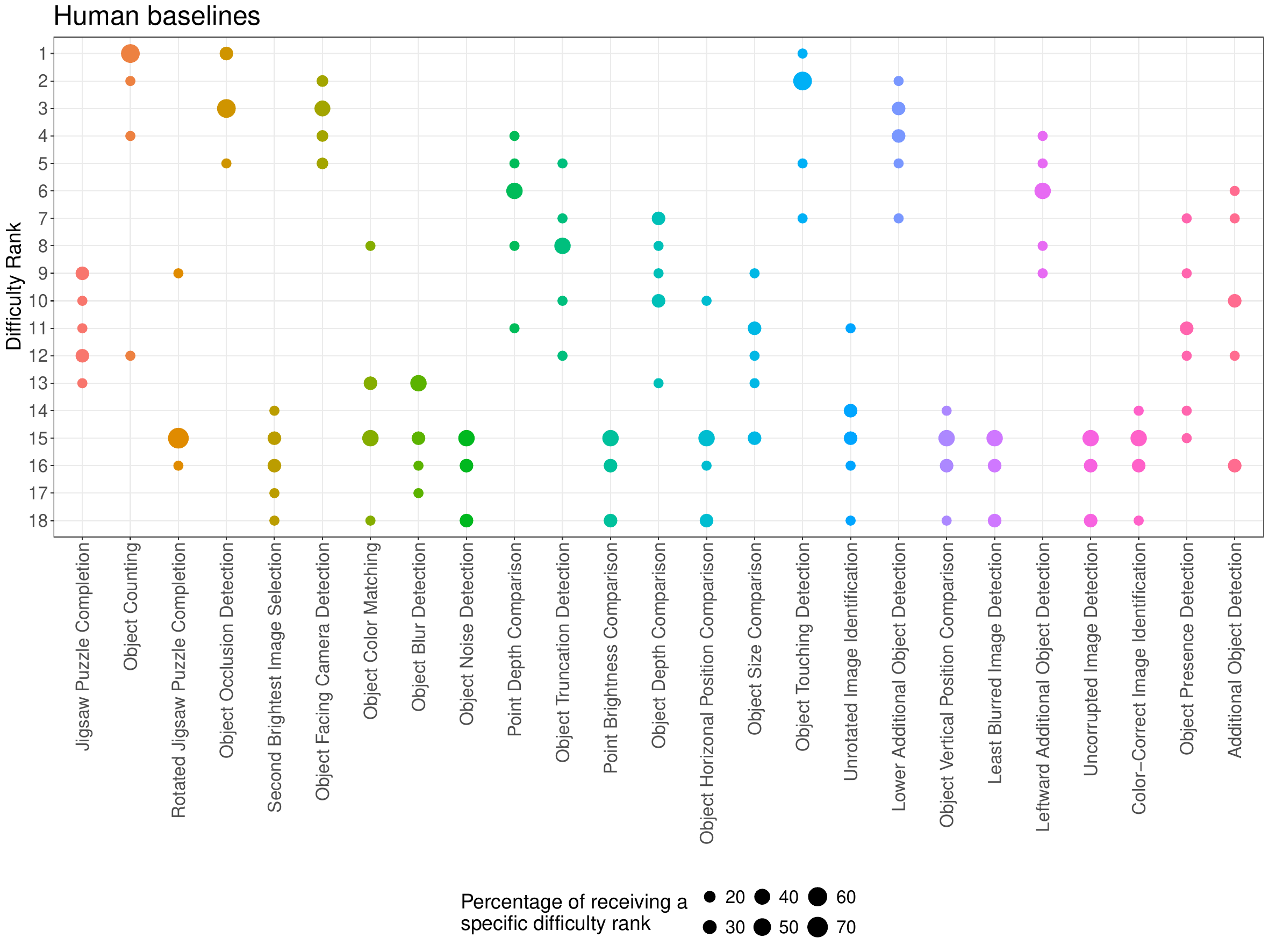}
    \caption{\textbf{Human performance reveals distinct patterns of task difficulty compared to models.} Ranking of vision tasks from easiest to most challenging based on human evaluator performance, providing a baseline for understanding natural visual capabilities.}    \label{fig:fig5_taskdifficulty_human}
\end{figure}

\begin{figure}[H]
    \centering
    \includegraphics[width=1\linewidth]{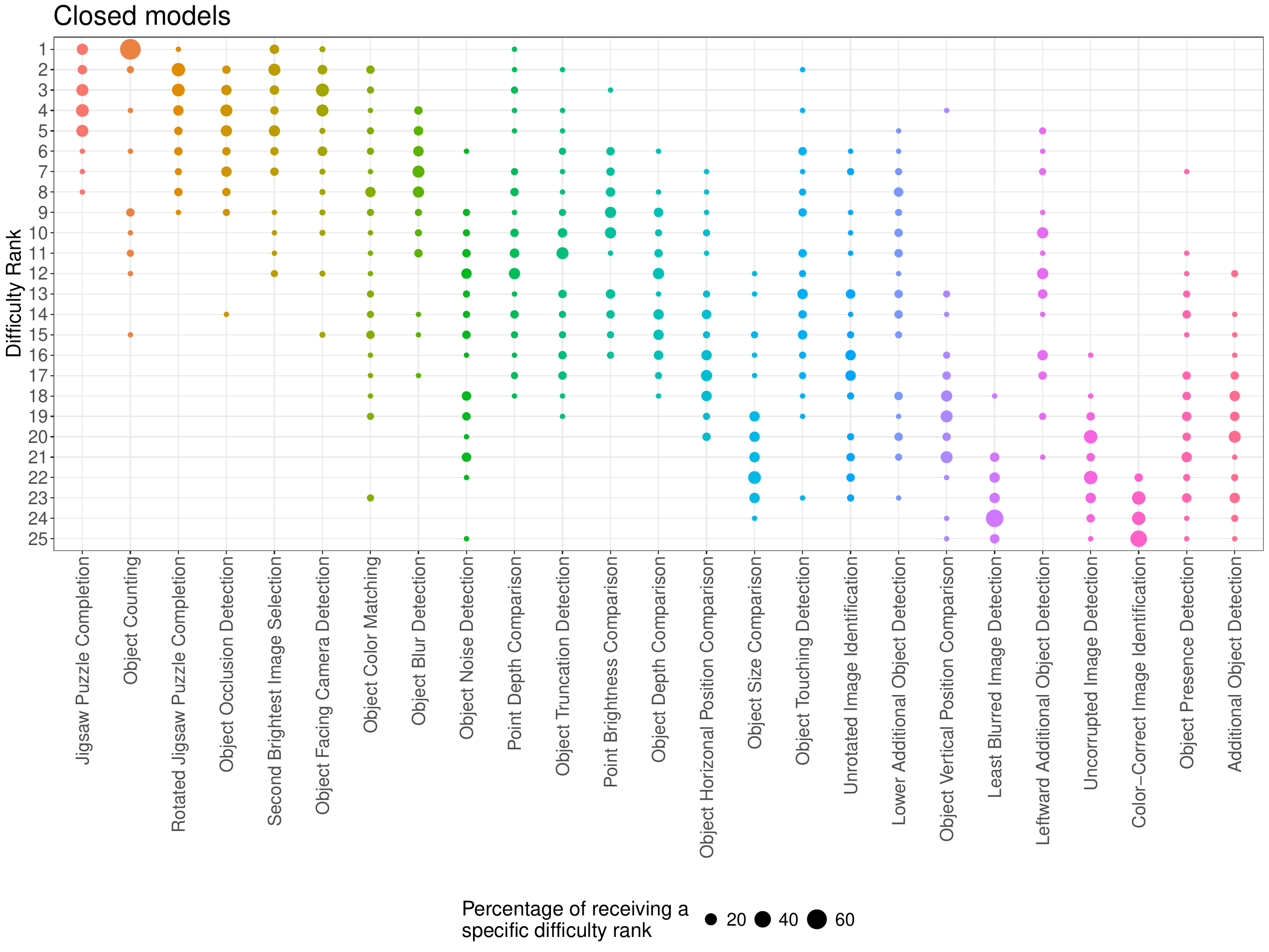}
    \caption{\textbf{Clsoed-sourced vision-language models demonstrate shared patterns in task difficulty despite architectural differences.} Aggregated ranking of tasks from easiest to most challenging, revealing systematic strengths and limitations common across all evaluated closed-source models.}    \label{fig:fig5_taskdifficulty_closed}
\end{figure}

\begin{figure}[H]
    \centering
    \includegraphics[width=1\linewidth]{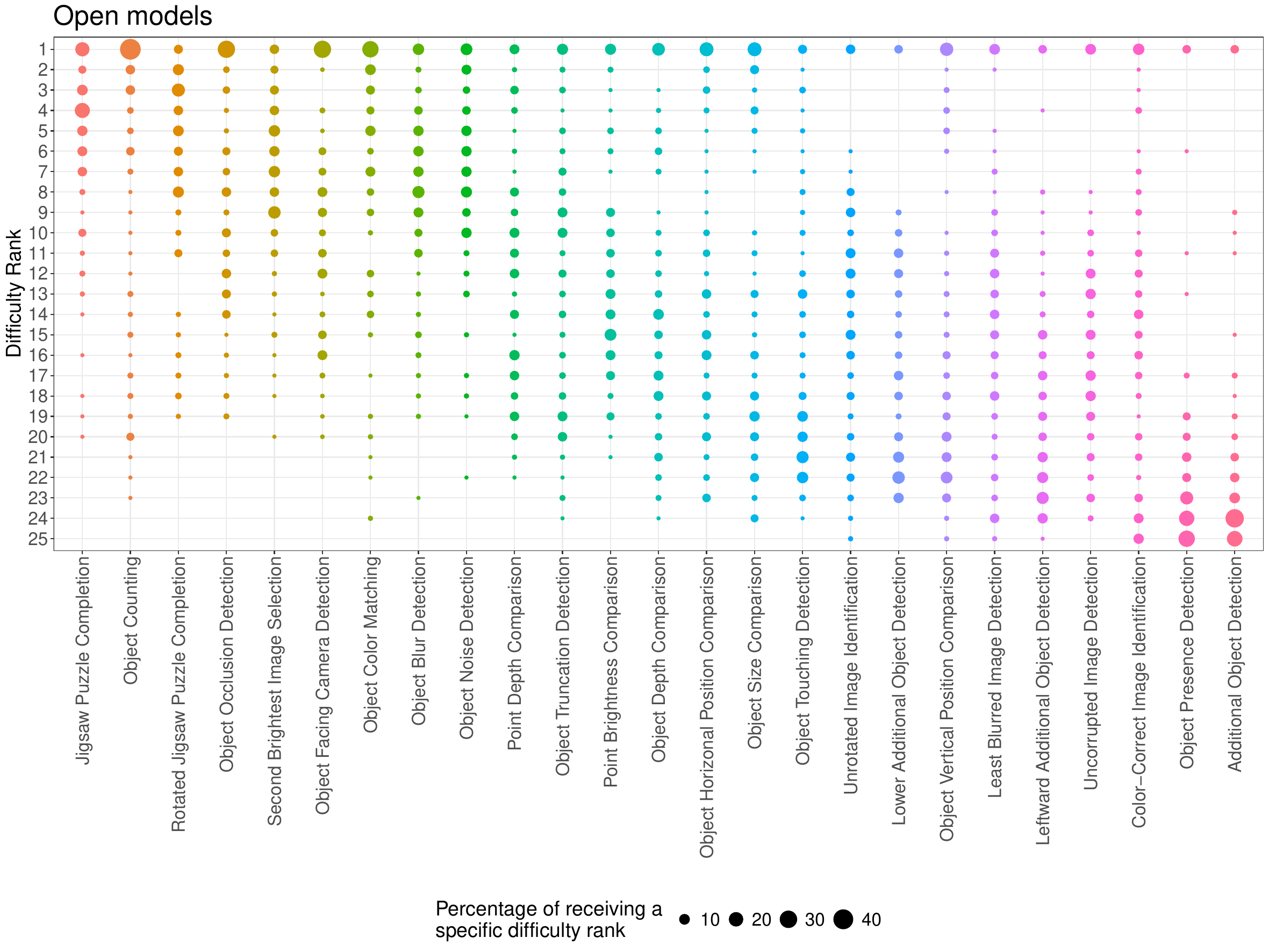}
    \caption{\textbf{Publicly available vision-language models exhibit consistent patterns in task difficulty across architectures.} Aggregated ranking of tasks from easiest to most challenging, revealing systematic strengths and limitations shared across all evaluated open-source models.}    \label{fig:fig5_taskdifficulty_open}
\end{figure}

 \textbf{Task ranking differs between models and human raters.} The plots shows the difficulty of tasks based on aggregated model scores/human scores (1 = hardest task, 25 = easiest task). The radius of the blob indicates how often a task was assigned a difficulty rank when considering all seven domains and all models (n = 5 for closed models; n = 16 for open models; n = 21 for all models; n = 1 for humans as majority vote over several raters). The larger the plot, the higher the percentage it achieved a specific rank. The hardest tasks on average across domains are (1) T7.2 “Jigsaw Puzzle Completion”, (2), T1.2 “Object Counting”, (3), T7.1 “Rotated Jigsaw Puzzle Completion”, (4), T2.1 “Object Occlusion Detection”, and (5) T5.2 “Second Brightest Image Selection”. The easiest task on average was T1.3 “Additional Object Presence Detection”. We display aggregated all models in~\autoref{fig:fig5_taskdifficulty_all}, human baselines in~\autoref{fig:fig5_taskdifficulty_human}, all closed-source models in~\autoref{fig:fig5_taskdifficulty_closed}, and all open-source models in~\autoref{fig:fig5_taskdifficulty_open}.

\newpage

\newpage

\FloatBarrier
\newpage

\subsection{Accuracies Across Datasets}
We display the regular Accuracy scores for all models per dataset in~\autoref{tab:model_accuracies} and per task in~\autoref{tab:model_performance_part2}.
\begin{table*}[h]
\centering
\begin{tabular}{l|c|ccccccc}
\toprule
Model & Overall & wildlife & animals & kitti & person & vehicles & food & kitchen \\
&  & \includegraphics[width=0.03\textwidth]{figures/icons_iclr/wildlife.png} & \includegraphics[width=0.03\textwidth]{figures/icons_iclr/animals.png} & \includegraphics[width=0.03\textwidth]{figures/icons_iclr/kitti.png} & \includegraphics[width=0.03\textwidth]{figures/icons_iclr/person.png} & \includegraphics[width=0.03\textwidth]{figures/icons_iclr/vehicles.png} & \includegraphics[width=0.03\textwidth]{figures/icons_iclr/food.png} & \includegraphics[width=0.03\textwidth]{figures/icons_iclr/kitchen.png} \\
\midrule
humans & 93.66 & 93.43 & 93.87 & 94.60 & 95.15 & 92.41 & 93.58 & 92.59 \\
\toprule
Gemini\_1.5\_pro & \cellcolor[rgb]{1,0.843,0}72.44 & \cellcolor[rgb]{1,0.843,0}74.58 & \cellcolor[rgb]{1,0.843,0}75.35 & \cellcolor[rgb]{1,0.843,0}77.96 & \cellcolor[rgb]{1,0.843,0}70.84 & \cellcolor[rgb]{1,0.843,0}71.74 & \cellcolor[rgb]{1,0.843,0}69.98 & \cellcolor[rgb]{1,0.843,0}66.60 \\
GPT-4o & \cellcolor[rgb]{0.753,0.753,0.753}69.79 & \cellcolor[rgb]{0.565,0.933,0.565}71.35 & \cellcolor[rgb]{0.565,0.933,0.565}71.35 & \cellcolor[rgb]{0.753,0.753,0.753}76.16 & \cellcolor[rgb]{0.753,0.753,0.753}69.01 & \cellcolor[rgb]{0.753,0.753,0.753}69.25 & \cellcolor[rgb]{0.753,0.753,0.753}67.00 & \cellcolor[rgb]{0.804,0.498,0.196}64.42 \\
Claude\_3.5\_Sonnet & \cellcolor[rgb]{0.804,0.498,0.196}69.00 & \cellcolor[rgb]{0.753,0.753,0.753}73.72 & \cellcolor[rgb]{0.804,0.498,0.196}74.28 & \cellcolor[rgb]{0.565,0.933,0.565}72.40 & \cellcolor[rgb]{0.804,0.498,0.196}65.34 & \cellcolor[rgb]{0.565,0.933,0.565}67.75 & \cellcolor[rgb]{0.565,0.933,0.565}65.69 & \cellcolor[rgb]{0.565,0.933,0.565}63.82 \\
Qwen2\_72B & \cellcolor[rgb]{0.565,0.933,0.565}68.76 & \cellcolor[rgb]{0.867,0.627,0.867}70.79 & \cellcolor[rgb]{0.753,0.753,0.753}75.00 & \cellcolor[rgb]{0.804,0.498,0.196}74.63 & \cellcolor[rgb]{0.678,0.847,0.902}61.70 & \cellcolor[rgb]{0.804,0.498,0.196}68.23 & \cellcolor[rgb]{0.804,0.498,0.196}66.22 & \cellcolor[rgb]{0.753,0.753,0.753}64.74 \\
Llama\_3.2\_90B & \cellcolor[rgb]{0.678,0.847,0.902}65.93 & \cellcolor[rgb]{0.678,0.847,0.902}71.33 & \cellcolor[rgb]{0.678,0.847,0.902}70.21 & \cellcolor[rgb]{0.867,0.627,0.867}68.63 & \cellcolor[rgb]{0.565,0.933,0.565}64.62 & \cellcolor[rgb]{0.867,0.627,0.867}63.13 & \cellcolor[rgb]{0.678,0.847,0.902}62.87 & \cellcolor[rgb]{0.867,0.627,0.867}60.75 \\
Gemini\_1.5\_flash & \cellcolor[rgb]{0.867,0.627,0.867}65.72 & \cellcolor[rgb]{0.804,0.498,0.196}71.53 & \cellcolor[rgb]{0.867,0.627,0.867}70.16 & \cellcolor[rgb]{0.678,0.847,0.902}70.83 & \cellcolor[rgb]{0.867,0.627,0.867}60.23 & \cellcolor[rgb]{0.678,0.847,0.902}65.17 & \cellcolor[rgb]{0.867,0.627,0.867}61.14 & \cellcolor[rgb]{0.678,0.847,0.902}61.01 \\
Qwen2\_7B & \cellcolor[rgb]{1,0.419,0.419}59.71 & \cellcolor[rgb]{1,0.419,0.419}67.82 & \cellcolor[rgb]{1,0.419,0.419}65.01 & \cellcolor[rgb]{0.529,0.808,0.922}62.22 & 51.49 & \cellcolor[rgb]{1,0.419,0.419}59.32 & \cellcolor[rgb]{1,0.419,0.419}57.34 & \cellcolor[rgb]{1,0.419,0.419}54.80 \\
Molmo\_7B & \cellcolor[rgb]{0.529,0.808,0.922}57.89 & \cellcolor[rgb]{0.529,0.808,0.922}61.11 & \cellcolor[rgb]{0.529,0.808,0.922}61.57 & \cellcolor[rgb]{1,0.9,0.8}59.77 & \cellcolor[rgb]{1,0.419,0.419}57.09 & \cellcolor[rgb]{0.529,0.808,0.922}56.12 & \cellcolor[rgb]{0.529,0.808,0.922}56.37 & \cellcolor[rgb]{0.529,0.808,0.922}53.19 \\
Pixtral & \cellcolor[rgb]{0.596,0.984,0.596}56.80 & \cellcolor[rgb]{0.596,0.984,0.596}59.77 & \cellcolor[rgb]{1,0.9,0.8}59.17 & \cellcolor[rgb]{1,0.419,0.419}63.20 & \cellcolor[rgb]{0.529,0.808,0.922}54.44 & \cellcolor[rgb]{0.596,0.984,0.596}56.08 & \cellcolor[rgb]{0.596,0.984,0.596}54.03 & \cellcolor[rgb]{0.596,0.984,0.596}50.90 \\
GPT-4o-mini & \cellcolor[rgb]{1,0.9,0.8}54.90 & \cellcolor[rgb]{1,0.9,0.8}59.19 & \cellcolor[rgb]{0.596,0.984,0.596}59.42 & \cellcolor[rgb]{0.596,0.984,0.596}60.80 & \cellcolor[rgb]{1,0.9,0.8}53.63 & \cellcolor[rgb]{1,0.9,0.8}53.71 & 46.90 & \cellcolor[rgb]{1,0.9,0.8}50.66 \\
LLaVA-NeXt\_34B & 53.26 & 57.47 & 56.67 & 53.57 & \cellcolor[rgb]{0.596,0.984,0.596}54.26 & 50.72 & \cellcolor[rgb]{1,0.9,0.8}53.11 & 47.02 \\
Llama\_3.2\_11B & 50.03 & 55.14 & 54.51 & 51.16 & 47.03 & 47.02 & 47.30 & 48.05 \\
Phi-3.5\_Vision & 49.18 & 58.14 & 51.83 & 47.86 & 44.46 & 45.43 & 49.21 & 47.34 \\
InternVL2-40B & 48.14 & 49.48 & 49.98 & 50.64 & 47.37 & 47.98 & 44.69 & 46.85 \\
LLaVA-NeXT\_7B & 44.52 & 48.84 & 46.44 & 41.38 & 42.39 & 41.52 & 47.27 & 43.79 \\
Phi-3\_Vision & 44.39 & 48.99 & 47.41 & 42.43 & 41.62 & 42.12 & 45.97 & 42.18 \\
InternVL2-8B & 41.28 & 43.72 & 45.60 & 46.76 & 35.59 & 38.95 & 39.26 & 39.05 \\
PaliGemma\_3B\_448x448 & 40.66 & 47.07 & 45.02 & 36.00 & 43.14 & 37.82 & 40.64 & 34.94 \\
PaliGemma\_3B\_224x224 & 36.43 & 41.26 & 39.73 & 31.57 & 38.26 & 34.13 & 37.32 & 32.76 \\
InternVL2-1B & 16.69 & 18.05 & 15.51 & 18.31 & 15.94 & 14.54 & 19.58 & 14.89 \\
Florence-2 & 15.47 & 17.80 & 16.02 & 15.01 & 18.29 & 13.85 & 16.67 & 10.65 \\
Chameleon\_7B & 0.00 & 0.00 & 0.00 & 0.00 & 0.00 & 0.00 & 0.00 & 0.00 \\
\bottomrule
\end{tabular}
\caption{\textbf{Model Accuracies across different datasets.} Performance varies greatly between domain datasets, highlighting the need for in-domain validation. For each column, the top 10 models are highlighted: \textcolor[rgb]{1,0.843,0}{\rule{0.5cm}{0.5cm}} 1st place (Gold) \quad \textcolor[rgb]{0.753,0.753,0.753}{\rule{0.5cm}{0.5cm}} 2nd place (Silver) \quad \textcolor[rgb]{0.804,0.498,0.196}{\rule{0.5cm}{0.5cm}} 3rd place (Bronze) \quad \textcolor[rgb]{0.565,0.933,0.565}{\rule{0.5cm}{0.5cm}} 4th place \quad \textcolor[rgb]{0.678,0.847,0.902}{\rule{0.5cm}{0.5cm}} 5th place \quad \textcolor[rgb]{0.867,0.627,0.867}{\rule{0.5cm}{0.5cm}} 6th place \quad \textcolor[rgb]{1,0.419,0.419}{\rule{0.5cm}{0.5cm}} 7th place \quad \textcolor[rgb]{0.529,0.808,0.922}{\rule{0.5cm}{0.5cm}} 8th place \quad \textcolor[rgb]{0.596,0.984,0.596}{\rule{0.5cm}{0.5cm}} 9th place \quad \textcolor[rgb]{1,0.9,0.8}{\rule{0.5cm}{0.5cm}} 10th place. The 'Overall' column represents the mean accuracy across all datasets.}
\label{tab:model_accuracies}
\end{table*}

\newpage




\subsection{Ranking Variability Across Datasets}
Scatter plots illustrating ranking diversity first present a combined view across all domains, followed by separate plots for each domain. These visualizations reveal the variations in rankings both globally and within datasets.

\begin{figure}[H]
    \centering
    \includegraphics[width=0.7\linewidth]{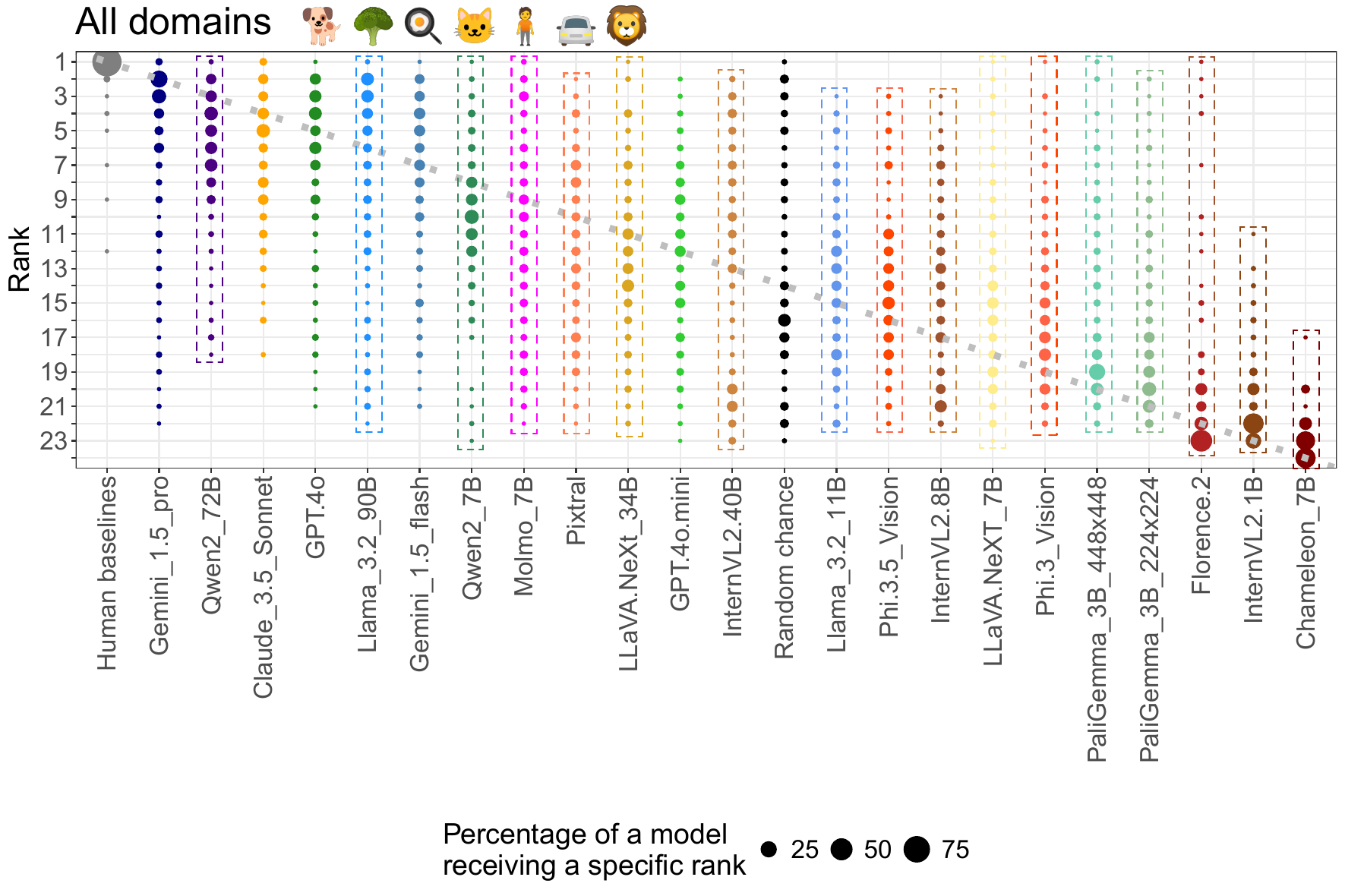}
    \caption{\textbf{Ranking variation across models for all datasets combined.} Depicted as a scatter plot. The radius of each blob at position $(Model\_i, rank\_j)$ is proportional to the percentage that model $Model\_i$ achieved rank $j$. Open models are indicated by a dashed border.}
    \label{app:fig_ranking_scatter_all}
\end{figure}

\begin{figure}[H]
    \centering
    \includegraphics[width=0.71\linewidth]{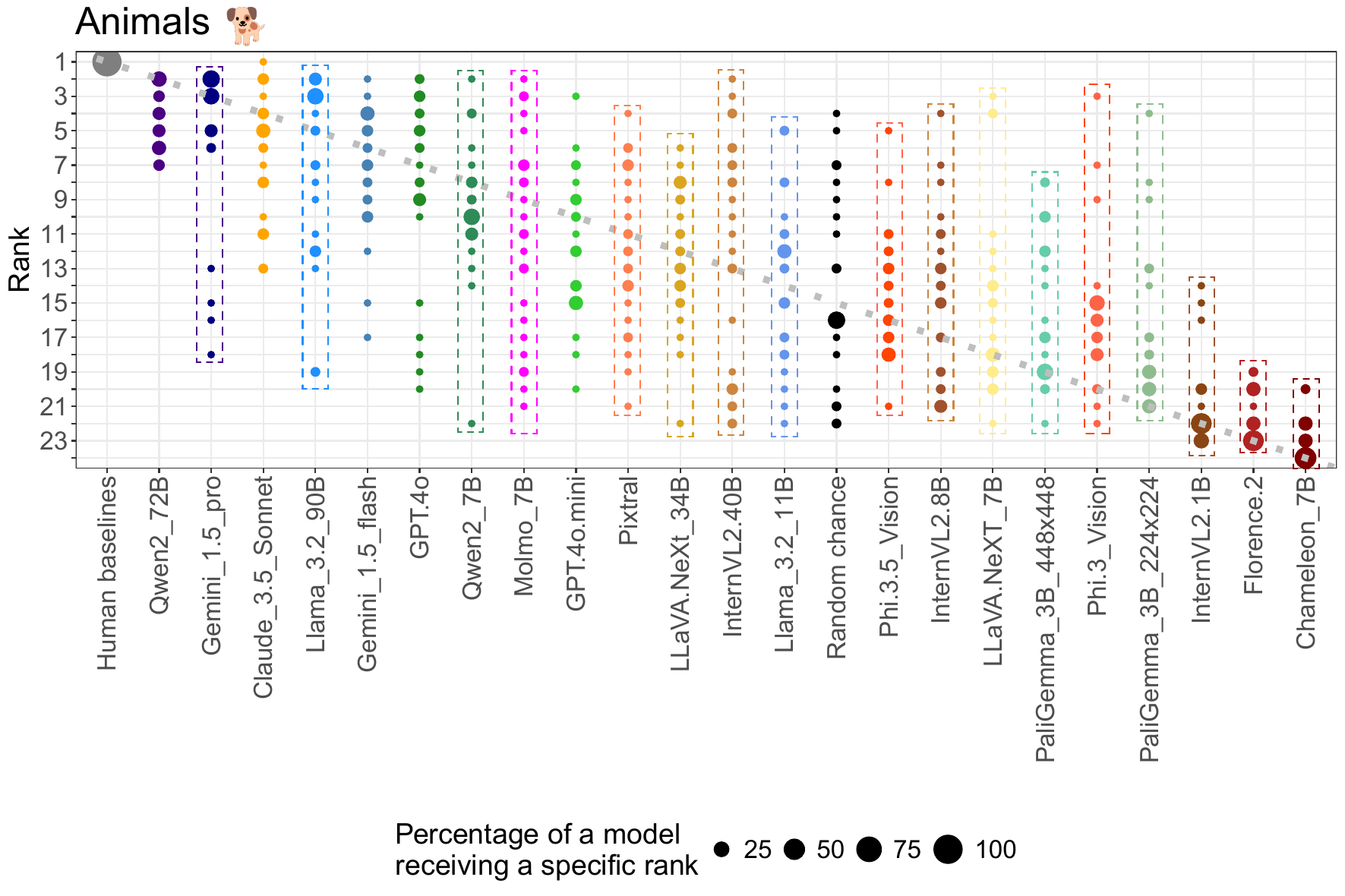}
    \caption{\textbf{Ranking variation across models for the animals dataset.} Depicted as a scatter plot. The radius of each blob at position $(Model\_i, rank\_j)$ is proportional to the percentage that model $Model\_i$ achieved rank $j$. Open models are indicated by a dashed border.}
    \label{app:fig_ranking_scatter_animals}
\end{figure}

\newpage

\begin{figure}[H]
    \centering
    \includegraphics[width=0.71\linewidth]{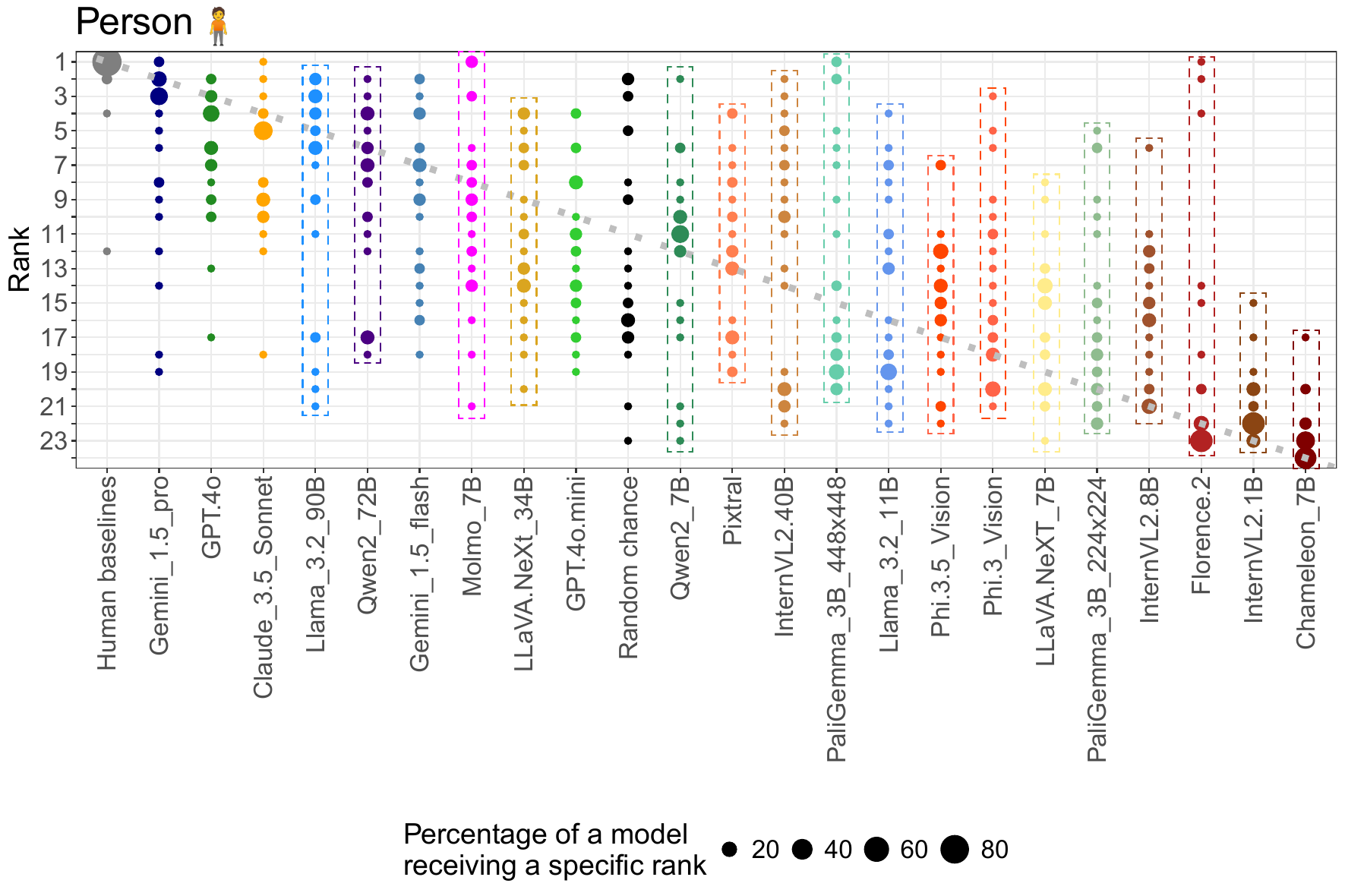}
    \caption{\textbf{Ranking variation across models for the person dataset.} Depicted as a scatter plot. The radius of each blob at position $(Model\_i, rank\_j)$ is proportional to the percentage that model $Model\_i$ achieved rank $j$. Open models are indicated by a dashed border.}
    \label{app:fig_ranking_scatter_person}
\end{figure}

\begin{figure}[H]
    \centering
    \includegraphics[width=0.71\linewidth]{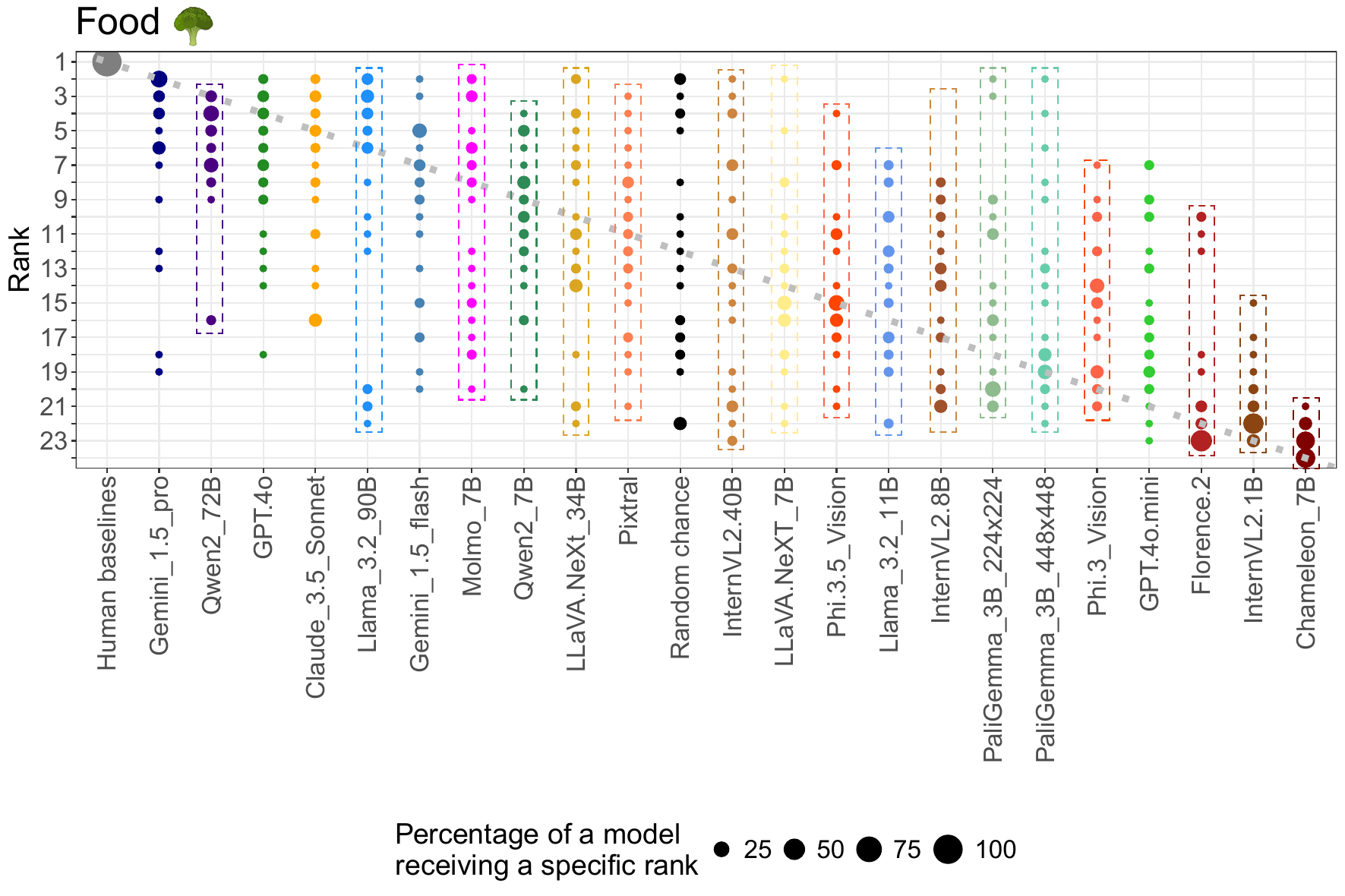}
    \caption{\textbf{Ranking variation across models for the food dataset.} Depicted as a scatter plot. The radius of each blob at position $(Model\_i, rank\_j)$ is proportional to the percentage that model $Model\_i$ achieved rank $j$. Open models are indicated by a dashed border.}
    \label{app:fig_ranking_scatter_food}
\end{figure}

\newpage

\begin{figure}[H]
    \centering
    \includegraphics[width=0.7\linewidth]{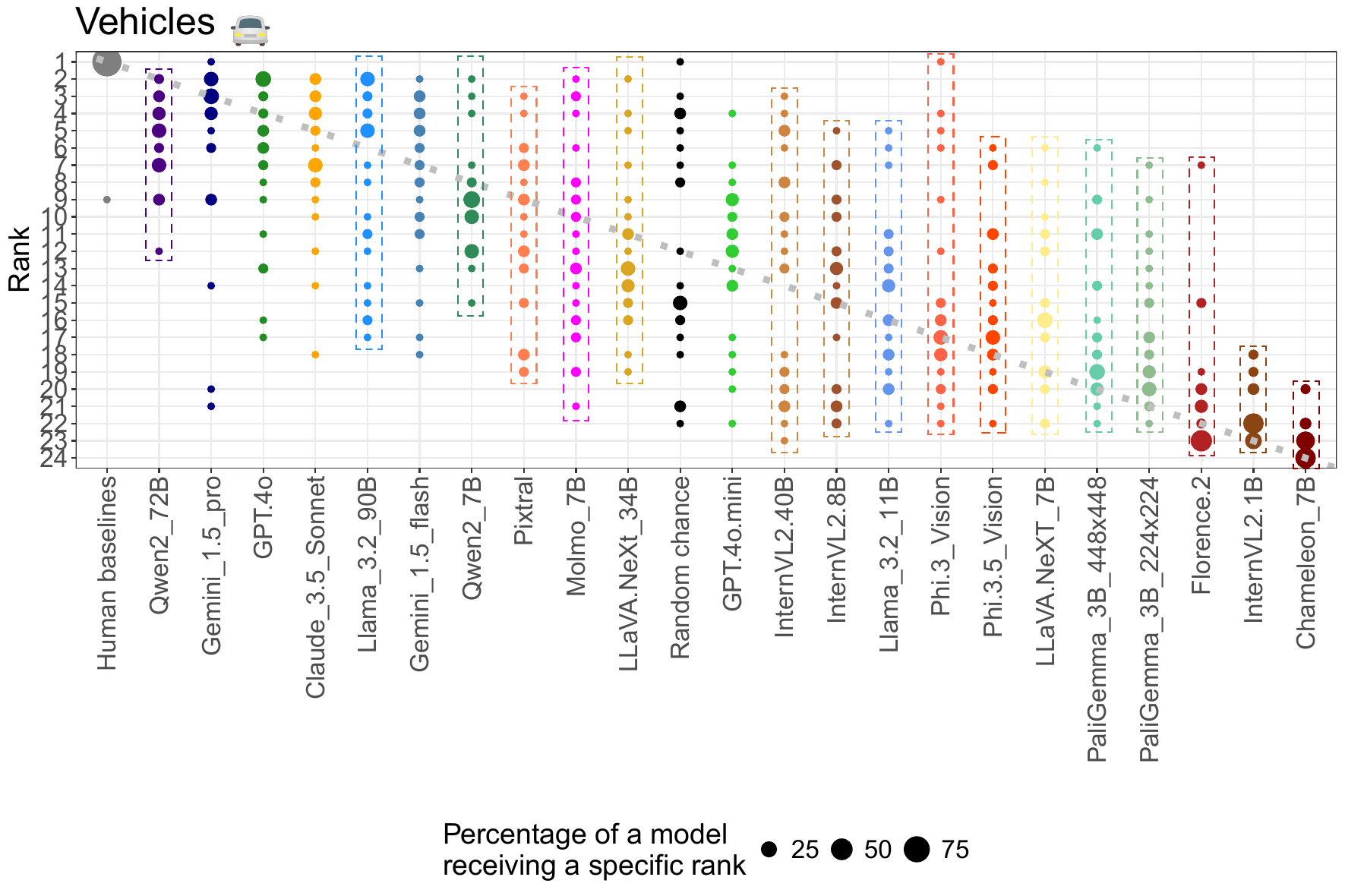}
    \caption{\textbf{Ranking variation across models for the vehicles dataset.} Depicted as a scatter plot. The radius of each blob at position $(Model\_i, rank\_j)$ is proportional to the percentage that model $Model\_i$ achieved rank $j$. Open models are indicated by a dashed border.}
    \label{app:fig_ranking_scatter_vehicles}
\end{figure}

\begin{figure}[H]
    \centering
    \includegraphics[width=0.7\linewidth]{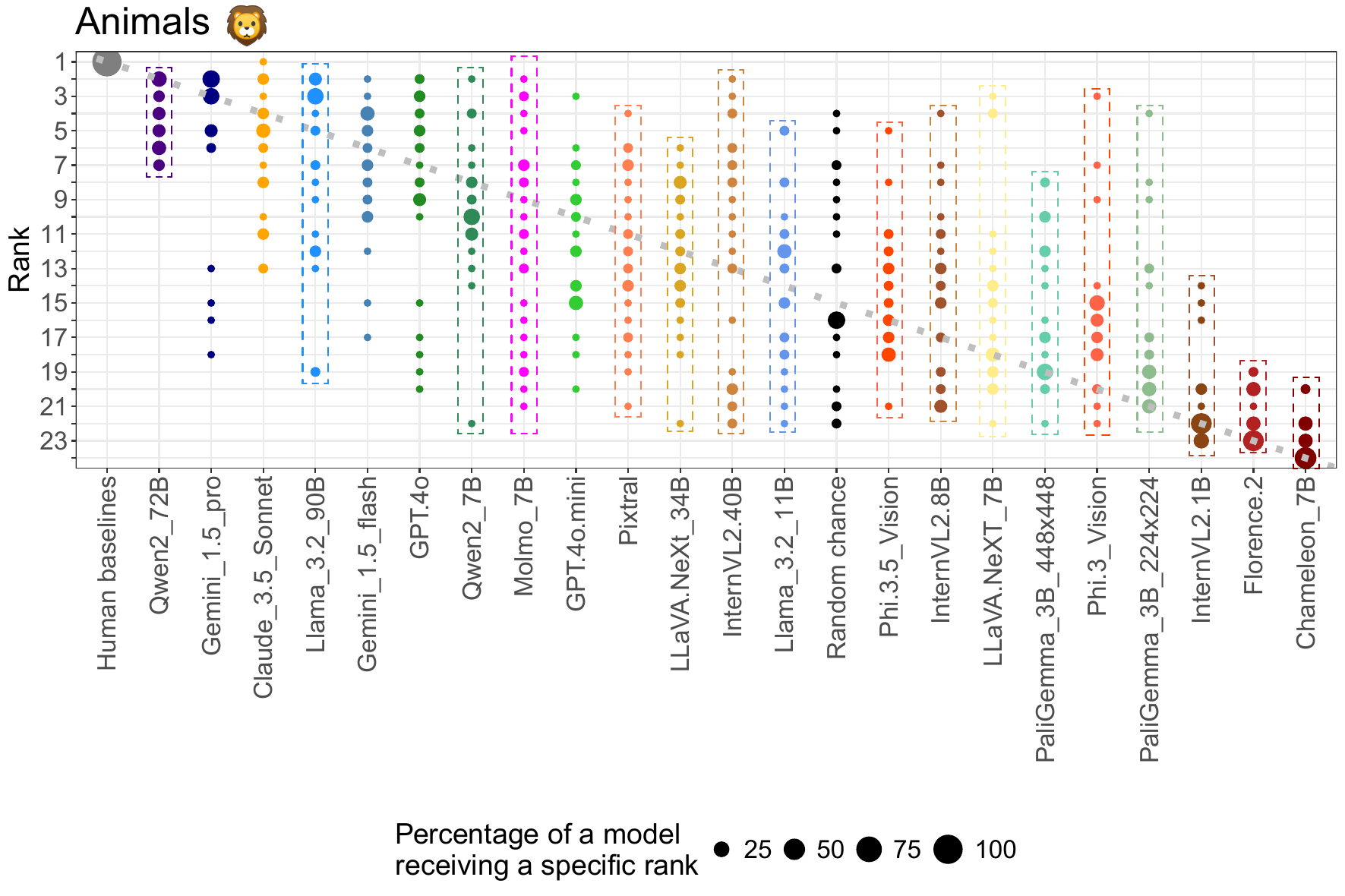}
    \caption{\textbf{Ranking variation across models for the wildlife dataset.} Depicted as a scatter plot. The radius of each blob at position $(Model\_i, rank\_j)$ is proportional to the percentage that model $Model\_i$ achieved rank $j$. Open models are indicated by a dashed border.}
    \label{app:fig_ranking_scatter_wildlife}
\end{figure}

\newpage
\clearpage
\FloatBarrier

\begin{figure}[H]
    \centering
    \includegraphics[width=0.7\linewidth]{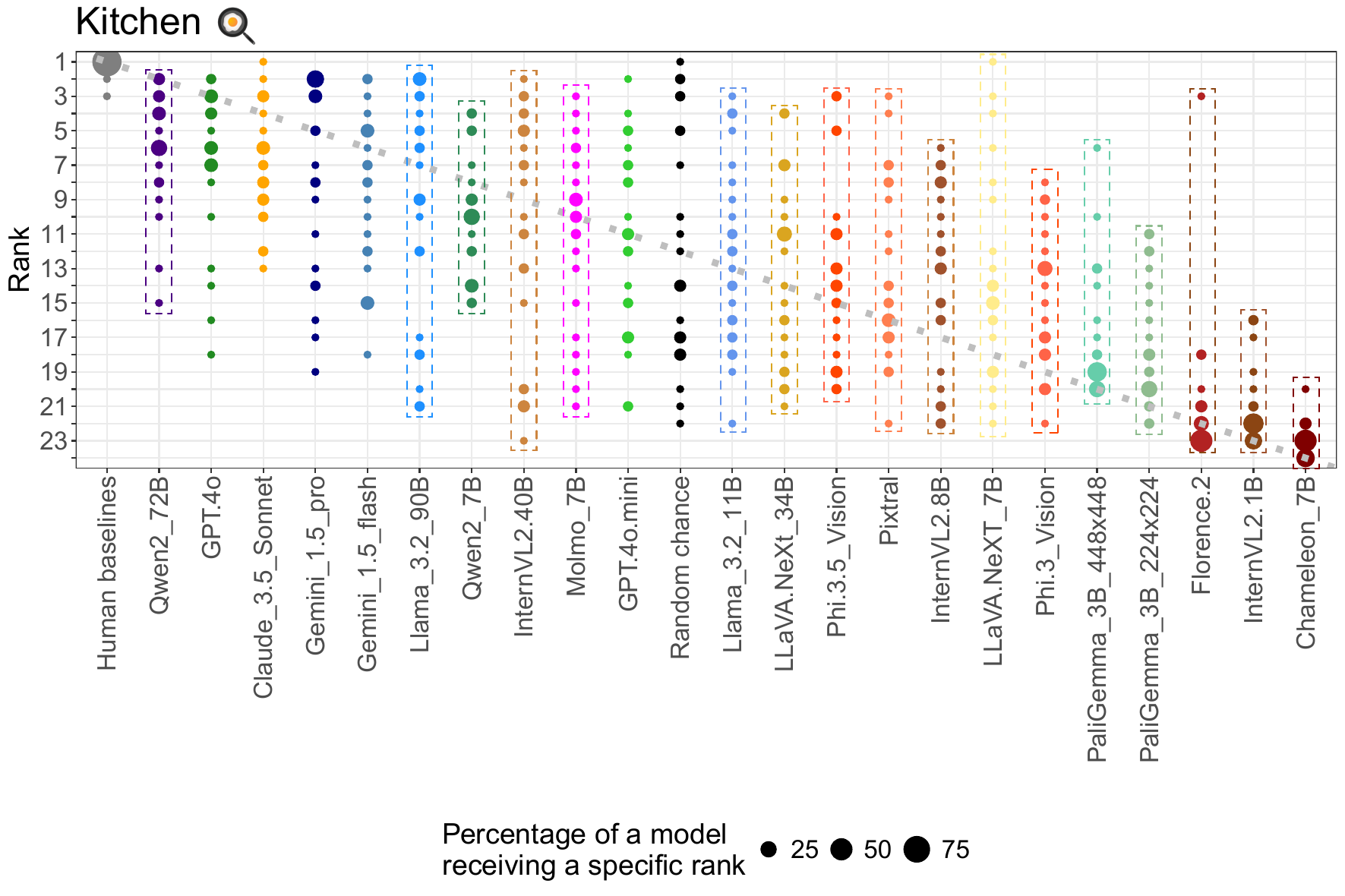}
    \caption{\textbf{Ranking variation across models for the kitchen dataset.} Depicted as a scatter plot. The radius of each blob at position $(Model\_i, rank\_j)$ is proportional to the percentage that model $Model\_i$ achieved rank $j$. Open models are indicated by a dashed border.}
    \label{app:fig_ranking_scatter_kitchen}
\end{figure}

\begin{figure}[H]
    \centering
    \includegraphics[width=0.71\linewidth]{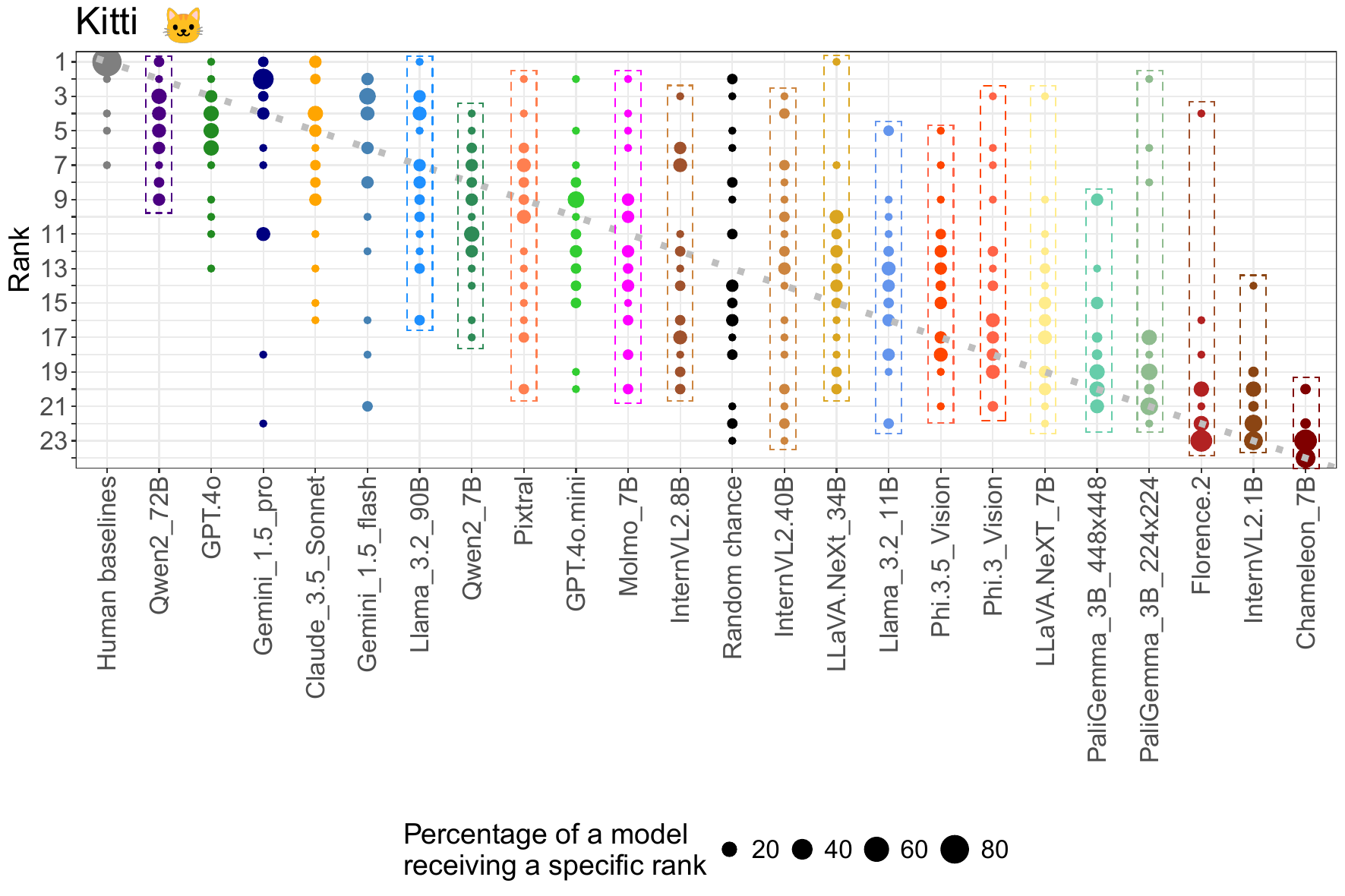}
    \caption{\textbf{Ranking variation across models for the kitti dataset.} Depicted as a scatter plot. The radius of each blob at position $(Model\_i, rank\_j)$ is proportional to the percentage that model $Model\_i$ achieved rank $j$. Open models are indicated by a dashed border.}
    \label{app:fig_ranking_scatter_kitti}
\end{figure}

\newpage

\subsection{Task Difficulty Comparison by Metadata Source and Ambiguity}
\label{appendix_fig6_instance}
\begin{figure}[h]
    \centering
    \includegraphics[width=\linewidth]{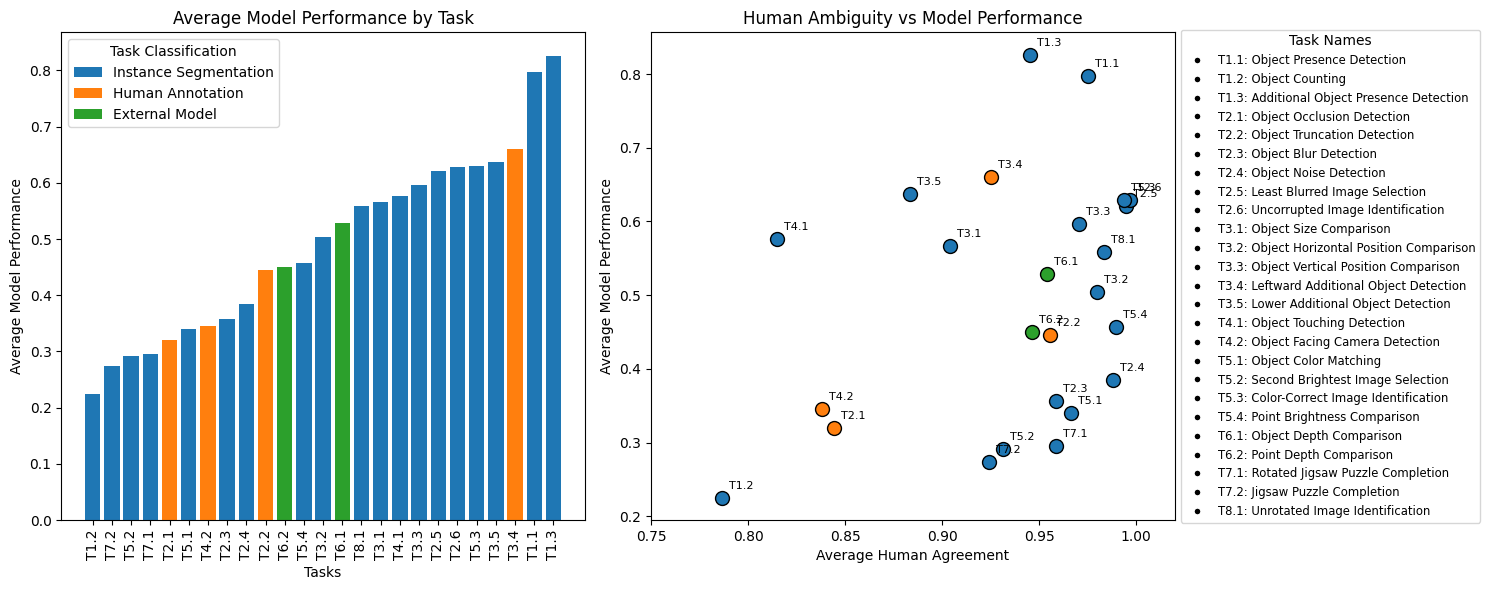}
    \caption{\textbf{Instance segmentations alone allow for the extraction of hard tasks.} (a) Tasks were classified in those extractable directly from instance segmentations (blue), requiring external models (green) and requiring human annotations (red). (b) Human ambiguity plotted against model performance.}
    \label{fig:fig6_instance}
\end{figure}

\newpage

\subsection{Task Diversity: Model Accuracies Across Tasks}
\label{appendix_spider_figures}
\begin{figure}
    \centering
    \includegraphics[width=0.8\linewidth]{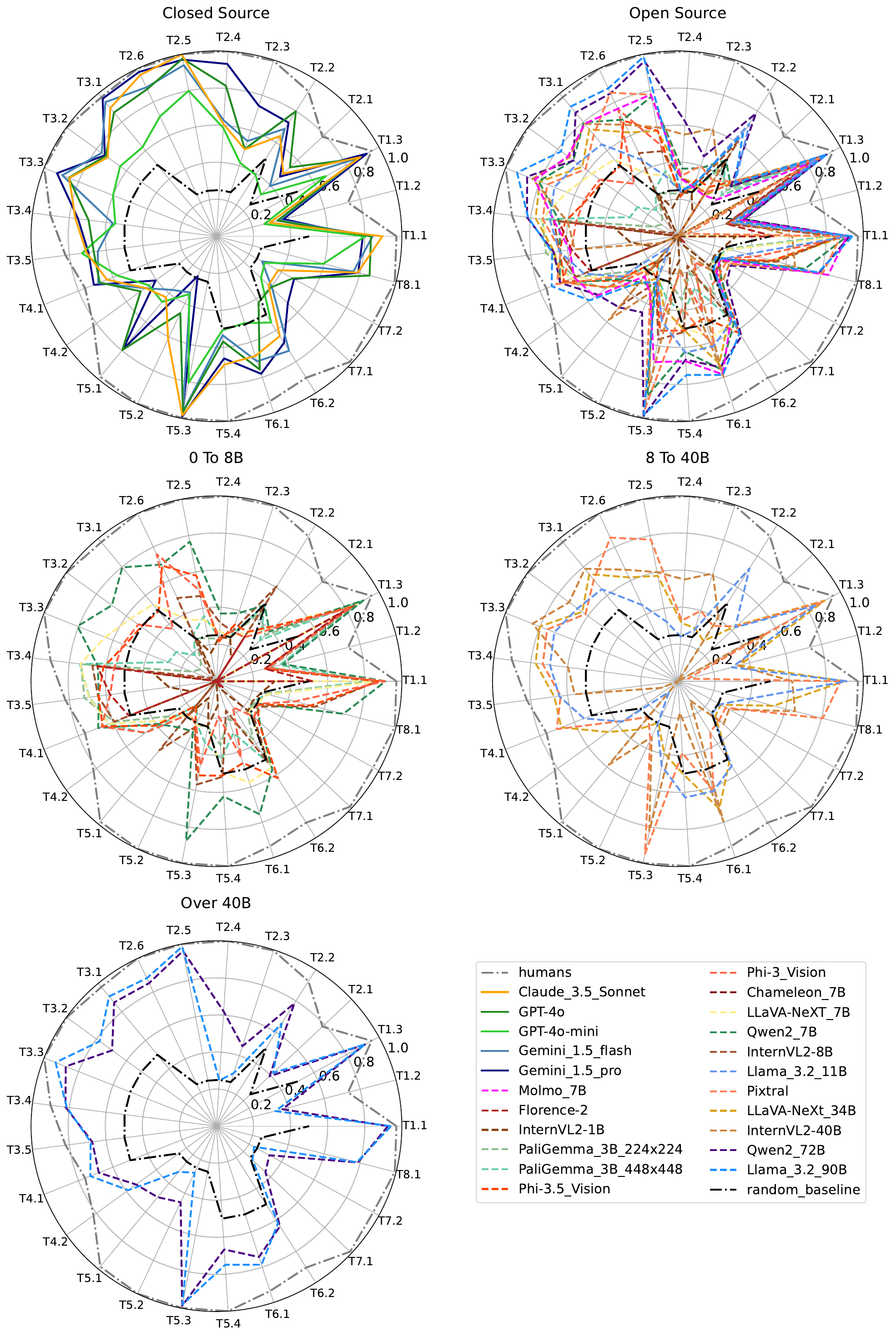}
    \caption{\textbf{Model architecture and size significantly impact performance patterns across diverse vision tasks.} Spider plots reveal distinct performance profiles between open-source (dashed lines) and closed-source (solid lines) models across our comprehensive task framework. Each axis represents task-specific accuracy, demonstrating how different model characteristics influence capabilities.}
    \label{app:spider_plot}
\end{figure}

Here, we present the performance of models across all tasks, using Accuracy (see~\autoref{tab:model_performance_part2}). Spider plots for closed-source, open-source, and varying model sizes are displayed in~\autoref{app:spider_plot}.

\begin{table*}[h]
\centering
\scriptsize 
\begin{tabular}{l|c|ccccccccccccc}
\toprule
Model & Rank & T1.1 & T1.2 & T1.3 & T2.1 & T2.2 & T2.3 & T2.4 & T2.5 & T2.6 & T3.1 & T3.2 & T3.3 & T3.4 \\
\midrule
humans &  & 96.88 & 68.31 & 95.24 & 78.93 & 92.92 & 99.51 & 100.00 & 100.00 & 100.00 & 97.21 & 99.83 & 99.90 & 89.74 \\
\midrule
Gemini\_1.5\_pro & \cellcolor[rgb]{1,0.843,0}1 & 78.23 & \cellcolor[rgb]{0.804,0.498,0.196}33.68 & \cellcolor[rgb]{1,0.843,0}92.32 & \cellcolor[rgb]{0.804,0.498,0.196}47.74 & \cellcolor[rgb]{0.804,0.498,0.196}72.91 & \cellcolor[rgb]{1,0.843,0}75.09 & \cellcolor[rgb]{1,0.843,0}93.33 & \cellcolor[rgb]{0.565,0.933,0.565}97.11 & \cellcolor[rgb]{1,0.843,0}98.15 & \cellcolor[rgb]{1,0.843,0}96.22 & \cellcolor[rgb]{0.804,0.498,0.196}75.95 & \cellcolor[rgb]{0.753,0.753,0.753}92.18 & \cellcolor[rgb]{0.678,0.847,0.902}74.85 \\
GPT-4o & \cellcolor[rgb]{0.753,0.753,0.753}2 & 83.45 & \cellcolor[rgb]{0.529,0.808,0.922}30.35 & 87.17 & \cellcolor[rgb]{1,0.843,0}54.94 & \cellcolor[rgb]{1,0.843,0}80.41 & \cellcolor[rgb]{0.678,0.847,0.902}50.45 & \cellcolor[rgb]{0.753,0.753,0.753}82.36 & \cellcolor[rgb]{0.804,0.498,0.196}97.62 & \cellcolor[rgb]{0.678,0.847,0.902}87.92 & \cellcolor[rgb]{0.678,0.847,0.902}87.07 & \cellcolor[rgb]{0.565,0.933,0.565}74.99 & \cellcolor[rgb]{0.678,0.847,0.902}85.63 & \cellcolor[rgb]{1,0.9,0.8}70.45 \\
Claude\_3.5\_Sonnet & \cellcolor[rgb]{0.804,0.498,0.196}3 & \cellcolor[rgb]{1,0.419,0.419}88.98 & 25.70 & \cellcolor[rgb]{0.596,0.984,0.596}88.46 & \cellcolor[rgb]{0.753,0.753,0.753}49.96 & \cellcolor[rgb]{1,0.419,0.419}64.56 & \cellcolor[rgb]{0.565,0.933,0.565}50.46 & \cellcolor[rgb]{0.565,0.933,0.565}63.32 & \cellcolor[rgb]{1,0.843,0}99.83 & \cellcolor[rgb]{0.753,0.753,0.753}96.37 & \cellcolor[rgb]{0.565,0.933,0.565}89.89 & \cellcolor[rgb]{0.867,0.627,0.867}74.08 & \cellcolor[rgb]{0.867,0.627,0.867}85.23 & \cellcolor[rgb]{0.565,0.933,0.565}76.66 \\
Qwen2\_72B & \cellcolor[rgb]{0.565,0.933,0.565}4 & \cellcolor[rgb]{0.753,0.753,0.753}92.10 & \cellcolor[rgb]{1,0.843,0}34.11 & \cellcolor[rgb]{0.804,0.498,0.196}91.18 & 39.80 & \cellcolor[rgb]{0.753,0.753,0.753}78.67 & \cellcolor[rgb]{0.867,0.627,0.867}45.76 & \cellcolor[rgb]{0.678,0.847,0.902}62.40 & \cellcolor[rgb]{0.678,0.847,0.902}96.39 & \cellcolor[rgb]{1,0.419,0.419}85.35 & \cellcolor[rgb]{0.867,0.627,0.867}86.97 & \cellcolor[rgb]{0.529,0.808,0.922}71.26 & \cellcolor[rgb]{0.565,0.933,0.565}86.96 & \cellcolor[rgb]{1,0.843,0}82.08 \\
Llama\_3.2\_90B & \cellcolor[rgb]{0.678,0.847,0.902}5 & \cellcolor[rgb]{1,0.843,0}93.82 & \cellcolor[rgb]{0.596,0.984,0.596}30.11 & \cellcolor[rgb]{0.753,0.753,0.753}91.46 & \cellcolor[rgb]{0.596,0.984,0.596}41.42 & \cellcolor[rgb]{0.867,0.627,0.867}65.21 & 30.21 & 24.82 & \cellcolor[rgb]{0.753,0.753,0.753}98.53 & \cellcolor[rgb]{0.565,0.933,0.565}88.38 & \cellcolor[rgb]{0.804,0.498,0.196}90.75 & \cellcolor[rgb]{1,0.843,0}79.31 & \cellcolor[rgb]{1,0.843,0}93.00 & \cellcolor[rgb]{0.753,0.753,0.753}81.21 \\
Gemini\_1.5\_flash & \cellcolor[rgb]{0.867,0.627,0.867}6 & 84.21 & \cellcolor[rgb]{0.565,0.933,0.565}31.33 & \cellcolor[rgb]{1,0.9,0.8}87.58 & 39.87 & \cellcolor[rgb]{0.678,0.847,0.902}68.89 & \cellcolor[rgb]{0.804,0.498,0.196}54.81 & \cellcolor[rgb]{0.804,0.498,0.196}63.51 & \cellcolor[rgb]{0.867,0.627,0.867}94.22 & \cellcolor[rgb]{0.804,0.498,0.196}89.21 & \cellcolor[rgb]{0.753,0.753,0.753}94.00 & \cellcolor[rgb]{0.753,0.753,0.753}77.79 & \cellcolor[rgb]{0.804,0.498,0.196}89.16 & 55.65 \\
Qwen2\_7B & \cellcolor[rgb]{1,0.419,0.419}7 & \cellcolor[rgb]{0.565,0.933,0.565}90.65 & \cellcolor[rgb]{0.753,0.753,0.753}33.82 & \cellcolor[rgb]{0.565,0.933,0.565}90.11 & 39.87 & \cellcolor[rgb]{0.596,0.984,0.596}48.14 & \cellcolor[rgb]{0.596,0.984,0.596}39.38 & \cellcolor[rgb]{0.529,0.808,0.922}38.14 & \cellcolor[rgb]{1,0.9,0.8}76.98 & 69.56 & \cellcolor[rgb]{1,0.419,0.419}80.16 & \cellcolor[rgb]{0.596,0.984,0.596}70.88 & \cellcolor[rgb]{0.596,0.984,0.596}80.03 & 65.24 \\
Molmo\_7B & \cellcolor[rgb]{0.529,0.808,0.922}8 & \cellcolor[rgb]{0.867,0.627,0.867}90.23 & \cellcolor[rgb]{1,0.419,0.419}30.88 & 84.30 & 28.99 & 25.84 & 27.54 & 30.74 & \cellcolor[rgb]{0.596,0.984,0.596}78.23 & \cellcolor[rgb]{0.529,0.808,0.922}76.44 & \cellcolor[rgb]{0.529,0.808,0.922}79.53 & \cellcolor[rgb]{0.678,0.847,0.902}74.20 & \cellcolor[rgb]{0.529,0.808,0.922}82.49 & 67.28 \\
Pixtral & \cellcolor[rgb]{0.596,0.984,0.596}9 & 86.84 & 7.14 & \cellcolor[rgb]{0.529,0.808,0.922}88.75 & \cellcolor[rgb]{0.867,0.627,0.867}43.64 & 37.04 & \cellcolor[rgb]{0.529,0.808,0.922}41.20 & \cellcolor[rgb]{0.596,0.984,0.596}34.18 & \cellcolor[rgb]{0.529,0.808,0.922}78.29 & \cellcolor[rgb]{0.867,0.627,0.867}85.89 & 71.29 & 63.09 & 75.85 & \cellcolor[rgb]{0.596,0.984,0.596}72.88 \\
GPT-4o-mini & \cellcolor[rgb]{1,0.9,0.8}10 & 79.86 & 23.35 & 67.68 & 33.83 & 39.76 & \cellcolor[rgb]{1,0.419,0.419}43.70 & \cellcolor[rgb]{0.867,0.627,0.867}60.91 & \cellcolor[rgb]{1,0.419,0.419}80.17 & \cellcolor[rgb]{1,0.9,0.8}72.04 & 62.47 & 65.35 & 59.01 & 54.33 \\
LLaVA-NeXt\_34B & 11 & \cellcolor[rgb]{0.529,0.808,0.922}88.84 & \cellcolor[rgb]{0.867,0.627,0.867}31.02 & \cellcolor[rgb]{0.678,0.847,0.902}90.07 & \cellcolor[rgb]{0.565,0.933,0.565}44.10 & 25.49 & 30.05 & \cellcolor[rgb]{1,0.9,0.8}31.79 & 57.86 & 61.61 & \cellcolor[rgb]{1,0.9,0.8}74.50 & 64.41 & \cellcolor[rgb]{1,0.9,0.8}79.34 & \cellcolor[rgb]{0.804,0.498,0.196}76.97 \\
Llama\_3.2\_11B & 12 & \cellcolor[rgb]{0.804,0.498,0.196}90.84 & 26.84 & 84.86 & \cellcolor[rgb]{0.529,0.808,0.922}42.20 & \cellcolor[rgb]{0.565,0.933,0.565}72.40 & 29.61 & 24.06 & 34.08 & 45.83 & 64.14 & \cellcolor[rgb]{1,0.9,0.8}67.97 & 70.72 & 68.50 \\
Phi-3.5\_Vision & 13 & \cellcolor[rgb]{1,0.9,0.8}86.88 & 23.33 & 82.71 & \cellcolor[rgb]{0.678,0.847,0.902}43.93 & 31.32 & \cellcolor[rgb]{1,0.9,0.8}33.07 & 22.15 & 58.01 & 69.47 & 48.38 & 49.98 & 51.05 & 61.35 \\
InternVL2-40B & 14 & 63.66 & \cellcolor[rgb]{1,0.9,0.8}29.59 & 55.06 & 0.00 & 39.81 & \cellcolor[rgb]{0.753,0.753,0.753}62.62 & \cellcolor[rgb]{1,0.419,0.419}57.09 & 60.56 & 66.63 & \cellcolor[rgb]{0.596,0.984,0.596}78.00 & \cellcolor[rgb]{1,0.419,0.419}72.34 & \cellcolor[rgb]{1,0.419,0.419}82.92 & 60.35 \\
LLaVA-NeXT\_7B & 15 & \cellcolor[rgb]{0.596,0.984,0.596}87.08 & 0.00 & 86.71 & \cellcolor[rgb]{1,0.419,0.419}43.45 & 30.54 & 25.41 & 26.43 & 33.54 & 36.81 & 53.94 & 57.34 & 63.21 & \cellcolor[rgb]{0.529,0.808,0.922}73.26 \\
Phi-3\_Vision & 16 & \cellcolor[rgb]{0.678,0.847,0.902}90.54 & 24.99 & 82.70 & \cellcolor[rgb]{1,0.9,0.8}41.00 & 27.84 & 28.65 & 23.84 & 50.31 & \cellcolor[rgb]{0.596,0.984,0.596}75.56 & 36.55 & 51.77 & 52.41 & 55.61 \\
InternVL2-8B & 17 & 86.78 & 24.47 & 85.39 & 0.06 & \cellcolor[rgb]{0.529,0.808,0.922}61.33 & 28.01 & 31.17 & 46.47 & 49.34 & 7.09 & 3.66 & 6.19 & 68.01 \\
PaliGemma\_3B\_448x448 & 18 & 79.68 & \cellcolor[rgb]{0.678,0.847,0.902}31.31 & \cellcolor[rgb]{0.867,0.627,0.867}89.50 & 31.74 & \cellcolor[rgb]{1,0.9,0.8}47.41 & 26.39 & 23.78 & 37.00 & 28.39 & 17.74 & 26.39 & 26.59 & \cellcolor[rgb]{0.867,0.627,0.867}74.15 \\
PaliGemma\_3B\_224x224 & 19 & 79.85 & 26.06 & \cellcolor[rgb]{1,0.419,0.419}89.28 & 35.85 & 42.85 & 25.35 & 22.89 & 23.61 & 25.53 & 12.55 & 10.39 & 13.15 & \cellcolor[rgb]{1,0.419,0.419}73.53 \\
InternVL2-1B & 20 & 31.55 & 0.00 & 47.84 & 0.00 & 14.20 & 24.17 & 24.58 & 15.81 & 25.16 & 0.00 & 0.00 & 0.00 & 34.83 \\
Florence-2 & 21 & 50.98 & 26.96 & 77.61 & 0.00 & 28.27 & 0.00 & 0.00 & 0.11 & 0.11 & 0.08 & 0.17 & 0.80 & 64.93 \\
Chameleon\_7B & 22 & 0.00 & 0.00 & 0.00 & 0.00 & 0.00 & 0.00 & 0.00 & 0.00 & 0.00 & 0.00 & 0.00 & 0.00 & 0.00 \\
\bottomrule
\end{tabular}
\label{tab:model_performance_part1}
\end{table*}

\begin{table*}[h]
\centering
\scriptsize 
\begin{tabular}{l|c|cccccccccccc}
\toprule
Model & Rank & T3.5 & T4.1 & T4.2 & T5.1 & T5.2 & T5.3 & T5.4 & T6.1 & T6.2 & T7.1 & T7.2 & T8.1 \\
\midrule
humans &  & 82.57 & 75.34 & 82.39 & 98.56 & 99.86 & 99.82 & 100.00 & 93.78 & 90.72 & 99.04 & 96.56 & 99.30 \\
\midrule
Gemini\_1.5\_pro & \cellcolor[rgb]{1,0.843,0}1 & \cellcolor[rgb]{0.678,0.847,0.902}69.02 & \cellcolor[rgb]{0.753,0.753,0.753}70.32 & 40.97 & \cellcolor[rgb]{0.753,0.753,0.753}81.41 & 23.64 & \cellcolor[rgb]{1,0.843,0}99.15 & \cellcolor[rgb]{0.678,0.847,0.902}66.80 & \cellcolor[rgb]{0.565,0.933,0.565}78.46 & \cellcolor[rgb]{0.753,0.753,0.753}68.36 & \cellcolor[rgb]{1,0.843,0}53.67 & \cellcolor[rgb]{1,0.843,0}47.86 & \cellcolor[rgb]{0.565,0.933,0.565}79.66 \\
GPT-4o & \cellcolor[rgb]{0.753,0.753,0.753}2 & \cellcolor[rgb]{0.596,0.984,0.596}66.58 & \cellcolor[rgb]{0.529,0.808,0.922}67.21 & \cellcolor[rgb]{0.565,0.933,0.565}51.08 & \cellcolor[rgb]{1,0.843,0}82.01 & \cellcolor[rgb]{0.753,0.753,0.753}46.09 & \cellcolor[rgb]{0.678,0.847,0.902}96.57 & \cellcolor[rgb]{0.529,0.808,0.922}57.50 & \cellcolor[rgb]{0.678,0.847,0.902}75.79 & 43.25 & \cellcolor[rgb]{0.804,0.498,0.196}41.58 & \cellcolor[rgb]{0.753,0.753,0.753}46.66 & \cellcolor[rgb]{1,0.843,0}85.76 \\
Claude\_3.5\_Sonnet & \cellcolor[rgb]{0.804,0.498,0.196}3 & \cellcolor[rgb]{0.804,0.498,0.196}69.86 & 63.54 & \cellcolor[rgb]{0.529,0.808,0.922}47.05 & \cellcolor[rgb]{1,0.419,0.419}46.37 & \cellcolor[rgb]{1,0.843,0}61.09 & \cellcolor[rgb]{0.804,0.498,0.196}98.36 & \cellcolor[rgb]{0.753,0.753,0.753}69.65 & \cellcolor[rgb]{1,0.9,0.8}68.48 & \cellcolor[rgb]{0.565,0.933,0.565}64.15 & \cellcolor[rgb]{0.565,0.933,0.565}39.64 & \cellcolor[rgb]{0.804,0.498,0.196}37.98 & \cellcolor[rgb]{0.678,0.847,0.902}79.61 \\
Qwen2\_72B & \cellcolor[rgb]{0.565,0.933,0.565}4 & \cellcolor[rgb]{0.867,0.627,0.867}68.21 & \cellcolor[rgb]{0.678,0.847,0.902}68.30 & \cellcolor[rgb]{0.804,0.498,0.196}53.98 & \cellcolor[rgb]{0.565,0.933,0.565}52.56 & \cellcolor[rgb]{0.804,0.498,0.196}45.81 & \cellcolor[rgb]{0.753,0.753,0.753}98.63 & \cellcolor[rgb]{0.565,0.933,0.565}66.89 & \cellcolor[rgb]{1,0.419,0.419}74.69 & \cellcolor[rgb]{0.804,0.498,0.196}64.21 & \cellcolor[rgb]{0.678,0.847,0.902}37.12 & \cellcolor[rgb]{0.565,0.933,0.565}32.74 & \cellcolor[rgb]{0.867,0.627,0.867}79.51 \\
Llama\_3.2\_90B & \cellcolor[rgb]{0.678,0.847,0.902}5 & \cellcolor[rgb]{0.529,0.808,0.922}67.43 & \cellcolor[rgb]{1,0.843,0}72.45 & \cellcolor[rgb]{1,0.843,0}58.78 & \cellcolor[rgb]{1,0.9,0.8}35.68 & \cellcolor[rgb]{1,0.9,0.8}28.37 & \cellcolor[rgb]{0.565,0.933,0.565}97.95 & \cellcolor[rgb]{1,0.843,0}75.00 & \cellcolor[rgb]{0.753,0.753,0.753}78.99 & \cellcolor[rgb]{0.678,0.847,0.902}62.49 & 27.54 & 23.22 & \cellcolor[rgb]{1,0.419,0.419}78.99 \\
Gemini\_1.5\_flash & \cellcolor[rgb]{0.867,0.627,0.867}6 & 63.95 & \cellcolor[rgb]{0.867,0.627,0.867}68.17 & \cellcolor[rgb]{0.678,0.847,0.902}49.15 & \cellcolor[rgb]{0.529,0.808,0.922}41.06 & \cellcolor[rgb]{1,0.419,0.419}33.76 & \cellcolor[rgb]{0.867,0.627,0.867}96.46 & \cellcolor[rgb]{0.596,0.984,0.596}52.53 & \cellcolor[rgb]{0.596,0.984,0.596}71.80 & \cellcolor[rgb]{1,0.843,0}73.40 & \cellcolor[rgb]{0.753,0.753,0.753}42.23 & \cellcolor[rgb]{0.596,0.984,0.596}27.55 & \cellcolor[rgb]{0.529,0.808,0.922}77.20 \\
Qwen2\_7B & \cellcolor[rgb]{1,0.419,0.419}7 & 64.52 & \cellcolor[rgb]{0.565,0.933,0.565}68.53 & 39.00 & 33.40 & \cellcolor[rgb]{0.596,0.984,0.596}30.22 & \cellcolor[rgb]{0.529,0.808,0.922}87.46 & \cellcolor[rgb]{1,0.419,0.419}61.72 & \cellcolor[rgb]{0.867,0.627,0.867}75.51 & \cellcolor[rgb]{1,0.419,0.419}56.72 & \cellcolor[rgb]{0.529,0.808,0.922}32.73 & 25.15 & \cellcolor[rgb]{0.596,0.984,0.596}70.42 \\
Molmo\_7B & \cellcolor[rgb]{0.529,0.808,0.922}8 & 58.12 & 60.06 & \cellcolor[rgb]{0.753,0.753,0.753}56.05 & 28.79 & \cellcolor[rgb]{0.678,0.847,0.902}35.34 & 69.54 & \cellcolor[rgb]{0.804,0.498,0.196}67.60 & \cellcolor[rgb]{0.804,0.498,0.196}78.68 & 53.21 & 28.87 & 26.05 & \cellcolor[rgb]{0.753,0.753,0.753}84.32 \\
Pixtral & \cellcolor[rgb]{0.596,0.984,0.596}9 & 63.37 & \cellcolor[rgb]{0.804,0.498,0.196}70.13 & 36.51 & 28.00 & \cellcolor[rgb]{0.565,0.933,0.565}35.47 & \cellcolor[rgb]{1,0.419,0.419}94.85 & 38.95 & 63.01 & 32.66 & \cellcolor[rgb]{1,0.9,0.8}30.41 & \cellcolor[rgb]{0.867,0.627,0.867}29.98 & \cellcolor[rgb]{0.804,0.498,0.196}81.35 \\
GPT-4o-mini & \cellcolor[rgb]{1,0.9,0.8}10 & \cellcolor[rgb]{1,0.843,0}72.69 & 57.68 & \cellcolor[rgb]{1,0.9,0.8}45.14 & \cellcolor[rgb]{0.867,0.627,0.867}48.57 & \cellcolor[rgb]{0.867,0.627,0.867}34.58 & \cellcolor[rgb]{1,0.9,0.8}81.02 & 51.40 & 49.48 & \cellcolor[rgb]{0.596,0.984,0.596}55.02 & \cellcolor[rgb]{1,0.419,0.419}33.01 & \cellcolor[rgb]{1,0.419,0.419}29.64 & 59.98 \\
LLaVA-NeXt\_34B & 11 & \cellcolor[rgb]{1,0.9,0.8}65.84 & \cellcolor[rgb]{1,0.419,0.419}67.52 & \cellcolor[rgb]{0.596,0.984,0.596}46.67 & \cellcolor[rgb]{0.596,0.984,0.596}35.69 & 24.83 & 39.62 & 51.00 & \cellcolor[rgb]{0.529,0.808,0.922}72.04 & 52.33 & 28.19 & 26.04 & 53.84 \\
Llama\_3.2\_11B & 12 & 63.69 & 55.04 & 36.83 & 7.73 & 26.22 & 46.86 & \cellcolor[rgb]{0.867,0.627,0.867}62.65 & 62.63 & \cellcolor[rgb]{1,0.9,0.8}54.41 & 25.69 & \cellcolor[rgb]{1,0.9,0.8}26.14 & 38.49 \\
Phi-3.5\_Vision & 13 & 64.30 & \cellcolor[rgb]{1,0.9,0.8}66.02 & 40.56 & 17.26 & 27.28 & 51.81 & 51.41 & 45.56 & \cellcolor[rgb]{0.867,0.627,0.867}61.98 & \cellcolor[rgb]{0.596,0.984,0.596}30.67 & 26.10 & 52.27 \\
InternVL2-40B & 14 & 56.75 & 9.18 & 0.00 & \cellcolor[rgb]{0.804,0.498,0.196}60.92 & \cellcolor[rgb]{0.529,0.808,0.922}33.40 & \cellcolor[rgb]{0.596,0.984,0.596}82.67 & 16.54 & \cellcolor[rgb]{1,0.843,0}79.59 & 11.92 & \cellcolor[rgb]{0.867,0.627,0.867}34.83 & \cellcolor[rgb]{0.678,0.847,0.902}32.43 & \cellcolor[rgb]{1,0.9,0.8}65.90 \\
LLaVA-NeXT\_7B & 15 & \cellcolor[rgb]{0.565,0.933,0.565}69.50 & 62.59 & 35.97 & 23.37 & 26.02 & 24.81 & 49.25 & 56.65 & \cellcolor[rgb]{0.529,0.808,0.922}55.59 & 23.77 & 24.97 & 28.64 \\
Phi-3\_Vision & 16 & 58.98 & \cellcolor[rgb]{0.596,0.984,0.596}66.22 & \cellcolor[rgb]{1,0.419,0.419}47.50 & 22.42 & 26.04 & 46.52 & 18.66 & 42.42 & 16.59 & 27.91 & 25.88 & 44.22 \\
InternVL2-8B & 17 & 64.84 & 59.73 & 0.00 & \cellcolor[rgb]{0.678,0.847,0.902}51.82 & 27.64 & 56.74 & \cellcolor[rgb]{1,0.9,0.8}51.50 & 18.60 & 51.13 & 27.32 & \cellcolor[rgb]{0.529,0.808,0.922}28.19 & 54.81 \\
PaliGemma\_3B\_448x448 & 18 & \cellcolor[rgb]{1,0.419,0.419}67.97 & 60.55 & \cellcolor[rgb]{0.867,0.627,0.867}47.86 & 16.22 & 25.23 & 33.53 & 41.82 & 27.86 & 51.58 & 24.47 & 22.28 & 28.83 \\
PaliGemma\_3B\_224x224 & 19 & \cellcolor[rgb]{0.753,0.753,0.753}71.00 & 58.38 & 31.76 & 23.41 & 26.30 & 24.59 & 34.08 & 19.91 & 37.08 & 22.13 & 24.20 & 27.14 \\
InternVL2-1B & 20 & 25.56 & 14.39 & 0.00 & 24.38 & 10.64 & 24.42 & 12.82 & 0.00 & 5.82 & 23.82 & 22.14 & 24.71 \\
Florence-2 & 21 & 57.33 & 49.88 & 0.00 & 0.00 & 0.70 & 0.00 & 0.17 & 0.26 & 3.56 & 4.83 & 0.00 & 0.65 \\
Chameleon\_7B & 22 & 0.00 & 0.00 & 0.00 & 0.00 & 0.00 & 0.00 & 0.00 & 0.00 & 0.00 & 0.00 & 0.00 & 0.00 \\
\bottomrule
\end{tabular}
\caption{\textbf{Model performance across tasks displayed as regular Accuracy.} For each column, the top 10 models are highlighted: \textcolor[rgb]{1,0.843,0}{\rule{0.5cm}{0.5cm}} 1st place (Gold) \quad \textcolor[rgb]{0.753,0.753,0.753}{\rule{0.5cm}{0.5cm}} 2nd place (Silver) \quad \textcolor[rgb]{0.804,0.498,0.196}{\rule{0.5cm}{0.5cm}} 3rd place (Bronze) \quad \textcolor[rgb]{0.565,0.933,0.565}{\rule{0.5cm}{0.5cm}} 4th place \quad \textcolor[rgb]{0.678,0.847,0.902}{\rule{0.5cm}{0.5cm}} 5th place \quad \textcolor[rgb]{0.867,0.627,0.867}{\rule{0.5cm}{0.5cm}} 6th place \quad \textcolor[rgb]{1,0.419,0.419}{\rule{0.5cm}{0.5cm}} 7th place \quad \textcolor[rgb]{0.529,0.808,0.922}{\rule{0.5cm}{0.5cm}} 8th place \quad \textcolor[rgb]{0.596,0.984,0.596}{\rule{0.5cm}{0.5cm}} 9th place \quad \textcolor[rgb]{1,0.9,0.8}{\rule{0.5cm}{0.5cm}} 10th place. }
\label{tab:model_performance_part2}
\end{table*}

\newpage

\newpage

\subsection{Accuracy\%(t) Curves for All Models}
\label{appendix_full_auc_curves}
\begin{figure}[H]
    \centering
    \includegraphics[width=\linewidth]{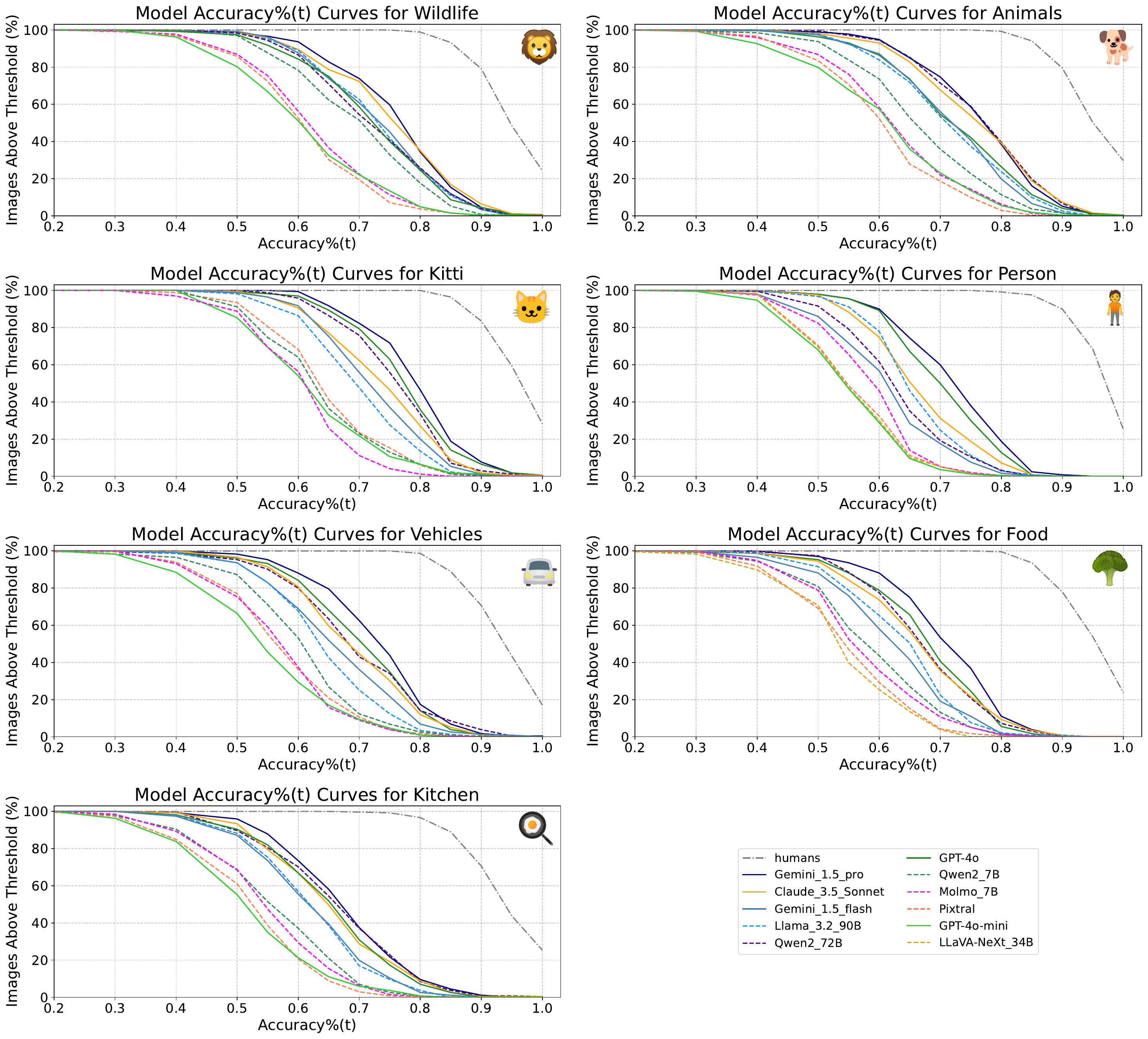}
    \caption{\textbf{Accuracy\% (t) curves for the top 10 models across each dataset,} with a maximum score of 1. Humans (dashed-dotted grey line) consistently achieve the highest performance. Dashed lines indicate open-source models, while solid lines represent closed-source models. The area under the Accuracy\% (t) curves, detailed in \autoref{tab:model_aucs}, highlights significant variations in model rankings across domain-specific datasets for the same tasks.}
    \label{fig:app_auc_curves_top_10}
\end{figure}

\begin{figure}[h]
    \centering
    \includegraphics[width=\linewidth]{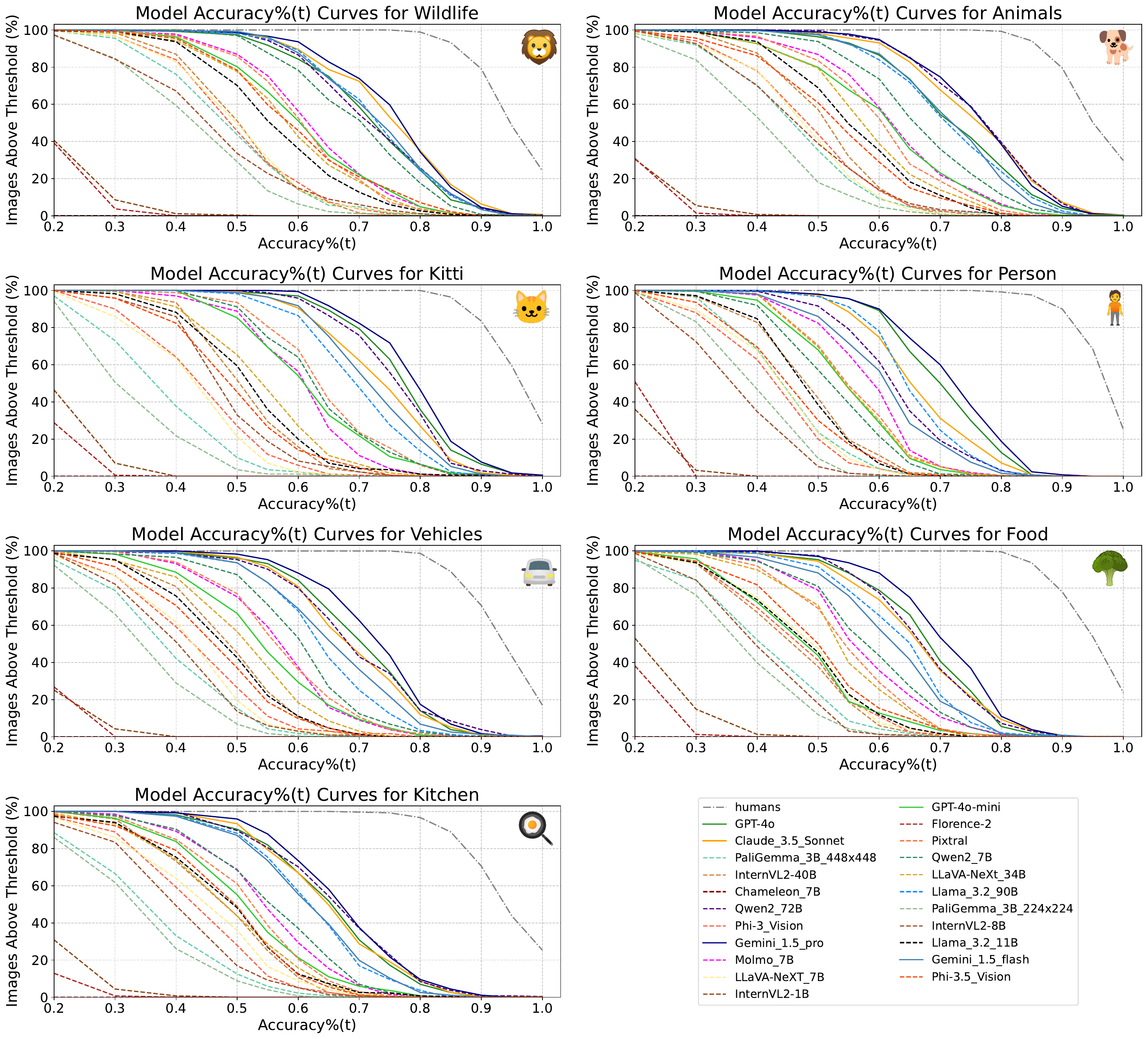}
    \caption{\textbf{Accuracy\% (t) curves for all models across each dataset,} with a maximum score of 1. Humans (dashed-dotted grey line) consistently achieve the highest performance. Dashed lines indicate open-source models, while solid lines represent closed-source models. The area under the Accuracy\% (t) curves, detailed in \autoref{tab:model_aucs}, highlights significant variations in model rankings across domain-specific datasets for the same tasks.}
    \label{fig:app_auc_curves_all_models}
\end{figure}

Here we present the Accuracy\%(t) curves for all models and datasets.
The Accuracy\%(t) metric represents the percentage of images for which at least a specified proportion of questions are correctly answered. The thresholds for each curve are [0.2, 0.3, 0.4, 0.5, 0.55, 0.6, 0.65, 0.7, 0.75, 0.8, 0.85, 0.9, 0.95, 1.0]. First we display the top 10 models per dataset in \autoref{fig:app_auc_curves_top_10}, for all 22 models in \autoref{fig:app_auc_curves_all_models}, and Area under the Accuracy\%(t) Curves in \autoref{tab:model_aucs}.

\newpage

\FloatBarrier  
\newpage

\begin{table*}[h]
\centering
\begin{tabular}{l|c|ccccccc}
\toprule
Model & Overall & animals & food & kitchen & kitti & person & vehicles & wildlife \\
&  & \includegraphics[width=0.03\textwidth]{figures/icons_iclr/animals.png} & \includegraphics[width=0.03\textwidth]{figures/icons_iclr/food.png} & \includegraphics[width=0.03\textwidth]{figures/icons_iclr/kitchen.png} & \includegraphics[width=0.03\textwidth]{figures/icons_iclr/kitti.png} & \includegraphics[width=0.03\textwidth]{figures/icons_iclr/person.png} & \includegraphics[width=0.03\textwidth]{figures/icons_iclr/vehicles.png} & \includegraphics[width=0.03\textwidth]{figures/icons_iclr/wildlife.png} \\
\midrule
humans & 74.28 & 74.40 & 74.34 & 73.08 & 75.18 & 75.89 & 73.01 & 74.10 \\
\toprule
Gemini\_1.5\_pro & \cellcolor[rgb]{1,0.843,0}52.95 & \cellcolor[rgb]{0.753,0.753,0.753}56.02 & \cellcolor[rgb]{1,0.843,0}50.39 & \cellcolor[rgb]{1,0.843,0}46.81 & \cellcolor[rgb]{1,0.843,0}58.51 & \cellcolor[rgb]{1,0.843,0}51.35 & \cellcolor[rgb]{1,0.843,0}52.17 & \cellcolor[rgb]{1,0.843,0}55.42 \\
GPT-4o & \cellcolor[rgb]{0.753,0.753,0.753}50.19 & \cellcolor[rgb]{0.565,0.933,0.565}51.87 & \cellcolor[rgb]{0.753,0.753,0.753}47.24 & \cellcolor[rgb]{0.565,0.933,0.565}44.52 & \cellcolor[rgb]{0.753,0.753,0.753}56.74 & \cellcolor[rgb]{0.753,0.753,0.753}49.59 & \cellcolor[rgb]{0.753,0.753,0.753}49.72 & \cellcolor[rgb]{0.867,0.627,0.867}51.64 \\
Claude\_3.5\_Sonnet & \cellcolor[rgb]{0.804,0.498,0.196}49.80 & \cellcolor[rgb]{0.804,0.498,0.196}55.36 & \cellcolor[rgb]{0.565,0.933,0.565}46.24 & \cellcolor[rgb]{0.804,0.498,0.196}44.73 & \cellcolor[rgb]{0.565,0.933,0.565}53.03 & \cellcolor[rgb]{0.804,0.498,0.196}45.92 & \cellcolor[rgb]{0.565,0.933,0.565}48.48 & \cellcolor[rgb]{0.753,0.753,0.753}54.85 \\
Qwen2\_72B & \cellcolor[rgb]{0.565,0.933,0.565}49.57 & \cellcolor[rgb]{1,0.843,0}56.15 & \cellcolor[rgb]{0.804,0.498,0.196}46.89 & \cellcolor[rgb]{0.753,0.753,0.753}45.63 & \cellcolor[rgb]{0.804,0.498,0.196}55.38 & \cellcolor[rgb]{0.678,0.847,0.902}42.27 & \cellcolor[rgb]{0.804,0.498,0.196}48.85 & \cellcolor[rgb]{0.678,0.847,0.902}51.82 \\
Llama\_3.2\_90B & \cellcolor[rgb]{0.678,0.847,0.902}46.57 & \cellcolor[rgb]{0.867,0.627,0.867}51.07 & \cellcolor[rgb]{0.678,0.847,0.902}43.16 & \cellcolor[rgb]{0.678,0.847,0.902}41.45 & \cellcolor[rgb]{0.867,0.627,0.867}49.25 & \cellcolor[rgb]{0.565,0.933,0.565}45.00 & \cellcolor[rgb]{0.867,0.627,0.867}43.66 & \cellcolor[rgb]{0.565,0.933,0.565}52.42 \\
Gemini\_1.5\_flash & \cellcolor[rgb]{0.867,0.627,0.867}46.33 & \cellcolor[rgb]{0.678,0.847,0.902}51.10 & \cellcolor[rgb]{0.867,0.627,0.867}41.61 & \cellcolor[rgb]{0.867,0.627,0.867}41.39 & \cellcolor[rgb]{0.678,0.847,0.902}51.56 & \cellcolor[rgb]{0.867,0.627,0.867}40.45 & \cellcolor[rgb]{0.678,0.847,0.902}45.60 & \cellcolor[rgb]{0.804,0.498,0.196}52.63 \\
Qwen2\_7B & \cellcolor[rgb]{1,0.419,0.419}40.35 & \cellcolor[rgb]{1,0.419,0.419}46.08 & \cellcolor[rgb]{1,0.419,0.419}37.87 & \cellcolor[rgb]{1,0.419,0.419}34.94 & \cellcolor[rgb]{0.529,0.808,0.922}42.81 & 31.93 & \cellcolor[rgb]{1,0.419,0.419}39.73 & \cellcolor[rgb]{1,0.419,0.419}49.11 \\
Molmo\_7B & \cellcolor[rgb]{0.529,0.808,0.922}38.27 & \cellcolor[rgb]{0.529,0.808,0.922}41.84 & \cellcolor[rgb]{0.529,0.808,0.922}36.78 & \cellcolor[rgb]{0.529,0.808,0.922}33.96 & \cellcolor[rgb]{1,0.9,0.8}39.78 & \cellcolor[rgb]{1,0.419,0.419}37.60 & \cellcolor[rgb]{0.596,0.984,0.596}36.29 & \cellcolor[rgb]{0.529,0.808,0.922}41.65 \\
Pixtral & \cellcolor[rgb]{0.596,0.984,0.596}37.34 & \cellcolor[rgb]{1,0.9,0.8}40.00 & \cellcolor[rgb]{0.596,0.984,0.596}34.23 & \cellcolor[rgb]{0.596,0.984,0.596}31.42 & \cellcolor[rgb]{1,0.419,0.419}43.69 & \cellcolor[rgb]{0.529,0.808,0.922}35.06 & \cellcolor[rgb]{0.529,0.808,0.922}36.56 & \cellcolor[rgb]{0.596,0.984,0.596}40.45 \\
GPT-4o-mini & \cellcolor[rgb]{1,0.9,0.8}35.50 & \cellcolor[rgb]{0.596,0.984,0.596}40.49 & 27.34 & \cellcolor[rgb]{1,0.9,0.8}31.03 & \cellcolor[rgb]{0.596,0.984,0.596}41.31 & \cellcolor[rgb]{1,0.9,0.8}34.08 & \cellcolor[rgb]{1,0.9,0.8}34.00 & \cellcolor[rgb]{1,0.9,0.8}40.29 \\
LLaVA-NeXt\_34B & 33.65 & 37.26 & \cellcolor[rgb]{1,0.9,0.8}33.23 & 27.67 & 33.63 & \cellcolor[rgb]{0.596,0.984,0.596}34.72 & 30.71 & 38.28 \\
Llama\_3.2\_11B & 30.44 & 35.35 & 27.17 & 28.06 & 31.75 & 27.44 & 27.19 & 36.10 \\
random\_chance & 30.24 & 29.39 & 30.31 & 30.86 & 29.78 & 30.39 & 29.70 & 31.27 \\
Phi-3.5\_Vision & 29.87 & 32.93 & 29.22 & 28.01 & 28.96 & 24.93 & 25.67 & 39.35 \\
InternVL2-40B & 28.48 & 30.60 & 25.16 & 27.14 & 30.76 & 27.85 & 27.89 & 29.95 \\
LLaVA-NeXT\_7B & 24.94 & 26.88 & 27.10 & 24.04 & 22.03 & 23.12 & 21.81 & 29.64 \\
Phi-3\_Vision & 24.89 & 27.88 & 26.05 & 22.74 & 23.41 & 22.12 & 22.60 & 29.44 \\
InternVL2-8B & 22.21 & 26.79 & 19.83 & 20.09 & 27.41 & 16.17 & 19.81 & 25.37 \\
PaliGemma\_3B\_448x448 & 21.44 & 25.70 & 20.86 & 15.81 & 17.04 & 24.13 & 18.40 & 28.16 \\
PaliGemma\_3B\_224x224 & 17.28 & 20.82 & 17.61 & 13.99 & 12.35 & 18.80 & 14.72 & 22.65 \\
InternVL2-1B & 2.62 & 2.14 & 4.26 & 2.06 & 3.03 & 2.12 & 1.68 & 3.03 \\
Florence-2 & 1.74 & 1.68 & 2.03 & 0.72 & 1.50 & 2.54 & 1.34 & 2.33 \\
Chameleon\_7B & 0.00 & 0.00 & 0.00 & 0.00 & 0.00 & 0.00 & 0.00 & 0.00 \\
\bottomrule
\end{tabular}
\caption{\textbf{Area under Accuracy\%(t) Curves for seven different datasets}. Curves calculated for threshholds at [0.2, 0.3, 0.4, 0.5, 0.55, 0.6, 0.65, 0.7, 0.75, 0.8, 0.85, 0.9, 0.95, 1.0] across each dataset. For each column, the top 10 models are highlighted: \textcolor[rgb]{1,0.843,0}{\rule{0.5cm}{0.5cm}} 1st place (Gold) \quad \textcolor[rgb]{0.753,0.753,0.753}{\rule{0.5cm}{0.5cm}} 2nd place (Silver) \quad \textcolor[rgb]{0.804,0.498,0.196}{\rule{0.5cm}{0.5cm}} 3rd place (Bronze) \quad \textcolor[rgb]{0.565,0.933,0.565}{\rule{0.5cm}{0.5cm}} 4th place \quad \textcolor[rgb]{0.678,0.847,0.902}{\rule{0.5cm}{0.5cm}} 5th place \quad \textcolor[rgb]{0.867,0.627,0.867}{\rule{0.5cm}{0.5cm}} 6th place \quad \textcolor[rgb]{1,0.419,0.419}{\rule{0.5cm}{0.5cm}} 7th place \quad \textcolor[rgb]{0.529,0.808,0.922}{\rule{0.5cm}{0.5cm}} 8th place \quad \textcolor[rgb]{0.596,0.984,0.596}{\rule{0.5cm}{0.5cm}} 9th place \quad \textcolor[rgb]{1,0.9,0.8}{\rule{0.5cm}{0.5cm}} 10th place. The 'Overall' column represents the mean area under the curve across all datasets. The substantial differences in model rankings across domain-specific datasets for identical tasks highlights the need for specific in-domain evaluation.}
\label{tab:model_aucs}
\end{table*}

\newpage

\subsection{Task Correlation}
\label{appendix_placeholder}

\begin{figure}[h]
    \centering
    \includegraphics[width=\linewidth]{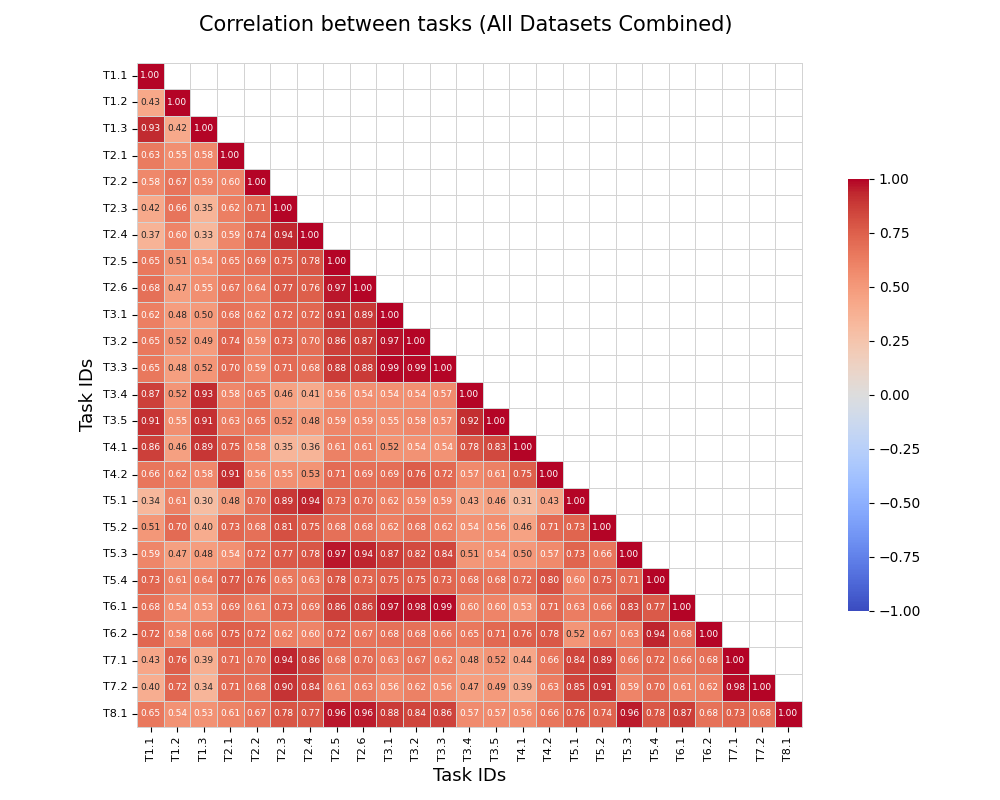}
    \caption{\textbf{Task performance correlations reveals insights on related visual capabilities.} Heatmap visualizing pairwise correlations between task performances across all datasets, with values ranging from -1 (anti-correlated) to 1 (highly correlated).}
    \label{fig:app_corr_heatmap}
\end{figure}

\begin{figure}[H]
    \centering
    \includegraphics[width=\linewidth]{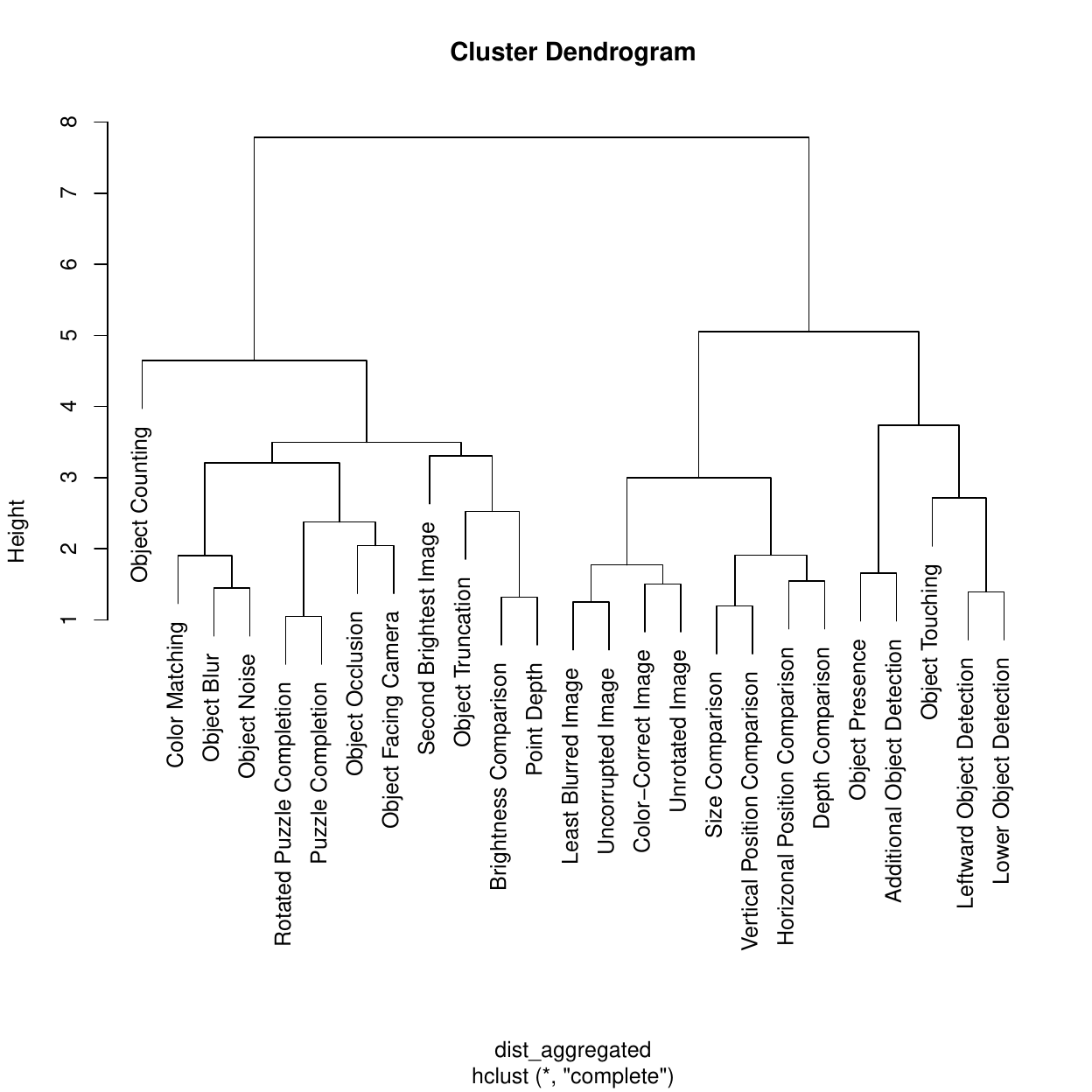}
    \caption{\textbf{Hierarchical clustering reveals natural groupings of visually related tasks across domains.} Dendrogram visualization of task relationships based on model performance across seven domains, confirming and extending the correlation patterns observed in the heatmap analysis.}
    \label{fig:app_corr_dendrogram}
\end{figure}

Here we present the correlation between tasks, quantified in the heatmap in~\autoref{fig:app_corr_heatmap} and visualized in the dendrogram in~\autoref{fig:app_corr_dendrogram}. The heatmap shows pairwise task correlations, while the dendrogram highlights clusters of tasks with similar performance patterns across models

\

\newpage

\subsection{Ranking Diversity Across Domains}